\documentclass[10pt,twocolumn,letterpaper]{article}
\usepackage[pagenumbers]{cvpr}

\usepackage{cuted} 

\usepackage{currfile} 

\usepackage{caption} 

\usepackage{booktabs}
\usepackage{array}
\usepackage{amsmath}
\usepackage{xcolor}
\usepackage{graphicx}
\usepackage{tikz}

\providecommand{\blank}{\underline{\hspace{1.2em}}}
\newcommand{\testTwoNPairs}{400}

\newcommand{\testTwoPassed}{9}

\newcommand{\testTwoGemmaO}{54}
\newcommand{\testTwoGemmaE}{0.94}
\newcommand{\testTwoGemmaAligned}{91}
\newcommand{\testTwoGemmaExcess}{58.3}
\newcommand{\testTwoGemmaExcessLo}{41.8}
\newcommand{\testTwoGemmaExcessHi}{73.9}
\newcommand{\testTwoGemmaRatio}{57.4}
\newcommand{\testTwoGemmaRatioLo}{22.8}
\newcommand{\testTwoGemmaRatioHi}{122.9}
\newcommand{\testTwoGemmaRepair}{4.5}

\newcommand{\testTwoGemmaTwelveO}{100}
\newcommand{\testTwoGemmaTwelveE}{4.44}
\newcommand{\testTwoGemmaTwelveAligned}{194}
\newcommand{\testTwoGemmaTwelveExcess}{49.3}
\newcommand{\testTwoGemmaTwelveExcessLo}{36.5}
\newcommand{\testTwoGemmaTwelveExcessHi}{63.2}
\newcommand{\testTwoGemmaTwelveRatio}{22.5}
\newcommand{\testTwoGemmaTwelveRatioLo}{12.3}
\newcommand{\testTwoGemmaTwelveRatioHi}{56.6}
\newcommand{\testTwoGemmaTwelveRepair}{8.3}

\newcommand{\testTwoQwenO}{51}
\newcommand{\testTwoQwenE}{3.83}
\newcommand{\testTwoQwenAligned}{167}
\newcommand{\testTwoQwenExcess}{28.2}
\newcommand{\testTwoQwenExcessLo}{19.0}
\newcommand{\testTwoQwenExcessHi}{39.3}
\newcommand{\testTwoQwenRatio}{13.3}
\newcommand{\testTwoQwenRatioLo}{7.3}
\newcommand{\testTwoQwenRatioHi}{54.6}
\newcommand{\testTwoQwenRepair}{4.2}

\newcommand{\testTwoQwenVLO}{125}
\newcommand{\testTwoQwenVLE}{10.24}
\newcommand{\testTwoQwenVLAligned}{252}
\newcommand{\testTwoQwenVLExcess}{45.5}
\newcommand{\testTwoQwenVLExcessLo}{34.7}
\newcommand{\testTwoQwenVLExcessHi}{55.9}
\newcommand{\testTwoQwenVLRatio}{12.2}
\newcommand{\testTwoQwenVLRatioLo}{6.7}
\newcommand{\testTwoQwenVLRatioHi}{25.3}
\newcommand{\testTwoQwenVLRepair}{10.4}

\newcommand{\testTwoQwenFourO}{84}
\newcommand{\testTwoQwenFourE}{6.03}
\newcommand{\testTwoQwenFourAligned}{216}
\newcommand{\testTwoQwenFourExcess}{36.1}
\newcommand{\testTwoQwenFourExcessLo}{25.8}
\newcommand{\testTwoQwenFourExcessHi}{46.7}
\newcommand{\testTwoQwenFourRatio}{13.9}
\newcommand{\testTwoQwenFourRatioLo}{6.9}
\newcommand{\testTwoQwenFourRatioHi}{32.6}
\newcommand{\testTwoQwenFourRepair}{7.0}

\newcommand{\testTwoInternO}{99}
\newcommand{\testTwoInternE}{4.99}
\newcommand{\testTwoInternAligned}{206}
\newcommand{\testTwoInternExcess}{45.6}
\newcommand{\testTwoInternExcessLo}{34.5}
\newcommand{\testTwoInternExcessHi}{57.7}
\newcommand{\testTwoInternRatio}{19.8}
\newcommand{\testTwoInternRatioLo}{9.9}
\newcommand{\testTwoInternRatioHi}{37.6}
\newcommand{\testTwoInternRepair}{8.2}

\newcommand{\testTwoMistralO}{69}
\newcommand{\testTwoMistralE}{8.29}
\newcommand{\testTwoMistralAligned}{194}
\newcommand{\testTwoMistralExcess}{31.3}
\newcommand{\testTwoMistralExcessLo}{20.4}
\newcommand{\testTwoMistralExcessHi}{43.4}
\newcommand{\testTwoMistralRatio}{8.3}
\newcommand{\testTwoMistralRatioLo}{4.8}
\newcommand{\testTwoMistralRatioHi}{20.4}
\newcommand{\testTwoMistralRepair}{5.8}

\newcommand{\testTwoGlmO}{22}
\newcommand{\testTwoGlmE}{0.70}
\newcommand{\testTwoGlmAligned}{83}
\newcommand{\testTwoGlmExcess}{25.7}
\newcommand{\testTwoGlmExcessLo}{10.2}
\newcommand{\testTwoGlmExcessHi}{43.1}
\newcommand{\testTwoGlmRatio}{31.2}
\newcommand{\testTwoGlmRatioLo}{8.3}
\newcommand{\testTwoGlmRatioHi}{160.4}
\newcommand{\testTwoGlmRepair}{1.8}

\newcommand{\testTwoLlavaO}{50}
\newcommand{\testTwoLlavaE}{4.07}
\newcommand{\testTwoLlavaAligned}{121}
\newcommand{\testTwoLlavaExcess}{38.0}
\newcommand{\testTwoLlavaExcessLo}{26.0}
\newcommand{\testTwoLlavaExcessHi}{51.4}
\newcommand{\testTwoLlavaRatio}{12.3}
\newcommand{\testTwoLlavaRatioLo}{5.4}
\newcommand{\testTwoLlavaRatioHi}{38.7}
\newcommand{\testTwoLlavaRepair}{4.2}

\newcommand{\testTwoMolmoO}{12}
\newcommand{\testTwoMolmoE}{2.65}
\newcommand{\testTwoMolmoAligned}{85}
\newcommand{\testTwoMolmoExcess}{11.0}
\newcommand{\testTwoMolmoExcessLo}{2.0}
\newcommand{\testTwoMolmoExcessHi}{22.0}
\newcommand{\testTwoMolmoRatio}{4.5}
\newcommand{\testTwoMolmoRatioLo}{1.5}
\newcommand{\testTwoMolmoRatioHi}{13.6}
\newcommand{\testTwoMolmoRepair}{1.0}

\newcommand{\testTwoRatioMin}{4.5}

\newcommand{\testTwoRatioPassedMin}{8.3}
\newcommand{\testTwoRatioPassedMax}{57.4}

\newcommand{\testTwoCiFactorMedian}{4.7}

\newcommand{\testTwoRecEMin}{0.18}
\newcommand{\testTwoRecEMax}{0.56}

\newcommand{\testTwoRecESum}{1.5}

\newcommand{\testTwoSmoothMaxShift}{1.2}

\newcommand{\testThreeNItems}{300}
\newcommand{\testThreeNWriters}{101}
\newcommand{\testThreePassed}{10}
\newcommand{\testThreeGemmaDelta}{7.0}
\newcommand{\testThreeGemmaDeltaLo}{3.5}
\newcommand{\testThreeGemmaDeltaHi}{11.0}
\newcommand{\testThreeGemmaPos}{2.1}
\newcommand{\testThreeGemmaPosLo}{0.4}
\newcommand{\testThreeGemmaPosHi}{4.5}

\newcommand{\testThreeGemmaCropDelta}{4.7}
\newcommand{\testThreeGemmaYN}{6.6}
\newcommand{\testThreeGemmaXN}{\ensuremath{-}0.4}
\newcommand{\testThreeGemmaTwelveDelta}{23.9}
\newcommand{\testThreeGemmaTwelveDeltaLo}{17.4}
\newcommand{\testThreeGemmaTwelveDeltaHi}{30.6}
\newcommand{\testThreeGemmaTwelvePos}{1.9}
\newcommand{\testThreeGemmaTwelvePosLo}{0.2}
\newcommand{\testThreeGemmaTwelvePosHi}{4.1}

\newcommand{\testThreeGemmaTwelveCropDelta}{21.6}
\newcommand{\testThreeGemmaTwelveYN}{23.0}
\newcommand{\testThreeGemmaTwelveXN}{\ensuremath{-}0.9}
\newcommand{\testThreeQwenDelta}{6.7}
\newcommand{\testThreeQwenDeltaLo}{3.8}
\newcommand{\testThreeQwenDeltaHi}{10.1}
\newcommand{\testThreeQwenPos}{4.4}
\newcommand{\testThreeQwenPosLo}{2.2}
\newcommand{\testThreeQwenPosHi}{7.1}

\newcommand{\testThreeQwenCropDelta}{3.1}
\newcommand{\testThreeQwenYN}{6.0}
\newcommand{\testThreeQwenXN}{\ensuremath{-}0.7}
\newcommand{\testThreeQwenVLDelta}{29.1}
\newcommand{\testThreeQwenVLDeltaLo}{23.1}
\newcommand{\testThreeQwenVLDeltaHi}{35.6}
\newcommand{\testThreeQwenVLPos}{2.9}
\newcommand{\testThreeQwenVLPosLo}{0.8}
\newcommand{\testThreeQwenVLPosHi}{5.2}

\newcommand{\testThreeQwenVLCropDelta}{27.5}
\newcommand{\testThreeQwenVLYN}{27.2}
\newcommand{\testThreeQwenVLXN}{\ensuremath{-}1.9}
\newcommand{\testThreeQwenFourDelta}{22.4}
\newcommand{\testThreeQwenFourDeltaLo}{17.3}
\newcommand{\testThreeQwenFourDeltaHi}{27.8}
\newcommand{\testThreeQwenFourPos}{0.4}
\newcommand{\testThreeQwenFourPosLo}{\ensuremath{-}1.8}
\newcommand{\testThreeQwenFourPosHi}{2.6}

\newcommand{\testThreeQwenFourCropDelta}{21.9}
\newcommand{\testThreeQwenFourYN}{21.9}
\newcommand{\testThreeQwenFourXN}{\ensuremath{-}0.6}
\newcommand{\testThreeInternDelta}{12.0}
\newcommand{\testThreeInternDeltaLo}{8.0}
\newcommand{\testThreeInternDeltaHi}{16.3}
\newcommand{\testThreeInternPos}{0.0}
\newcommand{\testThreeInternPosLo}{\ensuremath{-}0.9}
\newcommand{\testThreeInternPosHi}{0.9}

\newcommand{\testThreeInternCropDelta}{9.8}
\newcommand{\testThreeInternYN}{11.4}
\newcommand{\testThreeInternXN}{\ensuremath{-}0.6}
\newcommand{\testThreeMistralDelta}{7.7}
\newcommand{\testThreeMistralDeltaLo}{4.6}
\newcommand{\testThreeMistralDeltaHi}{11.2}
\newcommand{\testThreeMistralPos}{2.0}
\newcommand{\testThreeMistralPosLo}{\ensuremath{-}0.4}
\newcommand{\testThreeMistralPosHi}{4.7}

\newcommand{\testThreeMistralCropDelta}{7.4}
\newcommand{\testThreeMistralYN}{7.2}
\newcommand{\testThreeMistralXN}{\ensuremath{-}0.4}
\newcommand{\testThreeGlmDelta}{2.1}
\newcommand{\testThreeGlmDeltaLo}{0.5}
\newcommand{\testThreeGlmDeltaHi}{4.3}
\newcommand{\testThreeGlmPos}{0.3}
\newcommand{\testThreeGlmPosLo}{0.0}
\newcommand{\testThreeGlmPosHi}{1.2}

\newcommand{\testThreeGlmCropDelta}{0.0}
\newcommand{\testThreeGlmYN}{0.9}
\newcommand{\testThreeGlmXN}{\ensuremath{-}1.2}
\newcommand{\testThreeLlavaDelta}{8.8}
\newcommand{\testThreeLlavaDeltaLo}{5.3}
\newcommand{\testThreeLlavaDeltaHi}{12.7}
\newcommand{\testThreeLlavaPos}{0.2}
\newcommand{\testThreeLlavaPosLo}{\ensuremath{-}0.3}
\newcommand{\testThreeLlavaPosHi}{1.0}

\newcommand{\testThreeLlavaCropDelta}{5.6}
\newcommand{\testThreeLlavaYN}{7.8}
\newcommand{\testThreeLlavaXN}{\ensuremath{-}1.0}
\newcommand{\testThreeMolmoDelta}{4.4}
\newcommand{\testThreeMolmoDeltaLo}{2.0}
\newcommand{\testThreeMolmoDeltaHi}{7.4}
\newcommand{\testThreeMolmoPos}{0.0}
\newcommand{\testThreeMolmoPosLo}{\ensuremath{-}1.7}
\newcommand{\testThreeMolmoPosHi}{1.8}

\newcommand{\testThreeMolmoCropDelta}{3.1}
\newcommand{\testThreeMolmoYN}{4.1}
\newcommand{\testThreeMolmoXN}{\ensuremath{-}0.3}

\newcommand{\testThreeDeltaPassedMin}{2.1}
\newcommand{\testThreeDeltaPassedMax}{29.1}
\newcommand{\testThreeCropDeltaMin}{0.0}
\newcommand{\testThreeCropDeltaMax}{27.5}
\newcommand{\testThreePosMin}{0.0}
\newcommand{\testThreePosMax}{4.4}
\newcommand{\testThreePosAbove}{4}

\newcommand{\testThreeInternPosUnparsedMin}{71}
\newcommand{\testThreeInternPosUnparsedMax}{73}

\newcommand{\revRecReadings}{1{,}600}
\newcommand{\revRecRepairs}{3}
\newcommand{\revRecPowerMin}{0.55}
\newcommand{\revRecPowerMax}{0.92}

\newcommand{\revThreeGemmaParseX}{96}
\newcommand{\revThreeGemmaParseY}{96}
\newcommand{\revThreeGemmaTwelveParseX}{93}
\newcommand{\revThreeGemmaTwelveParseY}{90}
\newcommand{\revThreeQwenParseX}{93}
\newcommand{\revThreeQwenParseY}{89}
\newcommand{\revThreeQwenVLParseX}{90}
\newcommand{\revThreeQwenVLParseY}{87}
\newcommand{\revThreeQwenFourParseX}{89}
\newcommand{\revThreeQwenFourParseY}{88}
\newcommand{\revThreeInternParseX}{93}
\newcommand{\revThreeInternParseY}{84}
\newcommand{\revThreeMistralParseX}{35}
\newcommand{\revThreeMistralParseY}{33}
\newcommand{\revThreeGlmParseX}{87}
\newcommand{\revThreeGlmParseY}{83}
\newcommand{\revThreeLlavaParseX}{91}
\newcommand{\revThreeLlavaParseY}{88}
\newcommand{\revThreeMolmoParseX}{41}
\newcommand{\revThreeMolmoParseY}{43}

\newcommand{\sixNPairs}{300}
\newcommand{\sixNWriters}{58}

\newcommand{\sixPassedA}{10}

\newcommand{\sixRatioLoMin}{2.31}
\newcommand{\sixRatioLoMax}{7.78}
\newcommand{\sixExcessMin}{16.6}
\newcommand{\sixExcessMax}{44.8}
\newcommand{\sixDigitsAbove}{9}
\newcommand{\sixDigitsRatioMin}{4.8}
\newcommand{\sixDigitsRatioMax}{43.0}

\newcommand{\sixNStarts}{600}
\newcommand{\sixDeltaOther}{548}

\newcommand{\mchPatchQwenT}{0.02}

\newcommand{\mchPatchQwenO}{\ensuremath{-}0.02}

\newcommand{\mchPatchQwenP}{0.96}

\newcommand{\mchPairsQwen}{22}

\newcommand{\mchPOneQwen}{0.44}
\newcommand{\mchPOneQwenLo}{0.25}
\newcommand{\mchPOneQwenHi}{0.63}
\newcommand{\mchNullQwen}{0.62}

\newcommand{\mchProbeXQwen}{100.0}
\newcommand{\mchProbeYQwen}{99.3}

\newcommand{\phYdQwen}{80.3}

\newcommand{\mchPatchQwenVLT}{0.01}

\newcommand{\mchPatchQwenVLO}{0.00}

\newcommand{\mchPatchQwenVLP}{0.99}

\newcommand{\mchPairsQwenVL}{91}

\newcommand{\mchPOneQwenVL}{0.86}
\newcommand{\mchPOneQwenVLLo}{0.78}
\newcommand{\mchPOneQwenVLHi}{0.92}
\newcommand{\mchNullQwenVL}{0.79}

\newcommand{\mchProbeXQwenVL}{99.2}
\newcommand{\mchProbeYQwenVL}{99.2}

\newcommand{\phYdQwenVL}{57.0}

\newcommand{\mchPatchQwenFourT}{0.00}

\newcommand{\mchPatchQwenFourO}{0.03}

\newcommand{\mchPatchQwenFourP}{1.00}

\newcommand{\mchPairsQwenFour}{71}

\newcommand{\mchPOneQwenFour}{0.75}
\newcommand{\mchPOneQwenFourLo}{0.64}
\newcommand{\mchPOneQwenFourHi}{0.84}
\newcommand{\mchNullQwenFour}{0.91}

\newcommand{\mchPOnePassed}{2}

\newcommand{\tostMargin}{0.10}
\newcommand{\tostQwenTLo}{\ensuremath{-}0.070}
\newcommand{\tostQwenTHi}{0.096}

\newcommand{\tostQwenVLTLo}{\ensuremath{-}0.003}
\newcommand{\tostQwenVLTHi}{0.019}

\newcommand{\pfxDQwen}{0.95}

\newcommand{\pfxFQwen}{0.20}

\newcommand{\pfxDopQwen}{0.94}
\newcommand{\pfxDopQwenLo}{0.86}
\newcommand{\pfxDopQwenHi}{0.99}
\newcommand{\pfxDresQwen}{0.04}
\newcommand{\pfxDresQwenLo}{\ensuremath{-}0.01}
\newcommand{\pfxDresQwenHi}{0.11}

\newcommand{\pfxPairsQwen}{21}

\newcommand{\pfxDQwenVL}{0.78}

\newcommand{\pfxFQwenVL}{0.94}

\newcommand{\pfxDopQwenVL}{0.66}
\newcommand{\pfxDopQwenVLLo}{0.58}
\newcommand{\pfxDopQwenVLHi}{0.74}
\newcommand{\pfxDresQwenVL}{0.78}
\newcommand{\pfxDresQwenVLLo}{0.66}
\newcommand{\pfxDresQwenVLHi}{0.90}

\newcommand{\pfxPairsQwenVL}{91}

\newcommand{\pfxPassed}{1}
\newcommand{\qaDissNames}{Gemma-4-12B, InternVL3.5-38B, Mistral-Small-3.2, Molmo2-O-7B}
\newcommand{\qaClearNames}{Gemma-4-12B, Qwen3.8-27B, Qwen3.5-9B, InternVL3.5-38B, Mistral-Small-3.2, Molmo2-O-7B}
\newcommand{\qaClearTested}{8}
\newcommand{\qaDissTested}{8}
\newcommand{\qaLeakGemma}{2.9}
\newcommand{\qaLeakGemmaLo}{1.2}
\newcommand{\qaLeakGemmaHi}{86.2}
\newcommand{\qaLeakOGemma}{6}
\newcommand{\qaLeakErrGemma}{7}

\newcommand{\qaKnowGemma}{2.5}
\newcommand{\qaKnowGemmaLo}{1.2}
\newcommand{\qaKnowGemmaHi}{88.0}
\newcommand{\qaKnowOGemma}{5}
\newcommand{\qaKnowErrGemma}{6}

\newcommand{\qaClearExGemma}{\ensuremath{-}1.4}
\newcommand{\qaClearExGemmaLo}{\ensuremath{-}1.9}
\newcommand{\qaClearExGemmaHi}{\ensuremath{-}0.5}
\newcommand{\qaQOneCompGemma}{1.3}

\newcommand{\qaQTwoCompGemma}{89.3}
\newcommand{\qaQFourCompGemma}{2.0}
\newcommand{\qaQThreeDpGemma}{3.64}
\newcommand{\qaQThreeCGemma}{0.41}
\newcommand{\qaQThreeHitGemma}{0.92}
\newcommand{\qaQThreeFaGemma}{0.01}
\newcommand{\qaTextDpGemma}{5.92}

\newcommand{\qaMaskGemma}{0.0}
\newcommand{\qaTextGemma}{0.0}

\newcommand{\qaLeakGemmaTwelve}{6.9}
\newcommand{\qaLeakGemmaTwelveLo}{2.7}
\newcommand{\qaLeakGemmaTwelveHi}{16.2}
\newcommand{\qaLeakOGemmaTwelve}{37}
\newcommand{\qaLeakErrGemmaTwelve}{41}

\newcommand{\qaKnowGemmaTwelve}{8.7}
\newcommand{\qaKnowGemmaTwelveLo}{3.1}
\newcommand{\qaKnowGemmaTwelveHi}{92.1}
\newcommand{\qaKnowOGemmaTwelve}{21}
\newcommand{\qaKnowErrGemmaTwelve}{22}

\newcommand{\qaClearExGemmaTwelve}{70.9}
\newcommand{\qaClearExGemmaTwelveLo}{38.0}
\newcommand{\qaClearExGemmaTwelveHi}{90.2}
\newcommand{\qaQOneCompGemmaTwelve}{8.2}

\newcommand{\qaQTwoCompGemmaTwelve}{85.3}
\newcommand{\qaQFourCompGemmaTwelve}{0.2}
\newcommand{\qaQThreeDpGemmaTwelve}{1.50}
\newcommand{\qaQThreeCGemmaTwelve}{0.16}
\newcommand{\qaQThreeHitGemmaTwelve}{0.72}
\newcommand{\qaQThreeFaGemmaTwelve}{0.18}
\newcommand{\qaTextDpGemmaTwelve}{2.68}

\newcommand{\qaMaskGemmaTwelve}{2.3}
\newcommand{\qaTextGemmaTwelve}{0.0}

\newcommand{\qaLeakQwen}{8.6}
\newcommand{\qaLeakQwenLo}{1.1}
\newcommand{\qaLeakQwenHi}{95.4}
\newcommand{\qaLeakOQwen}{9}
\newcommand{\qaLeakErrQwen}{12}

\newcommand{\qaKnowQwen}{9.2}
\newcommand{\qaKnowQwenLo}{1.5}
\newcommand{\qaKnowQwenHi}{99.4}
\newcommand{\qaKnowOQwen}{5}
\newcommand{\qaKnowErrQwen}{7}

\newcommand{\qaClearExQwen}{91.3}
\newcommand{\qaClearExQwenLo}{62.1}
\newcommand{\qaClearExQwenHi}{99.1}
\newcommand{\qaQOneCompQwen}{2.0}

\newcommand{\qaQTwoCompQwen}{82.4}
\newcommand{\qaQFourCompQwen}{2.2}
\newcommand{\qaQThreeDpQwen}{1.44}
\newcommand{\qaQThreeCQwen}{0.84}
\newcommand{\qaQThreeHitQwen}{0.45}
\newcommand{\qaQThreeFaQwen}{0.06}
\newcommand{\qaTextDpQwen}{2.95}

\newcommand{\qaMaskQwen}{54.2}
\newcommand{\qaTextQwen}{19.3}

\newcommand{\qaLeakQwenVL}{4.0}
\newcommand{\qaLeakQwenVLLo}{1.1}
\newcommand{\qaLeakQwenVLHi}{27.0}
\newcommand{\qaLeakOQwenVL}{11}
\newcommand{\qaLeakErrQwenVL}{29}

\newcommand{\qaKnowQwenVL}{6.5}
\newcommand{\qaKnowQwenVLLo}{0.0}
\newcommand{\qaKnowQwenVLHi}{275.1}
\newcommand{\qaKnowOQwenVL}{3}
\newcommand{\qaKnowErrQwenVL}{5}

\newcommand{\qaClearExQwenVL}{94.9}
\newcommand{\qaClearExQwenVLLo}{81.2}
\newcommand{\qaClearExQwenVLHi}{99.7}
\newcommand{\qaQOneCompQwenVL}{2.4}

\newcommand{\qaQTwoCompQwenVL}{81.3}
\newcommand{\qaQFourCompQwenVL}{2.9}
\newcommand{\qaQThreeDpQwenVL}{1.84}
\newcommand{\qaQThreeCQwenVL}{0.39}
\newcommand{\qaQThreeHitQwenVL}{0.70}
\newcommand{\qaQThreeFaQwenVL}{0.10}
\newcommand{\qaTextDpQwenVL}{2.49}

\newcommand{\qaMaskQwenVL}{63.6}
\newcommand{\qaTextQwenVL}{1.2}

\newcommand{\qaLeakQwenFour}{0.0}
\newcommand{\qaLeakQwenFourLo}{0.0}
\newcommand{\qaLeakQwenFourHi}{0.0}
\newcommand{\qaLeakOQwenFour}{0}
\newcommand{\qaLeakErrQwenFour}{3}

\newcommand{\qaKnowOQwenFour}{0}
\newcommand{\qaKnowErrQwenFour}{0}

\newcommand{\qaQOneCompQwenFour}{0.0}

\newcommand{\qaQTwoCompQwenFour}{79.6}
\newcommand{\qaQFourCompQwenFour}{4.0}
\newcommand{\qaQThreeDpQwenFour}{1.44}
\newcommand{\qaQThreeCQwenFour}{\ensuremath{-}0.82}
\newcommand{\qaQThreeHitQwenFour}{0.94}
\newcommand{\qaQThreeFaQwenFour}{0.54}
\newcommand{\qaTextDpQwenFour}{1.95}

\newcommand{\qaMaskQwenFour}{57.4}
\newcommand{\qaTextQwenFour}{5.9}

\newcommand{\qaLeakIntern}{9.0}
\newcommand{\qaLeakInternLo}{3.4}
\newcommand{\qaLeakInternHi}{59.6}
\newcommand{\qaLeakOIntern}{19}
\newcommand{\qaLeakErrIntern}{22}

\newcommand{\qaKnowIntern}{8.7}
\newcommand{\qaKnowInternLo}{3.3}
\newcommand{\qaKnowInternHi}{59.4}
\newcommand{\qaKnowOIntern}{18}
\newcommand{\qaKnowErrIntern}{19}

\newcommand{\qaClearExIntern}{76.1}
\newcommand{\qaClearExInternLo}{57.0}
\newcommand{\qaClearExInternHi}{93.5}
\newcommand{\qaQOneCompIntern}{4.2}

\newcommand{\qaQTwoCompIntern}{72.9}
\newcommand{\qaQFourCompIntern}{1.1}
\newcommand{\qaQThreeDpIntern}{1.14}
\newcommand{\qaQThreeCIntern}{1.66}
\newcommand{\qaQThreeHitIntern}{0.14}
\newcommand{\qaQThreeFaIntern}{0.01}
\newcommand{\qaTextDpIntern}{4.93}

\newcommand{\qaMaskIntern}{7.7}
\newcommand{\qaTextIntern}{1.2}

\newcommand{\qaLeakMistral}{11.0}
\newcommand{\qaLeakMistralLo}{4.1}
\newcommand{\qaLeakMistralHi}{38.2}
\newcommand{\qaLeakOMistral}{64}
\newcommand{\qaLeakErrMistral}{95}

\newcommand{\qaKnowMistral}{21.0}
\newcommand{\qaKnowMistralLo}{7.7}
\newcommand{\qaKnowMistralHi}{75.3}
\newcommand{\qaKnowOMistral}{45}
\newcommand{\qaKnowErrMistral}{50}

\newcommand{\qaClearExMistral}{71.0}
\newcommand{\qaClearExMistralLo}{47.8}
\newcommand{\qaClearExMistralHi}{91.2}
\newcommand{\qaQOneCompMistral}{14.2}

\newcommand{\qaQTwoCompMistral}{41.8}
\newcommand{\qaQFourCompMistral}{19.6}
\newcommand{\qaQThreeDpMistral}{0.11}
\newcommand{\qaQThreeCMistral}{0.88}
\newcommand{\qaQThreeHitMistral}{0.20}
\newcommand{\qaQThreeFaMistral}{0.17}
\newcommand{\qaTextDpMistral}{1.42}

\newcommand{\qaMaskMistral}{40.0}
\newcommand{\qaTextMistral}{0.1}

\newcommand{\qaLeakGlm}{8.6}
\newcommand{\qaLeakGlmLo}{0.0}
\newcommand{\qaLeakGlmHi}{55.6}
\newcommand{\qaLeakOGlm}{5}
\newcommand{\qaLeakErrGlm}{17}

\newcommand{\qaKnowOGlm}{0}
\newcommand{\qaKnowErrGlm}{0}

\newcommand{\qaQOneCompGlm}{1.1}

\newcommand{\qaQTwoCompGlm}{33.8}
\newcommand{\qaQFourCompGlm}{1.3}

\newcommand{\qaTextDpGlm}{1.94}

\newcommand{\qaMaskGlm}{68.4}
\newcommand{\qaTextGlm}{0.3}

\newcommand{\qaLeakLlava}{3.0}
\newcommand{\qaLeakLlavaLo}{0.0}
\newcommand{\qaLeakLlavaHi}{14.4}
\newcommand{\qaLeakOLlava}{3}
\newcommand{\qaLeakErrLlava}{19}

\newcommand{\qaKnowLlava}{6.7}
\newcommand{\qaKnowLlavaLo}{0.0}
\newcommand{\qaKnowLlavaHi}{266.3}
\newcommand{\qaKnowOLlava}{2}
\newcommand{\qaKnowErrLlava}{4}

\newcommand{\qaClearExLlava}{36.6}
\newcommand{\qaClearExLlavaLo}{\ensuremath{-}1.6}
\newcommand{\qaClearExLlavaHi}{99.8}
\newcommand{\qaQOneCompLlava}{0.7}

\newcommand{\qaQTwoCompLlava}{79.3}
\newcommand{\qaQFourCompLlava}{0.0}
\newcommand{\qaQThreeDpLlava}{1.12}
\newcommand{\qaQThreeCLlava}{\ensuremath{-}0.31}
\newcommand{\qaQThreeHitLlava}{0.81}
\newcommand{\qaQThreeFaLlava}{0.40}
\newcommand{\qaTextDpLlava}{2.95}

\newcommand{\qaMaskLlava}{70.9}
\newcommand{\qaTextLlava}{0.0}

\newcommand{\qaLeakMolmo}{6.3}
\newcommand{\qaLeakMolmoLo}{2.0}
\newcommand{\qaLeakMolmoHi}{21.5}
\newcommand{\qaLeakOMolmo}{12}
\newcommand{\qaLeakErrMolmo}{59}

\newcommand{\qaKnowMolmo}{7.1}
\newcommand{\qaKnowMolmoLo}{2.0}
\newcommand{\qaKnowMolmoHi}{29.4}
\newcommand{\qaKnowOMolmo}{11}
\newcommand{\qaKnowErrMolmo}{36}

\newcommand{\qaClearExMolmo}{16.7}
\newcommand{\qaClearExMolmoLo}{2.6}
\newcommand{\qaClearExMolmoHi}{49.1}
\newcommand{\qaQOneCompMolmo}{2.7}

\newcommand{\qaQTwoCompMolmo}{72.0}
\newcommand{\qaQFourCompMolmo}{1.1}
\newcommand{\qaQThreeDpMolmo}{0.14}
\newcommand{\qaQThreeCMolmo}{0.94}
\newcommand{\qaQThreeHitMolmo}{0.19}
\newcommand{\qaQThreeFaMolmo}{0.16}
\newcommand{\qaTextDpMolmo}{1.50}

\newcommand{\qaMaskMolmo}{47.0}
\newcommand{\qaTextMolmo}{29.6}

\newcommand{\qaKnowRatioMin}{7.1}
\newcommand{\qaKnowRatioMax}{21.0}

\newcommand{\qaTextMin}{0.0}
\newcommand{\qaTextMax}{29.6}

\newcommand{\qaMaskMin}{0.0}
\newcommand{\qaMaskMax}{70.9}
\newcommand{\qaDissPassed}{4}
\newcommand{\qaClearPassed}{6}

\newcommand{\qaKnowQTwoGemma}{2.9}
\newcommand{\qaKnowQTwoGemmaLo}{1.2}
\newcommand{\qaKnowQTwoGemmaHi}{88.0}

\newcommand{\qaKnowQTwoGemmaTwelve}{7.9}
\newcommand{\qaKnowQTwoGemmaTwelveLo}{3.1}
\newcommand{\qaKnowQTwoGemmaTwelveHi}{19.4}

\newcommand{\qaKnowQTwoQwen}{8.5}
\newcommand{\qaKnowQTwoQwenLo}{1.1}
\newcommand{\qaKnowQTwoQwenHi}{95.5}

\newcommand{\qaKnowQTwoQwenVL}{5.5}
\newcommand{\qaKnowQTwoQwenVLLo}{1.9}
\newcommand{\qaKnowQTwoQwenVLHi}{29.9}

\newcommand{\qaKnowQTwoQwenFour}{0.0}
\newcommand{\qaKnowQTwoQwenFourLo}{0.0}
\newcommand{\qaKnowQTwoQwenFourHi}{0.0}

\newcommand{\qaKnowQTwoIntern}{8.7}
\newcommand{\qaKnowQTwoInternLo}{3.3}
\newcommand{\qaKnowQTwoInternHi}{59.4}
\newcommand{\qaQThreeNoTrueIntern}{86}
\newcommand{\qaKnowQTwoMistral}{19.9}
\newcommand{\qaKnowQTwoMistralLo}{7.2}
\newcommand{\qaKnowQTwoMistralHi}{76.3}

\newcommand{\qaKnowQTwoGlm}{15.7}
\newcommand{\qaKnowQTwoGlmLo}{0.0}
\newcommand{\qaKnowQTwoGlmHi}{87.0}

\newcommand{\qaKnowQTwoLlava}{4.4}
\newcommand{\qaKnowQTwoLlavaLo}{0.0}
\newcommand{\qaKnowQTwoLlavaHi}{54.0}

\newcommand{\qaKnowQTwoMolmo}{7.1}
\newcommand{\qaKnowQTwoMolmoLo}{2.0}
\newcommand{\qaKnowQTwoMolmoHi}{29.3}

\newcommand{\colGemma}{8.4}
\newcommand{\colGemmaLo}{3.3}
\newcommand{\colGemmaHi}{19.4}
\newcommand{\colClearExGemma}{37.5}
\newcommand{\colClearExGemmaLo}{18.6}
\newcommand{\colClearExGemmaHi}{56.3}
\newcommand{\colRepGemma}{5.1}
\newcommand{\ineqTruthGemma}{1.61}
\newcommand{\ineqTruthGemmaLo}{0.00}
\newcommand{\ineqTruthGemmaHi}{3.61}
\newcommand{\ineqRepGemma}{0.3}
\newcommand{\colGemmaTwelve}{9.5}
\newcommand{\colGemmaTwelveLo}{4.3}
\newcommand{\colGemmaTwelveHi}{15.2}
\newcommand{\colClearExGemmaTwelve}{39.8}
\newcommand{\colClearExGemmaTwelveLo}{24.3}
\newcommand{\colClearExGemmaTwelveHi}{52.8}
\newcommand{\colRepGemmaTwelve}{10.7}
\newcommand{\ineqTruthGemmaTwelve}{0.57}
\newcommand{\ineqTruthGemmaTwelveLo}{0.00}
\newcommand{\ineqTruthGemmaTwelveHi}{0.96}
\newcommand{\ineqRepGemmaTwelve}{0.0}
\newcommand{\colQwen}{8.1}
\newcommand{\colQwenLo}{4.2}
\newcommand{\colQwenHi}{15.7}
\newcommand{\colClearExQwen}{11.9}
\newcommand{\colClearExQwenLo}{3.6}
\newcommand{\colClearExQwenHi}{21.5}
\newcommand{\colRepQwen}{4.2}
\newcommand{\ineqTruthQwen}{1.23}
\newcommand{\ineqTruthQwenLo}{0.84}
\newcommand{\ineqTruthQwenHi}{1.70}
\newcommand{\ineqRepQwen}{0.9}
\newcommand{\colQwenVL}{8.6}
\newcommand{\colQwenVLLo}{4.3}
\newcommand{\colQwenVLHi}{14.6}
\newcommand{\colClearExQwenVL}{16.8}
\newcommand{\colClearExQwenVLLo}{7.7}
\newcommand{\colClearExQwenVLHi}{26.8}
\newcommand{\colRepQwenVL}{4.0}
\newcommand{\ineqTruthQwenVL}{1.62}
\newcommand{\ineqTruthQwenVLLo}{1.09}
\newcommand{\ineqTruthQwenVLHi}{3.05}
\newcommand{\ineqRepQwenVL}{1.2}
\newcommand{\colQwenFour}{3.1}
\newcommand{\colQwenFourLo}{0.4}
\newcommand{\colQwenFourHi}{8.6}
\newcommand{\colClearExQwenFour}{1.5}
\newcommand{\colClearExQwenFourLo}{\ensuremath{-}3.9}
\newcommand{\colClearExQwenFourHi}{11.0}
\newcommand{\colRepQwenFour}{1.1}
\newcommand{\ineqTruthQwenFour}{1.32}
\newcommand{\ineqTruthQwenFourLo}{0.92}
\newcommand{\ineqTruthQwenFourHi}{1.79}
\newcommand{\ineqRepQwenFour}{0.5}
\newcommand{\colIntern}{16.7}
\newcommand{\colInternLo}{9.9}
\newcommand{\colInternHi}{23.5}
\newcommand{\colClearExIntern}{26.5}
\newcommand{\colClearExInternLo}{17.9}
\newcommand{\colClearExInternHi}{36.2}
\newcommand{\colRepIntern}{8.9}
\newcommand{\ineqTruthIntern}{1.19}
\newcommand{\ineqTruthInternLo}{0.84}
\newcommand{\ineqTruthInternHi}{1.70}
\newcommand{\ineqRepIntern}{0.2}
\newcommand{\colMistral}{21.8}
\newcommand{\colMistralLo}{12.8}
\newcommand{\colMistralHi}{29.8}
\newcommand{\colClearExMistral}{21.6}
\newcommand{\colClearExMistralLo}{12.1}
\newcommand{\colClearExMistralHi}{31.4}
\newcommand{\colRepMistral}{9.9}
\newcommand{\ineqTruthMistral}{1.30}
\newcommand{\ineqTruthMistralLo}{0.75}
\newcommand{\ineqTruthMistralHi}{1.83}
\newcommand{\ineqRepMistral}{0.0}
\newcommand{\colGlm}{0.0}
\newcommand{\colGlmLo}{0.0}
\newcommand{\colGlmHi}{0.0}

\newcommand{\colRepGlm}{0.0}

\newcommand{\ineqRepGlm}{0.0}
\newcommand{\colLlava}{10.0}
\newcommand{\colLlavaLo}{4.4}
\newcommand{\colLlavaHi}{19.8}
\newcommand{\colClearExLlava}{7.6}
\newcommand{\colClearExLlavaLo}{1.3}
\newcommand{\colClearExLlavaHi}{15.3}
\newcommand{\colRepLlava}{2.4}
\newcommand{\ineqTruthLlava}{0.83}
\newcommand{\ineqTruthLlavaLo}{0.32}
\newcommand{\ineqTruthLlavaHi}{1.59}
\newcommand{\ineqRepLlava}{0.2}
\newcommand{\colMolmo}{3.8}
\newcommand{\colMolmoLo}{0.0}
\newcommand{\colMolmoHi}{9.0}
\newcommand{\colClearExMolmo}{\ensuremath{-}1.4}
\newcommand{\colClearExMolmoLo}{\ensuremath{-}2.6}
\newcommand{\colClearExMolmoHi}{\ensuremath{-}0.5}
\newcommand{\colRepMolmo}{0.6}
\newcommand{\ineqTruthMolmo}{1.56}
\newcommand{\ineqTruthMolmoLo}{0.00}
\newcommand{\ineqTruthMolmoHi}{2.53}
\newcommand{\ineqRepMolmo}{0.0}

\newcommand{\ineqTruthMin}{0.57}
\newcommand{\ineqTruthMax}{1.62}

\newcommand{\colPassed}{7}
\newcommand{\colClearPassed}{6}
\newcommand{\colPairs}{240}
\newcommand{\ineqPairs}{240}

\newcommand{\sclPassed}{2}
\newcommand{\sclClearPassed}{2}

\newcommand{\bidFreeOQwen}{0.73}
\newcommand{\bidFreeOQwenLo}{0.54}
\newcommand{\bidFreeOQwenHi}{0.91}
\newcommand{\bidFreeTQwen}{0.08}

\newcommand{\bidRevOQwen}{0.49}
\newcommand{\bidRevOQwenLo}{0.23}
\newcommand{\bidRevOQwenHi}{0.78}
\newcommand{\bidPairsQwen}{22}
\newcommand{\bidWritersQwen}{20}

\newcommand{\bidFreeOQwenVL}{0.07}
\newcommand{\bidFreeOQwenVLLo}{0.01}
\newcommand{\bidFreeOQwenVLHi}{0.12}
\newcommand{\bidFreeTQwenVL}{0.01}

\newcommand{\bidRevOQwenVL}{0.18}
\newcommand{\bidRevOQwenVLLo}{0.09}
\newcommand{\bidRevOQwenVLHi}{0.38}
\newcommand{\bidPairsQwenVL}{91}
\newcommand{\bidWritersQwenVL}{59}

\newcommand{\bidHOnePassed}{2}

\newcommand{\sensDecisions}{176}

\newcommand{\qaNAnswers}{450}
\newcommand{\qaNFalse}{150}

\newcommand{\bidHolm}{0.0029}

\newcommand{\qOneNullPassed}{5}
\newcommand{\qOneNullNames}{Gemma-4-31B, Gemma-4-12B, Qwen3.8-27B, InternVL3.5-38B, Mistral-Small-3.2}
\newcommand{\qOneNullRatioMin}{7.4}
\newcommand{\qOneNullRatioMax}{18.3}
\newcommand{\qOneNullGemma}{17.8}
\newcommand{\qOneNullGemmaLo}{12.9}
\newcommand{\qOneNullGemmaHi}{37.9}
\newcommand{\qOneNullGemmaTwelve}{18.3}
\newcommand{\qOneNullGemmaTwelveLo}{11.8}
\newcommand{\qOneNullGemmaTwelveHi}{25.7}
\newcommand{\qOneNullQwen}{11.8}
\newcommand{\qOneNullQwenLo}{4.3}
\newcommand{\qOneNullQwenHi}{51.9}
\newcommand{\qOneNullQwenVL}{10.3}
\newcommand{\qOneNullQwenVLLo}{0.0}
\newcommand{\qOneNullQwenVLHi}{88.1}

\newcommand{\qOneNullIntern}{16.5}
\newcommand{\qOneNullInternLo}{9.5}
\newcommand{\qOneNullInternHi}{23.7}
\newcommand{\qOneNullMistral}{7.4}
\newcommand{\qOneNullMistralLo}{3.5}
\newcommand{\qOneNullMistralHi}{38.0}

\newcommand{\qOneNullLlava}{8.5}
\newcommand{\qOneNullLlavaLo}{0.0}
\newcommand{\qOneNullLlavaHi}{13.1}
\newcommand{\qOneNullMolmo}{2.8}
\newcommand{\qOneNullMolmoLo}{0.8}
\newcommand{\qOneNullMolmoHi}{18.0}

\newcommand{\qaSelPassed}{5}

\newcommand{\qaRegRuleMinimum}{3}
\newcommand{\qaRegRuleOf}{5}
\newcommand{\qaRegRulePassed}{2}
\newcommand{\qaRegRuleNames}{InternVL3.5-38B, Mistral-Small-3.2}
\newcommand{\qaRegRuleMet}{not met}

\newcommand{\qaLowoMinGemmaTwelve}{6.8}
\newcommand{\qaLowoMaxGemmaTwelve}{14.3}
\newcommand{\qaLowoWritersGemmaTwelve}{8}
\newcommand{\qaLowoShareGemmaTwelve}{24}

\newcommand{\qaLowoMinIntern}{7.7}
\newcommand{\qaLowoMaxIntern}{21.5}
\newcommand{\qaLowoWritersIntern}{13}
\newcommand{\qaLowoShareIntern}{11}

\newcommand{\qaLowoMinMistral}{17.1}
\newcommand{\qaLowoMaxMistral}{32.9}
\newcommand{\qaLowoWritersMistral}{20}
\newcommand{\qaLowoShareMistral}{13}

\newcommand{\qaLowoMinMolmo}{4.4}
\newcommand{\qaLowoMaxMolmo}{11.0}
\newcommand{\qaLowoWritersMolmo}{8}
\newcommand{\qaLowoShareMolmo}{27}

\newcommand{\qaCopyOGemma}{0}
\newcommand{\qaCopyErrGemma}{0}

\newcommand{\qaCopyOGemmaTwelve}{0}
\newcommand{\qaCopyErrGemmaTwelve}{0}

\newcommand{\qaCopyRateGemmaTwelve}{0.0}
\newcommand{\qaCopyRateHiGemmaTwelve}{0.8}

\newcommand{\qaCopyOQwen}{2}
\newcommand{\qaCopyErrQwen}{3}

\newcommand{\qaCopyOQwenVL}{0}
\newcommand{\qaCopyErrQwenVL}{3}

\newcommand{\qaCopyOQwenFour}{0}
\newcommand{\qaCopyErrQwenFour}{0}

\newcommand{\qaCopyOIntern}{0}
\newcommand{\qaCopyErrIntern}{3}

\newcommand{\qaCopyRateHiIntern}{0.8}

\newcommand{\qaCopyOMistral}{6}
\newcommand{\qaCopyErrMistral}{21}

\newcommand{\qaCopyOGlm}{0}
\newcommand{\qaCopyErrGlm}{0}

\newcommand{\qaCopyOLlava}{0}
\newcommand{\qaCopyErrLlava}{6}

\newcommand{\qaCopyOMolmo}{1}
\newcommand{\qaCopyErrMolmo}{12}

\newcommand{\qaHeldPairs}{300}
\newcommand{\qaHeldWriters}{105}
\newcommand{\qaHeldFamily}{4}
\newcommand{\qaHeldReplicated}{2}

\newcommand{\qaHeldCopyAnyMet}{0}

\newcommand{\qaHeldNullGemma}{6.4}
\newcommand{\qaHeldNullGemmaLo}{2.4}
\newcommand{\qaHeldNullGemmaHi}{30.2}

\newcommand{\qaHeldNullOGemma}{11}

\newcommand{\qaHeldNullErrorsGemma}{11}
\newcommand{\qaHeldNullGemmaTwelve}{11.9}
\newcommand{\qaHeldNullGemmaTwelveLo}{6.2}
\newcommand{\qaHeldNullGemmaTwelveHi}{26.8}

\newcommand{\qaHeldNullOGemmaTwelve}{36}

\newcommand{\qaHeldNullErrorsGemmaTwelve}{49}
\newcommand{\qaHeldNullQwen}{4.4}
\newcommand{\qaHeldNullQwenLo}{1.6}
\newcommand{\qaHeldNullQwenHi}{19.9}

\newcommand{\qaHeldNullOQwen}{16}

\newcommand{\qaHeldNullErrorsQwen}{46}
\newcommand{\qaHeldNullQwenVL}{4.2}
\newcommand{\qaHeldNullQwenVLLo}{1.6}
\newcommand{\qaHeldNullQwenVLHi}{11.6}

\newcommand{\qaHeldNullOQwenVL}{20}

\newcommand{\qaHeldNullErrorsQwenVL}{33}
\newcommand{\qaHeldNullQwenFour}{1.8}
\newcommand{\qaHeldNullQwenFourLo}{0.0}
\newcommand{\qaHeldNullQwenFourHi}{14.8}

\newcommand{\qaHeldNullOQwenFour}{3}

\newcommand{\qaHeldNullErrorsQwenFour}{24}
\newcommand{\qaHeldNullIntern}{12.7}
\newcommand{\qaHeldNullInternLo}{5.3}
\newcommand{\qaHeldNullInternHi}{28.4}

\newcommand{\qaHeldNullOIntern}{33}

\newcommand{\qaHeldNullErrorsIntern}{51}
\newcommand{\qaHeldNullMistral}{5.3}
\newcommand{\qaHeldNullMistralLo}{2.7}
\newcommand{\qaHeldNullMistralHi}{10.4}

\newcommand{\qaHeldNullOMistral}{51}

\newcommand{\qaHeldNullErrorsMistral}{94}
\newcommand{\qaHeldNullGlm}{0.4}
\newcommand{\qaHeldNullGlmLo}{0.0}
\newcommand{\qaHeldNullGlmHi}{4.8}

\newcommand{\qaHeldNullOGlm}{1}

\newcommand{\qaHeldNullErrorsGlm}{23}
\newcommand{\qaHeldNullLlava}{4.8}
\newcommand{\qaHeldNullLlavaLo}{1.4}
\newcommand{\qaHeldNullLlavaHi}{20.7}

\newcommand{\qaHeldNullOLlava}{10}

\newcommand{\qaHeldNullErrorsLlava}{27}
\newcommand{\qaHeldNullMolmo}{5.1}
\newcommand{\qaHeldNullMolmoLo}{2.2}
\newcommand{\qaHeldNullMolmoHi}{19.0}

\newcommand{\qaHeldNullOMolmo}{23}

\newcommand{\qaHeldNullErrorsMolmo}{49}
\newcommand{\qaHeldKnowPassed}{6}

\newcommand{\qaHeldKnowGemma}{20.9}
\newcommand{\qaHeldKnowGemmaLo}{3.7}
\newcommand{\qaHeldKnowGemmaHi}{151.4}

\newcommand{\qaHeldKnowOGemma}{6}

\newcommand{\qaHeldKnowErrorsGemma}{6}
\newcommand{\qaHeldKnowGemmaTwelve}{13.9}
\newcommand{\qaHeldKnowGemmaTwelveLo}{4.6}
\newcommand{\qaHeldKnowGemmaTwelveHi}{64.7}

\newcommand{\qaHeldKnowOGemmaTwelve}{23}

\newcommand{\qaHeldKnowErrorsGemmaTwelve}{33}
\newcommand{\qaHeldKnowQwen}{6.0}
\newcommand{\qaHeldKnowQwenLo}{2.0}
\newcommand{\qaHeldKnowQwenHi}{24.5}

\newcommand{\qaHeldKnowOQwen}{13}

\newcommand{\qaHeldKnowErrorsQwen}{28}
\newcommand{\qaHeldKnowQwenVL}{4.8}
\newcommand{\qaHeldKnowQwenVLLo}{0.9}
\newcommand{\qaHeldKnowQwenVLHi}{24.4}

\newcommand{\qaHeldKnowOQwenVL}{12}

\newcommand{\qaHeldKnowErrorsQwenVL}{18}
\newcommand{\qaHeldKnowQwenFour}{0.0}
\newcommand{\qaHeldKnowQwenFourLo}{0.0}
\newcommand{\qaHeldKnowQwenFourHi}{0.0}

\newcommand{\qaHeldKnowOQwenFour}{0}

\newcommand{\qaHeldKnowErrorsQwenFour}{3}
\newcommand{\qaHeldKnowIntern}{22.0}
\newcommand{\qaHeldKnowInternLo}{9.4}
\newcommand{\qaHeldKnowInternHi}{65.7}

\newcommand{\qaHeldKnowOIntern}{28}

\newcommand{\qaHeldKnowErrorsIntern}{39}
\newcommand{\qaHeldKnowMistral}{20.4}
\newcommand{\qaHeldKnowMistralLo}{9.9}
\newcommand{\qaHeldKnowMistralHi}{55.4}

\newcommand{\qaHeldKnowOMistral}{45}

\newcommand{\qaHeldKnowErrorsMistral}{51}
\newcommand{\qaHeldKnowGlm}{\blank{}}
\newcommand{\qaHeldKnowGlmLo}{\blank{}}
\newcommand{\qaHeldKnowGlmHi}{\blank{}}

\newcommand{\qaHeldKnowOGlm}{0}

\newcommand{\qaHeldKnowErrorsGlm}{0}
\newcommand{\qaHeldKnowLlava}{2.7}
\newcommand{\qaHeldKnowLlavaLo}{0.0}
\newcommand{\qaHeldKnowLlavaHi}{52.4}

\newcommand{\qaHeldKnowOLlava}{3}

\newcommand{\qaHeldKnowErrorsLlava}{11}
\newcommand{\qaHeldKnowMolmo}{22.8}
\newcommand{\qaHeldKnowMolmoLo}{7.9}
\newcommand{\qaHeldKnowMolmoHi}{72.7}

\newcommand{\qaHeldKnowOMolmo}{16}

\newcommand{\qaHeldKnowErrorsMolmo}{34}

\newcommand{\qaHeldCopyGemma}{4.3}
\newcommand{\qaHeldCopyGemmaLo}{0.0}
\newcommand{\qaHeldCopyGemmaHi}{30.2}

\newcommand{\qaHeldCopyOGemma}{7}

\newcommand{\qaHeldCopyErrorsGemma}{10}
\newcommand{\qaHeldCopyGemmaTwelve}{2.3}
\newcommand{\qaHeldCopyGemmaTwelveLo}{0.0}
\newcommand{\qaHeldCopyGemmaTwelveHi}{10.6}

\newcommand{\qaHeldCopyOGemmaTwelve}{2}

\newcommand{\qaHeldCopyErrorsGemmaTwelve}{17}
\newcommand{\qaHeldCopyQwen}{2.8}
\newcommand{\qaHeldCopyQwenLo}{0.0}
\newcommand{\qaHeldCopyQwenHi}{9.2}

\newcommand{\qaHeldCopyOQwen}{5}

\newcommand{\qaHeldCopyErrorsQwen}{38}
\newcommand{\qaHeldCopyQwenVL}{3.7}
\newcommand{\qaHeldCopyQwenVLLo}{1.3}
\newcommand{\qaHeldCopyQwenVLHi}{12.5}

\newcommand{\qaHeldCopyOQwenVL}{10}

\newcommand{\qaHeldCopyErrorsQwenVL}{21}
\newcommand{\qaHeldCopyQwenFour}{2.4}
\newcommand{\qaHeldCopyQwenFourLo}{0.0}
\newcommand{\qaHeldCopyQwenFourHi}{18.4}

\newcommand{\qaHeldCopyOQwenFour}{4}

\newcommand{\qaHeldCopyErrorsQwenFour}{24}
\newcommand{\qaHeldCopyIntern}{1.0}
\newcommand{\qaHeldCopyInternLo}{0.0}
\newcommand{\qaHeldCopyInternHi}{6.4}

\newcommand{\qaHeldCopyOIntern}{2}

\newcommand{\qaHeldCopyErrorsIntern}{22}
\newcommand{\qaHeldCopyMistral}{2.7}
\newcommand{\qaHeldCopyMistralLo}{0.8}
\newcommand{\qaHeldCopyMistralHi}{6.2}

\newcommand{\qaHeldCopyOMistral}{14}

\newcommand{\qaHeldCopyErrorsMistral}{61}
\newcommand{\qaHeldCopyGlm}{0.4}
\newcommand{\qaHeldCopyGlmLo}{0.0}
\newcommand{\qaHeldCopyGlmHi}{3.8}

\newcommand{\qaHeldCopyOGlm}{1}

\newcommand{\qaHeldCopyErrorsGlm}{25}
\newcommand{\qaHeldCopyLlava}{1.0}
\newcommand{\qaHeldCopyLlavaLo}{0.0}
\newcommand{\qaHeldCopyLlavaHi}{7.6}

\newcommand{\qaHeldCopyOLlava}{2}

\newcommand{\qaHeldCopyErrorsLlava}{21}
\newcommand{\qaHeldCopyMolmo}{3.0}
\newcommand{\qaHeldCopyMolmoLo}{0.0}
\newcommand{\qaHeldCopyMolmoHi}{14.8}

\newcommand{\qaHeldCopyOMolmo}{5}

\newcommand{\qaHeldCopyErrorsMolmo}{27}

\newcommand{\posSplitPassed}{9}

\newcommand{\posSplitEFactorMax}{1.9}

\newcommand{\nullPHolmMax}{1.3\times 10^{-5}}
\newcommand{\transHeldPairs}{300}
\newcommand{\transHeldWriters}{105}
\newcommand{\transHeldPassed}{9}
\newcommand{\transHeldRead}{10}
\newcommand{\transHeldMinimum}{8}

\newcommand{\transHeldTrueAboveTwo}{9}

\newcommand{\transHeldRatioPassedMin}{6.6}
\newcommand{\transHeldRatioPassedMax}{19.5}

\newcommand{\transHeldGlm}{2.1}
\newcommand{\transHeldGlmLo}{0.3}
\newcommand{\transHeldGlmHi}{9.3}

\newcommand{\placeboTransMin}{0.00}
\newcommand{\placeboTransMax}{0.65}
\newcommand{\placeboTransHiMax}{1.15}

\newcommand{\placeboQaMin}{0.00}
\newcommand{\placeboQaMax}{4.41}

\newcommand{\placeboQaHeldMin}{0.00}
\newcommand{\placeboQaHeldMax}{1.67}

\newcommand{\placeboStepMin}{0.00}
\newcommand{\placeboStepMax}{1.17}

\newcommand{\placeboStepAboveTwo}{0}
\newcommand{\legPassed}{9}
\newcommand{\legFalseMin}{62}
\newcommand{\legFalseMax}{341}
\newcommand{\legFalseOf}{400}
\newcommand{\legFailNames}{Molmo2-O-7B}

\newcommand{\legLoPassedMin}{5.0}
\newcommand{\legRateMin}{1.0}
\newcommand{\legRateMax}{10.4}

\newcommand{\frontierLunaO}{31}
\newcommand{\frontierLunaE}{4.0}
\newcommand{\frontierLunaRatio}{7.69}
\newcommand{\frontierLunaRatioLo}{3.71}
\newcommand{\frontierLunaRatioHi}{15.25}

\newcommand{\frontierLunaRepair}{3.4}
\newcommand{\frontierLunaExactT}{27.2}
\newcommand{\frontierLunaExactF}{16.8}
\newcommand{\frontierLunaEqForm}{33.2}
\newcommand{\frontierLunaTokens}{2847}
\newcommand{\frontierLunaCost}{1261}
\newcommand{\frontierLunaVersion}{gpt-6-luna-2026-09-22}
\newcommand{\frontierSolO}{10}
\newcommand{\frontierSolE}{2.9}
\newcommand{\frontierSolRatio}{3.46}
\newcommand{\frontierSolRatioLo}{0.85}
\newcommand{\frontierSolRatioHi}{7.52}
\newcommand{\frontierSolPholm}{0.029}
\newcommand{\frontierSolRepair}{1.1}
\newcommand{\frontierSolExactT}{94.1}
\newcommand{\frontierSolExactF}{86.1}
\newcommand{\frontierSolEqForm}{97.6}
\newcommand{\frontierSolTokens}{108}
\newcommand{\frontierSolCost}{1558}
\newcommand{\frontierSolVersion}{gpt-6.1-sol-2026-09-29}

\newcommand{\frontierReadings}{3600}

\newcommand{\degMinimum}{8}

\newcommand{\degCleanOSum}{431}
\newcommand{\degCleanAlignedSum}{1190}
\newcommand{\degCleanRepairShare}{36.2}
\newcommand{\degCleanESum}{42.2}
\newcommand{\degScanOSum}{235}
\newcommand{\degScanESum}{31.1}
\newcommand{\degScanAlignedSum}{825}
\newcommand{\degScanRepairShare}{28.5}

\newcommand{\degScanPassed}{7}

\newcommand{\degScanFailNames}{Mistral-Small-3.2, GLM-4.6V-Flash, Molmo2-O-7B}

\newcommand{\degBlurPassed}{9}

\newcommand{\degNoisePassed}{8}

\newcommand{\rfPassed}{7}

\newcommand{\rfQGeoMean}{0.64}

\newcommand{\vbRead}{10}
\newcommand{\vbPassed}{6}

\newcommand{\vbRemoved}{2}
\newcommand{\vbPersists}{1}
\newcommand{\vbQGeoMean}{0.53}

\newcommand{\vbRepairMin}{0.0}
\newcommand{\vbRepairMax}{6.2}
\newcommand{\vbRepairStdMin}{0.9}
\newcommand{\vbRepairStdMax}{8.0}
\newcommand{\vbExactTDropFive}{2}
\newcommand{\vbExactTDropMax}{12.3}

\newcommand{\thinkOffRatio}{8.9}
\newcommand{\thinkOffRatioLo}{3.4}
\newcommand{\thinkOffRatioHi}{21.8}
\newcommand{\thinkOffCtx}{6.7}
\newcommand{\thinkOffCtxLo}{3.8}
\newcommand{\thinkOffCtxHi}{10.1}
\newcommand{\thinkPairs}{300}
\newcommand{\thinkOnRatio}{10.7}
\newcommand{\thinkOnRatioLo}{4.4}
\newcommand{\thinkOnRatioHi}{19.9}
\newcommand{\thinkOnCtx}{13.0}
\newcommand{\thinkOnCtxLo}{8.7}
\newcommand{\thinkOnCtxHi}{17.7}

\newcommand{\swRemoved}{1}
\newcommand{\swPractical}{0}
\newcommand{\swExactTMax}{3.6}

\newcommand{\cropRemoved}{6}

\newcommand{\cropPractical}{0}
\newcommand{\cropVLMs}{10}
\newcommand{\cropQGeo}{0.16}
\newcommand{\cropExactDropMin}{24.0}
\newcommand{\cropExactDropMax}{84.9}
\newcommand{\loraPlainRemoved}{1}

\newcommand{\loraPlainPractical}{3}
\newcommand{\loraPlainGemmaTwelveRatioBase}{15.2}
\newcommand{\loraPlainGemmaTwelveRatio}{0.7}
\newcommand{\loraPlainGemmaTwelveQ}{0.04}
\newcommand{\loraPlainGemmaTwelveQLo}{0.00}
\newcommand{\loraPlainGemmaTwelveQHi}{0.37}
\newcommand{\loraPlainGemmaTwelveRepairDiff}{-6.6}
\newcommand{\loraPlainGemmaTwelveExactDiff}{4.9}
\newcommand{\loraPlainGemmaTwelveGenBase}{67.6}
\newcommand{\loraPlainGemmaTwelveGen}{53.1}
\newcommand{\loraPlainGemmaTwelveGenDiff}{-14.5}

\newcommand{\loraPlainQwenNineRatioBase}{7.3}
\newcommand{\loraPlainQwenNineRatio}{3.5}
\newcommand{\loraPlainQwenNineQ}{0.49}
\newcommand{\loraPlainQwenNineQLo}{0.24}
\newcommand{\loraPlainQwenNineQHi}{0.97}
\newcommand{\loraPlainQwenNineRepairDiff}{-4.7}
\newcommand{\loraPlainQwenNineExactDiff}{5.2}
\newcommand{\loraPlainQwenNineGenBase}{68.1}
\newcommand{\loraPlainQwenNineGen}{44.3}
\newcommand{\loraPlainQwenNineGenDiff}{-23.8}

\newcommand{\loraPlainQwenTwentySevenRatioBase}{6.6}
\newcommand{\loraPlainQwenTwentySevenRatio}{3.9}
\newcommand{\loraPlainQwenTwentySevenQ}{0.58}
\newcommand{\loraPlainQwenTwentySevenQLo}{0.14}
\newcommand{\loraPlainQwenTwentySevenQHi}{2.48}
\newcommand{\loraPlainQwenTwentySevenRepairDiff}{-3.4}
\newcommand{\loraPlainQwenTwentySevenExactDiff}{7.4}
\newcommand{\loraPlainQwenTwentySevenGenBase}{62.9}
\newcommand{\loraPlainQwenTwentySevenGen}{44.5}

\newcommand{\loraMixedRemoved}{0}
\newcommand{\loraMixedPassed}{0}
\newcommand{\loraMixedTransfer}{1}
\newcommand{\loraMixedPractical}{3}
\newcommand{\loraMixedGemmaTwelveRatioBase}{15.2}
\newcommand{\loraMixedGemmaTwelveRatio}{3.1}
\newcommand{\loraMixedGemmaTwelveQ}{0.20}
\newcommand{\loraMixedGemmaTwelveQLo}{0.02}
\newcommand{\loraMixedGemmaTwelveQHi}{1.63}
\newcommand{\loraMixedGemmaTwelveRepairDiff}{-6.1}
\newcommand{\loraMixedGemmaTwelveExactDiff}{3.9}
\newcommand{\loraMixedGemmaTwelveGenBase}{67.6}
\newcommand{\loraMixedGemmaTwelveGen}{75.9}

\newcommand{\loraMixedQwenNineRatioBase}{7.3}
\newcommand{\loraMixedQwenNineRatio}{4.4}
\newcommand{\loraMixedQwenNineQ}{0.60}
\newcommand{\loraMixedQwenNineQLo}{0.26}
\newcommand{\loraMixedQwenNineQHi}{1.17}
\newcommand{\loraMixedQwenNineRepairDiff}{-4.4}
\newcommand{\loraMixedQwenNineExactDiff}{5.1}
\newcommand{\loraMixedQwenNineGenBase}{68.1}
\newcommand{\loraMixedQwenNineGen}{71.7}

\newcommand{\loraMixedQwenTwentySevenRatioBase}{6.6}
\newcommand{\loraMixedQwenTwentySevenRatio}{3.1}
\newcommand{\loraMixedQwenTwentySevenQ}{0.47}
\newcommand{\loraMixedQwenTwentySevenQLo}{0.17}
\newcommand{\loraMixedQwenTwentySevenQHi}{1.14}
\newcommand{\loraMixedQwenTwentySevenRepairDiff}{-3.0}
\newcommand{\loraMixedQwenTwentySevenExactDiff}{7.4}
\newcommand{\loraMixedQwenTwentySevenGenBase}{62.9}
\newcommand{\loraMixedQwenTwentySevenGen}{70.2}

\newcommand{\fermatItems}{232}
\newcommand{\fermatPages}{182}
\newcommand{\fermatReadings}{696}
\newcommand{\fermatRepairMin}{47}
\newcommand{\fermatRepairMax}{222}
\newcommand{\fermatRateMin}{7}
\newcommand{\fermatRateMax}{32}
\newcommand{\fermatPlacebo}{0.04}
\newcommand{\fermatVLMs}{10}

\newcommand{\posParsedMin}{-0.1}
\newcommand{\posParsedMax}{4.5}

\newcommand{\fmEqAbove}{9}
\newcommand{\fmDigitAbove}{4}
\newcommand{\fmEqMin}{0.2}
\newcommand{\fmEqMax}{17.9}
\newcommand{\fmDigitMin}{-0.6}
\newcommand{\fmDigitMax}{3.3}

\newcommand{\fmSplitMax}{13.7}
\newcommand{\fmReverseMax}{1.4}
\newcommand{\fmAfterDigitMin}{0.9}
\newcommand{\fmAfterDigitMax}{12.1}
\newcommand{\fmGlmUnparsed}{95.8}

\newcommand{\toRepairMax}{0.7}
\newcommand{\toRepairHiMax}{1.2}
\newcommand{\toRepairZero}{5}

\newcommand{\toCopyMin}{88.6}
\newcommand{\toCopyMax}{100.0}

\newcommand{\lensEqMin}{5.7}
\newcommand{\lensEqMax}{10.6}
\newcommand{\lensDigitMin}{0.8}
\newcommand{\lensDigitMax}{3.2}
\newcommand{\lensRatioMin}{2.6}
\newcommand{\lensRatioMax}{8.8}
\newcommand{\lensFlipYMin}{5.0}
\newcommand{\lensFlipYMax}{15.0}
\newcommand{\lensFlipXMax}{0.7}

\newcommand{\lensReadableMin}{62}
\newcommand{\lensReadableMax}{87}

\newcommand{\atTrueO}{543}
\newcommand{\atTrueE}{27.1}
\newcommand{\atSwapO}{0}
\newcommand{\atSwapE}{2.5}
\newcommand{\atNearO}{2}
\newcommand{\atNearE}{15.8}

\newcommand{\rawConfRho}{0.13}
\newcommand{\rawHeldRho}{0.71}
\newcommand{\rawGlm}{1.8}
\newcommand{\rawGlmRatio}{31.2}
\newcommand{\rawQwenVL}{10.4}
\newcommand{\rawQwenVLRatio}{12.2}

\newcommand{\alignShareMin}{15}
\newcommand{\alignShareMax}{71}

\newcommand{\dhAbove}{9}
\newcommand{\dhRatioMin}{4.1}
\newcommand{\dhRatioMax}{27.5}
\newcommand{\dhAboveTwice}{7}
\newcommand{\dhAboveThrice}{5}

\newcommand{\hsTypical}{10.2}
\newcommand{\hsTypicalLo}{7.4}
\newcommand{\hsTypicalHi}{13.8}
\newcommand{\hsPredLo}{4.7}
\newcommand{\hsPredHi}{21.9}
\newcommand{\hsO}{431}
\newcommand{\hsEop}{42.2}
\newcommand{\hsEtrue}{34.9}
\newcommand{\hsEdig}{38.2}
\newcommand{\hsFalseMin}{2}
\newcommand{\hsFalseMax}{16}

\newcommand{\hcFalse}{60}
\newcommand{\hcTrue}{15}
\newcommand{\hcExactMin}{53}
\newcommand{\hcExactMax}{57}
\newcommand{\hcRepairs}{0}

\newcommand{\fcO}{11}
\newcommand{\fcE}{1.6}
\newcommand{\fcRatio}{6.84}
\newcommand{\fcRatioLo}{2.10}
\newcommand{\fcRatioHi}{28.14}
\newcommand{\fcP}{0.003}
\newcommand{\fcRepair}{1.2}
\newcommand{\fcExactF}{83.4}
\newcommand{\fcExactT}{94.3}
\newcommand{\fcEqForm}{92.8}
\newcommand{\fcTokens}{113}
\newcommand{\fcCost}{2{,}034}
\newcommand{\fcVersion}{claude-opus-5-5}
\newcommand{\fcReadings}{3600}

\newcommand{\stBaseMin}{9.2}
\newcommand{\stBaseMax}{14.1}
\newcommand{\stBestMin}{9.2}
\newcommand{\stBestMax}{12.3}

\newcommand{\koForcedBaseQ}{15.0}
\newcommand{\koForcedQ}{3.7}

\newcommand{\koForcedCtrlQ}{16.0}
\newcommand{\koGenBaseQ}{29.7}
\newcommand{\koGenQ}{7.0}

\newcommand{\koForcedBaseG}{12.0}
\newcommand{\koForcedG}{1.3}

\newcommand{\koForcedCtrlG}{11.7}
\newcommand{\koGenBaseG}{20.7}
\newcommand{\koGenG}{2.7}

\newcommand{\khfVLMs}{7}
\newcommand{\khfRepairBaseMin}{1.1}
\newcommand{\khfRepairBaseMax}{7.3}
\newcommand{\khfRepairMin}{0.2}
\newcommand{\khfRepairMax}{2.7}
\newcommand{\khfRepairLower}{6}

\newcommand{\khfQResolved}{1}
\newcommand{\khfQMin}{0.14}
\newcommand{\khfQMax}{0.86}
\newcommand{\khfRemoved}{0}
\newcommand{\khfPractical}{6}
\newcommand{\khfPassed}{4}

\newcommand{\khfExactDropFive}{1}

\newcommand{\kfVLMs}{7}

\newcommand{\kfNoneMin}{2.3}
\newcommand{\kfNoneMax}{15.7}
\newcommand{\kfFlowMin}{0.7}
\newcommand{\kfFlowMax}{6.0}
\newcommand{\kfRandomMaxShift}{0.7}

\newcommand{\vbBoth}{1}
\newcommand{\swBoth}{0}
\newcommand{\cropBoth}{0}
\newcommand{\loraPlainBoth}{1}
\newcommand{\loraMixedBoth}{0}

\newcommand{\khfBoth}{0}

\newcommand{\esRatioOne}{19.1}

\newcommand{\esRatioTwo}{14.6}

\newcommand{\esRatioThree}{12.5}

\newcommand{\srVLMs}{10}
\newcommand{\srPassed}{9}
\newcommand{\srDeltaMin}{0.7}
\newcommand{\srDeltaMax}{19.0}
\newcommand{\srRatio}{0.64}

\newcommand{\snSmoothLowMin}{4.3}
\newcommand{\snSmoothPassedMin}{9}
\newcommand{\snScanRatioMin}{3.1}
\newcommand{\snScanRatioMax}{13.5}

\newcommand{\gnZeroE}{5}

\newcommand{\gnOSum}{431}
\newcommand{\gnESum}{14.0}
\newcommand{\gnClearRepairs}{218}

\newcommand{\gnClearVLMs}{10}

\newcommand{\fpRepairSum}{1{,}040}
\newcommand{\fpPlaceboSum}{3}
\newcommand{\fpPlaceboZero}{8}

\definecolor{cvprblue}{rgb}{0.21,0.49,0.74}
\usepackage[pagebackref,breaklinks,colorlinks,allcolors=cvprblue]{hyperref}
\newcommand{\ci}[2]{\,[#1,\,#2]}

\newcommand{\data}{DiploMat}
\title{DiploMat: Faithful Transcription Benchmark for Handwritten Arithmetic}
\author{Takaya Kawakatsu\\Preferred Networks, Inc.\\{\tt\small kat.nii.ac.jp@gmail.com}\\{\small\href{https://researchmap.jp/t.kat}{\tt researchmap.jp/t.kat}}
\and
Ryo Ishiyama\\Kyushu University\\{\tt\small ryo.ishiyama@human.ait.kyushu-u.ac.jp}\\{\small\href{https://researchmap.jp/rishiyama}{\tt researchmap.jp/rishiyama}}}

\begin{document}
\maketitle

\begin{abstract}
Vision-language models (VLMs) are used to transcribe and grade handwritten mathematics, where their reading should report what was written.
To measure this, we introduce \data, a benchmark of false handwritten equations such as $37\times9=533$, composed from the recorded pen strokes of the CROHME competitions so that the written digits are known by construction, and a protocol that compares the observed count $O$ of \emph{repairs}, the transcriptions that change a written digit into the arithmetically correct one, against the count $E$ that the reader's own digit confusions would predict.
In a confirmatory test with held-out pairs whose criterion was fixed before the reading, \transHeldPassed{} of 10 open VLMs repair false equations with the lower bound of $O/E$ above 2; on clean renderings in the confusion-null test, repairs are \legRateMin--\legRateMax\% of readings of false equations.
Further analyses locate the effect in the use of the arithmetic rather than in the reading of the glyph: a glyph that is read as written when only its digit is asked for is repaired when the whole equation is transcribed (\testThreePassed{} of 10 VLMs, criterion fixed in advance); the same equations given as typed text are copied with at most \toRepairMax\% repairs; and preventing the generated positions from attending to the written operands lowers the context effect on same-glyph images in each of the \kfVLMs{} VLMs tested (post hoc), though on held-out pairs the reduction in $O/E$ is resolved for only \khfQResolved{} of \khfVLMs{}.
GPT-6 Luna meets the criterion ($O/E$ \frontierLunaRatio), and so does Claude Opus 5.5 (post hoc), whereas the test lacks power for GPT-6.1 Sol, which misreads too rarely.
Under scan-like degradation only \degScanPassed{} of 10 VLMs meet the criterion, below the \degMinimum{} required, and on photographed solutions the VLMs write the correct value in place of a wrong one in \fermatRateMin--\fermatRateMax\% of the readings, whereas an equally distant placebo value only \fpPlaceboSum{} times, far below expectation, so the registered null is uncalibrated.
\end{abstract}

\begin{figure*}[tb]
  \centering
  \includegraphics[width=\textwidth]{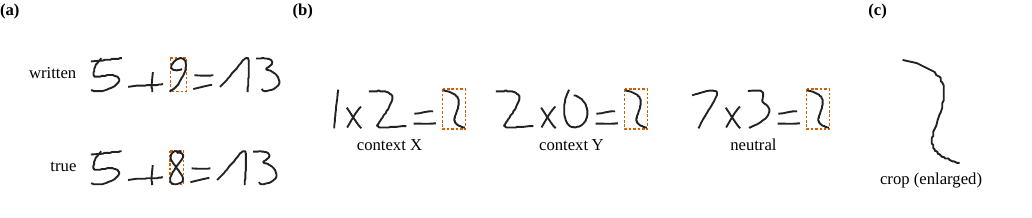}
  \caption{Stimuli from one writer's ink: (a) a false equation and its true equation, (b) one glyph in three contexts, (c) the glyph alone.}
  \label{fig:stimuli}
\end{figure*}

\section{Introduction}
\label{sec:intro}

Vision-language models (VLMs) are used to transcribe and to grade handwritten mathematics~\cite{yu2026grading,seong2026fix}.
A reading that reports what should have been written instead of what was written hides the errors that grading is meant to detect.
For example, suppose that a student has written $37\times9=533$ on a page.
In the failure we study, a VLM asked to transcribe the page writes $37\times9=333$, the correct product, although the digit written in that position is a 5.
Asked only for that digit, the same VLM reads the same pixels as written.

Human readers identify letters better in words than in isolation~\cite{reicher1969}, and a model that resolves an ambiguous digit from the surrounding arithmetic can read correct documents more accurately.
Whether such use of the arithmetic context is appropriate depends on the question.
A question about the correct result needs the arithmetic, whereas a question about what the student wrote requires the model to set it aside.
We call a reading \emph{faithful} when it reports the written symbols, and we measure whether VLMs use the arithmetic only when the question requires it.

We build \data{} from the online ink of the CROHME test expressions~\cite{crohme2016,crohme2019,crohme2023} (Fig.~\ref{fig:stimuli}).
A \emph{false equation} is a single-operation equation, composed from one writer's own glyphs, that differs from its \emph{true equation} in one digit; a \emph{repair} is a transcription of a false equation that equals its true equation.
Repairs alone do not show that a model uses arithmetic knowledge, because a noisy reader also changes digits.
We therefore compare every count of repairs with a \emph{confusion null}, the count expected from the same reader's own digit confusions.
We write $O$ for an observed count and $E$ for its confusion-null prediction, and use $O/E$ as the effect size.
Over-correction in handwritten solutions has been reported by counting corrected planted errors~\cite{seong2026fix}; such counts leave open whether the corrections exceed the reader's own misreadings and whether the written digit is misperceived or overridden; we answer the first with the confusion null and the second by holding the image fixed while the question changes.

VLMs change written digits toward the arithmetically correct result more often than a null from the same ink predicts, as confirmed on held-out pairs with a criterion fixed before their readings (Sec.~\ref{sec:evidence}).
This held-out test is the confirmatory result; secondary tests had their criteria fixed before their readings but after earlier tests, and analyses chosen after the data were seen are marked post hoc.
A placebo calibration shows that the null does not inflate $O/E$, while a null estimated from readings of the digits alone, which carry no arithmetic, gives the same decision for \sixDigitsAbove{} of 10 VLMs; a restriction to equations whose digits each reader reads correctly in isolation reveals that the repairs are not limited to illegible ink; and the same pixels that a VLM reports as written when asked only for the digit are repaired when it writes the whole equation, also within one reply that gives both (Sec.~\ref{sec:sameglyph}).
The repairs need the image, since typed copies of the same false equations are almost never repaired; a logit lens places the context effect in late layers, and blocking attention to the written operands reduces the same-glyph context effect in all \kfVLMs{} VLMs tested (post hoc; Sec.~\ref{sec:layers}).
These results hold for clean renderings; under scan-like degradation \degScanPassed{} of 10 VLMs meet the criterion, below the \degMinimum{} required, and on photographed handwritten solutions the VLMs also write correct values in place of wrong ones (Sec.~\ref{sec:further}).
Table~\ref{tab:claims} lists each claim with the tests that support it and their status.

\begin{table}[tb]
  \centering
  \footnotesize
  \setlength{\tabcolsep}{2.5pt}
  \begin{tabular}{@{}p{0.34\columnwidth}p{0.42\columnwidth}l@{}}
    \toprule
    claim & tests & status \\
    \midrule
    repair beyond confusions & held-out pairs (Table~\ref{tab:heldout}) & confirmatory \\
    & confusion-null, meaningless operator & secondary \\
    null not inflated & placebo, split and digits-only nulls & post hoc \\
    not limited to illegible ink & legible equations & post hoc \\
    arises when the equation is written & same-glyph test & secondary \\
    & one-reply format, typed copies, other arithmetic targets, logit lens, knockout & post hoc \\
    remedies trade repair for accuracy & four remedies & secondary \\
    \bottomrule
  \end{tabular}
  \caption{Claims, supporting tests and their status.}
  \label{tab:claims}
\end{table}

We make three contributions: (i) a protocol for faithful transcription that compares every count of repairs with a per-reader confusion null and varies the context around pixel-identical glyphs, with \data{} as its instance built from human ink and applicable to any transcription task whose faithful reading is known by construction; (ii) the finding, replicated on held-out pairs, that most of 10 VLMs repair false equations beyond what their own confusions predict, with a secondary same-glyph test and post hoc evidence that the written digit is overridden when the whole equation is written; and (iii) a comparison of remedies (Sec.~\ref{sec:remedies}), none of which brings more than one VLM below the criterion without losing accuracy.

\section{Related work}
\label{sec:related}

\paragraph{Priors over visual evidence.}
VLM errors are known to arise from language priors that override the evidence in the image~\cite{pope2023,vcd2024,deng2025words}.
On counterfactual images, for example a logo with an added stripe, VLMs answer from memorized knowledge~\cite{vlmsbiased2026}; they also report removed objects~\cite{he2025removed} and miss simple visual details~\cite{rahmanzadehgervi2024blind}.
Larger VLMs more often follow counterfactual priors, and the language prior is set in early layers before the visual evidence is integrated~\cite{grob2026bigger} while across 24 models the reliance on priors, including arithmetic ones, does not decrease with scale~\cite{visip2026}.
These studies vary the image with the question fixed, whereas we fix the image and vary the question, so that the same pixels require the prior under one question and forbid it under another.

\paragraph{Encoded but not used.}
On vision-centric tasks, a direct readout of the vision encoder outperforms the answers of the VLM~\cite{fu2025hidden}, and VLMs attend to the evidence while they answer wrongly~\cite{seeingnotbelieving2026}.
Nooralahzadeh \etal~\cite{arbitration2026} find that visual attributes remain decodable in 10 VLMs when the answer follows world knowledge, and their patching locates the causal effect in the image tokens rather than in the prompt text.
Golovanevsky \etal~\cite{golovanevsky2025pixels} use the logit lens to trace how predictions on counterfactual colors and sizes move between prior and image across layers, and language models compute arithmetic results using sparse heuristic neurons~\cite{nikankin2025arithmetic}, a candidate source of the observed prior.

\paragraph{Rewriting in transcription and grading.}
Lee \etal~\cite{faithc4} find that, on perturbed document images, general VLMs degrade more than OCR models in word error rate and trace rewriting in one VLM to its final-layer representation of the word.
WildHandBench~\cite{zhang2026wildhandbench} attributes reading errors on handwritten documents to language priors when the output is more probable than the truth under a language model, a criterion that ignores the visual confusions of the reader.
Post-OCR correction~\cite{rigaud2019postocr} moves the recognized text toward plausible text, a change that faithful reading has to avoid.
Seong \etal~\cite{seong2026fix} show that VLMs over-correct handwritten student solutions built with deliberate errors from FERMAT~\cite{fermat2025} and propose a metric that penalizes such corrections, while error detection in student work is benchmarked separately~\cite{errorradar2026}.
Unlike these studies, we fix the faithful reading by construction, and the prior that we measure is the result of the written arithmetic rather than a word or an object.
Specialist recognizers~\cite{comer2022,guan2024posformer,zhu2024ical,zhu2025tamer} of handwritten mathematics are trained on CROHME, and language-model fusion has been designed to correct their recognition errors~\cite{chen2023rlfn}.

\section{Benchmark, readers and measures}
\label{sec:method}

\paragraph{Terms.}
A \emph{reader} is a VLM or a recognizer.
An \emph{aligned misreading} is a wrong reading that keeps the operator and the digit count of every number.
The \emph{excess} is the number of observed minus expected repairs, divided by the number of aligned misreadings, in percentage points (hereafter simply points).
A target glyph is \emph{clear} for a reader when that reader transcribes it, cropped alone, as the written digit in all three readings.
A false equation is \emph{legible} for a reader when that reader transcribes its digits-only view, the digits without the operator and the equals sign, as written in all three readings (post hoc).

\paragraph{Ink and equations.}
We take the online ink and symbol segmentation of the test expressions from the CROHME 2016, 2019 and 2023 sets~\cite{crohme2016,crohme2019,crohme2023} in the CROHME 2023 release, distributed under CC BY-NC-SA 3.0, which allows non-commercial use and requires derivatives to carry the same license.
We split all writers into \emph{design writers}, whose ink was used in pilot runs before the criteria were fixed, and \emph{test writers}, whose ink was not.
An equation $a\circ b=c$ with $\circ\in\{+,-,\times\}$ is composed from one writer's glyphs and rendered at $448{\times}448$ pixels.
Each glyph is scaled so that digits take the writer's median digit height, glyphs share one baseline with a gap of a quarter of the glyph height, and the equation is centered in a square frame with a 6\% margin and is drawn with 2-pixel strokes.
A false equation differs from its true equation in exactly one digit of the result or of an operand, by a step drawn uniformly from $\pm1$, $\pm2$ and $\pm3$ without regard to confusability, and every number keeps its number of digits.
A \emph{pair} consists of one false equation and its true equation.
A repair writes the digit of the true equation at the changed position, an operand digit or a result digit; question answering uses only pairs with a changed result digit, for which the computed value is the result of the true equation.
For each equation we also render a view that has no arithmetic meaning, the \emph{meaningless-operator view}, the same image with its $\circ$ replaced by a handwritten triangle, which is the writer's own where available.
Three tests compare repairs with a null: the \emph{confusion-null test} (\testTwoNPairs{} pairs of test writers, null from the true equations), the \emph{meaningless-operator test} (\sixNPairs{} pairs of \sixNWriters{} further design writers, null from the meaningless-operator view) and its repetition on \emph{held-out pairs} (\transHeldPairs{} new pairs from \transHeldWriters{} test writers, same null); the \emph{same-glyph test} uses \testThreeNItems{} glyphs of \testThreeNWriters{} test writers.

\paragraph{Glyph assignment.}
We assign glyphs by position, so that a false equation and its true equation share every unchanged glyph instance and no pair uses a glyph instance twice.
An earlier version of the generator violated both conditions; all stimuli were regenerated after the decision rules were fixed and form the released benchmark used by every test.

\paragraph{Readers.}
The 10 VLMs in this study are \mbox{Gemma-4-31B} and 12B~\cite{gemma4}; \mbox{Qwen3.8-27B}~\cite{qwen38_27b}; \mbox{Qwen3.5-9B} and 4B~\cite{qwen35}; \mbox{Mistral-Small-3.2-24B}~\cite{mistralsmall32} (hereafter \mbox{Mistral-Small-3.2}); \mbox{InternVL3.5-38B}~\cite{internvl35}; \mbox{GLM-4.6V-Flash}~\cite{glm46vflash,glmv2025}; as well as \mbox{LLaVA-OneVision-2-8B}~\cite{an2026llavaonevision2} and \mbox{Molmo2-O-7B}~\cite{clark2026molmo2}.
All are served locally by vLLM on NVIDIA A100 GPUs in bfloat16 from fixed weight revisions on Hugging Face, with fixed seeds and token limits that the release records, and with the thinking mode of the chat template turned off for the models that have such a mode.
We read each image three times unless stated otherwise, once at temperature 0 and twice at temperature 1.0.
Four further readers are the recognizers CoMER~\cite{comer2022}, PosFormer~\cite{guan2024posformer}, TAMER~\cite{zhu2025tamer} and ICAL~\cite{zhu2024ical}, which are trained on CROHME training data that do not overlap with the CROHME test expressions used here and have had no large-scale language pretraining.

\paragraph{Prompts and scoring.}
For transcription, each one of the VLMs is given the prompt ``The image shows a handwritten mathematical expression. Transcribe it into LaTeX. Output only the LaTeX code, without \$ delimiters or explanations.''
The \emph{extraction rules} keep a single-line reply as it is and otherwise take the last line labeled as \LaTeX{} or the last code block.
We convert the extracted \LaTeX{} to numbers by removing markup and spaces and then reading off the operator, the equals sign and the digit strings between them; readings that cannot be converted are counted as unparsed and not repaired.

\paragraph{Confusion null.}
For each reader we count, in the aligned readings of a reference view, how often each written digit $w$ is read as a digit $v$.
We then add 0.1 to every count and normalize each row, which gives a $10{\times}10$ confusion matrix $C$ with $C[w,v]$ the probability of reading $v$ for a written $w$.
The reference view is the true equation in the confusion-null test and the meaningless-operator view in the meaningless-operator test; in that test, as registered, $C$ is estimated per role (result or operand) and per place (units or higher).
For an aligned misreading of a false equation, the probability that independent digit noise produces exactly the true equation is
\begin{equation}
  q=\frac{\prod_i C[w_i,t_i]}{1-\prod_i C[w_i,w_i]},
\end{equation}
where $i$ runs over all digit positions, $w_i$ and $t_i$ are the written and true digits, and the denominator is the probability that one or more digits are misread.
Because a repair is an exact true equation and hence always aligned, the null is applied only to aligned misreadings, which account for \alignShareMin--\alignShareMax\% of the misreadings of false equations per VLM in the held-out test.
The expected count $E$ is the sum of $q$ over the aligned misreadings, $O$ is the observed number of repairs, and the excess is $(O-E)$ divided by the number of aligned misreadings.
The null assumes that the digit errors are independent across positions and $C$ independent of the surrounding glyphs and of the position.
The main risk is an underestimate of $E$: a position with more misreadings than average, or an arithmetic prior that suppresses misreadings of true equations, would lower it.
We address this risk post hoc.
In the confusion-null test, with $C$ estimated per role (result or operand) and per place (units or higher), $E$ grows by at most a factor of \posSplitEFactorMax{}, and the same \posSplitPassed{} VLMs keep a lower bound above 2.
With $C$ taken from the digits-only view, which has neither operator nor equals sign and hence no arithmetic, \sixDigitsAbove{} of 10 VLMs keep a lower bound above 2 on the design-writer pairs ($O/E$ \sixDigitsRatioMin--\sixDigitsRatioMax).
Summed over the 10 VLMs on the held-out pairs, $E$ is \hsEtrue{} from true equations, \hsEdig{} from the digits-only view and \hsEop{} from the meaningless-operator view, against the \hsO{} observed repairs.
We also read the digits-only view of the held-out pairs; against this null, \dhAbove{} of 10 VLMs keep a lower bound of more than 2 ($O/E$ \dhRatioMin--\dhRatioMax), and \dhAboveTwice{} and \dhAboveThrice{} still do if $E$ is doubled or tripled.
In the confusion-null test, summed $O/E$ is \esRatioOne, \esRatioTwo{} and \esRatioThree{} for planted changes of size 1, 2 and 3 (post hoc), and replacing the added 0.1 by values from 0.01 to 1 shifts the excess by at most \testTwoSmoothMaxShift{} points, with the \snSmoothPassedMin{} passing VLMs keeping lower bounds of at least \snSmoothLowMin.
The split and smoothing analyses use the confusion-null test and the digits-only null uses both writer sets; against the confusion null, the confirmatory held-out test, decided against the meaningless-operator null, also gives a lower bound above 2 for \transHeldTrueAboveTwo{} of the 10 VLMs.
No null is free of assumptions; the values under the different nulls are estimates under different assumptions rather than independent confirmations.

\paragraph{Statistics and criterion.}
All intervals are 95\% percentile intervals from 10{,}000 bootstrap replicates that resample the writers and re-estimate $C$ each time.
A model meets the criterion if its Holm-adjusted $p$ (over the 10 VLMs) is below 0.05 and the lower bound of the 95\% interval of its $O/E$ is above 2, that is, if repairs are at least twice as frequent as its confusions predict.
A bound of 2 rather than 1 means that an underestimate of $E$ by up to a factor of two cannot by itself meet the criterion.
The bootstrap $p$ often sits at its floor, so decisions rest on the lower bound; post hoc $p$ values from the null distribution of $O$ are at most $\nullPHolmMax$ after Holm adjustment in the confusion-null test (Sec.~\ref{S-sec:stats}).
For the same-glyph effect, the weaker criterion on a difference requires a lower bound above 0.
A test over several VLMs is met when the number of VLMs that meet the criterion reaches a minimum fixed with the criterion; we call this minimum the \emph{rule} of the test.
To compare a model under two conditions, such as with and without a remedy, we use $Q$, the ratio of its $O/E$ under the two conditions, with a paired writer bootstrap.

\paragraph{Test history.}
The stimuli, measures and criteria of all the main tests were written in a version-controlled document before their readings were collected, and analyses chosen after the data were seen are labeled \emph{post hoc}; such analyses are descriptive and carry no decision.
The first transcription test, which required four of five VLMs to repair the false equations more often than they read the true equations as false, failed; since that comparison ignores how often each reader misreads digits, the confusion null was chosen after this outcome and then fixed before the stimuli of the second test were generated.
For this reason we consider as fully confirmatory only the repetition on \emph{held-out pairs}, which are new pairs from test writers generated and read after the confusion-null, meaningless-operator and same-glyph tests were run.
Five VLMs were planned and 10 were read; Table~\ref{S-tab:deviations} lists the deviations from the written plans of the main tests.

\paragraph{Using the benchmark.}
For any new model, the released code reports $O/E$ with its interval and also the same-glyph effect.
We recommend reporting the lower bounds of the ratio $O/E$ against both nulls and the same-glyph effect on clear glyphs; because the intervals are wide (for the models that meet the criterion, the upper bound of $O/E$ is a median \testTwoCiFactorMedian{} times the lower bound), the 400 pairs support the binary decision of the criterion rather than a ranking of models by $O/E$.
The test-writer sets should not be used for tuning; released hashes allow checking the images.

\paragraph{Data and release.}
The benchmark consists of the images and views, the writer, operator, changed position and step of every pair, and the protocol of this section; these, the prompts, the readings, a datasheet, Croissant metadata and the code for every number will be released in a versioned public repository with a persistent identifier, with the code under the MIT license and the data under CC BY-NC-SA 3.0, the license of the CROHME 2023 release~\cite{crohme2023}, which allows only non-commercial use.

\section{Repair in transcription}
\label{sec:evidence}

\begin{table}[tb]
  \centering
  \footnotesize
  \setlength{\tabcolsep}{2.5pt}
  \begin{tabular}{@{}lrrrcc@{}}
    \toprule
    reader & $O$ & rep., \% & $E$ & $O/E$ [95\% interval] & met \\
    \midrule
    Gemma-4-31B & 30 & 3.3 & 2.4 & 12.7 [4.6, 38.3] & yes \\
    Gemma-4-12B & 61 & 6.8 & 4.0 & 15.2 [6.8, 28.9] & yes \\
    Qwen3.8-27B & 41 & 4.6 & 6.2 & 6.6 [3.6, 12.3] & yes \\
    Qwen3.5-9B & 58 & 6.4 & 8.0 & 7.3 [4.0, 11.4] & yes \\
    Qwen3.5-4B & 72 & 8.0 & 3.7 & 19.5 [9.1, 32.6] & yes \\
    InternVL3.5-38B & 62 & 6.9 & 3.4 & 18.4 [6.8, 57.3] & yes \\
    Mistral-Small-3.2 & 49 & 5.4 & 5.4 & 9.1 [4.7, 23.0] & yes \\
    GLM-4.6V-Flash & 8 & 0.9 & 3.8 & 2.1 [0.3, 9.3] & no \\
    LLaVA-OneVision-2-8B & 24 & 2.7 & 3.0 & 7.9 [3.8, 20.9] & yes \\
    Molmo2-O-7B & 26 & 2.9 & 2.3 & 11.1 [5.3, 23.9] & yes \\
    \midrule
    GPT-6 Luna & 31 & 3.4 & 4.0 & 7.7 [3.7, 15.3] & yes \\
    GPT-6.1 Sol & 10 & 1.1 & 2.9 & 3.5 [0.8, 7.5] & no \\
    Claude Opus 5.5\textsuperscript{\dag} & 11 & 1.2 & 1.6 & 6.8 [2.1, 28.1] & yes \\
    \bottomrule

{}  \end{tabular}
  \caption{Confirmatory test on \transHeldPairs{} held-out pairs (\dag{} post hoc).}
  \label{tab:heldout}
\end{table}

\begin{table*}[tb]
  \centering
  \footnotesize
  \setlength{\tabcolsep}{3pt}
  \begin{tabular}{@{}lrrrccccc@{}}
    \toprule
    & \multicolumn{6}{c}{confusion-null test} & meaningless op. & same-glyph test \\
    \cmidrule(lr){2-7}
    reader & $O$ & rep., \% & $E$ & $O/E$ & temp.\ 0 (lo) & legible (lo) & $O/E$ & Y$-$X, points \\
    \midrule
    Gemma-4-31B & 54 & 4.5 & 0.9 & 57.4 [22.8, 122.9] & 50.8 (25.3) & 100.6 (33.1) & 9.6 [2.8, 67.0] & 7.0 [3.5, 11.0] \\
    Gemma-4-12B & 100 & 8.3 & 4.4 & 22.5 [12.3, 56.6] & 23.6 (13.7) & 19.8 (8.7) & 8.9 [4.0, 14.3] & 23.9 [17.4, 30.6] \\
    Qwen3.8-27B & 51 & 4.2 & 3.8 & 13.3 [7.3, 54.6] & 18.7 (10.0) & 11.5 (5.0) & 8.9 [3.4, 21.8] & 6.7 [3.8, 10.1] \\
    Qwen3.5-9B & 125 & 10.4 & 10.2 & 12.2 [6.7, 25.3] & 20.0 (12.0) & 11.8 (5.6) & 8.0 [4.3, 11.2] & 29.1 [23.1, 35.6] \\
    Qwen3.5-4B & 84 & 7.0 & 6.0 & 13.9 [6.9, 32.6] & 26.8 (13.6) & 15.7 (5.7) & 16.3 [7.8, 24.2] & 22.4 [17.3, 27.8] \\
    InternVL3.5-38B & 99 & 8.2 & 5.0 & 19.8 [9.9, 37.6] & 20.0 (10.7) & 22.5 (9.0) & 12.9 [5.3, 25.0] & 12.0 [8.0, 16.3] \\
    Mistral-Small-3.2 & 69 & 5.8 & 8.3 & 8.3 [4.8, 20.4] & 12.2 (6.5) & 18.1 (5.5) & 11.6 [4.9, 33.5] & 7.7 [4.6, 11.2] \\
    GLM-4.6V-Flash & 22 & 1.8 & 0.7 & 31.2 [8.3, 160.4] & 37.1 (12.9) & 54.4 (6.1) & 7.2 [2.3, 25.8] & 2.1 [0.5, 4.3] \\
    LLaVA-OneVision-2-8B & 50 & 4.2 & 4.1 & 12.3 [5.4, 38.7] & 10.2 (4.4) & 15.1 (6.6) & 8.0 [3.1, 17.2] & 8.8 [5.3, 12.7] \\
    Molmo2-O-7B & 12 & 1.0 & 2.6 & 4.5 [1.5, 13.6] & 1.1 (0.0) & 0.0 (0.0) & 8.6 [3.1, 20.5] & 4.4 [2.0, 7.4] \\
    \midrule
    four recognizers & 3 & -- & 1.5 & -- & -- & -- & -- & 0.0 \\
    \bottomrule

{}
  \end{tabular}
  \caption{Transcription repair against the confusion and meaningless-operator nulls.}
  \label{tab:phenomenon}
\end{table*}

\begin{figure}[tb]
  \centering
  \includegraphics[width=\columnwidth]{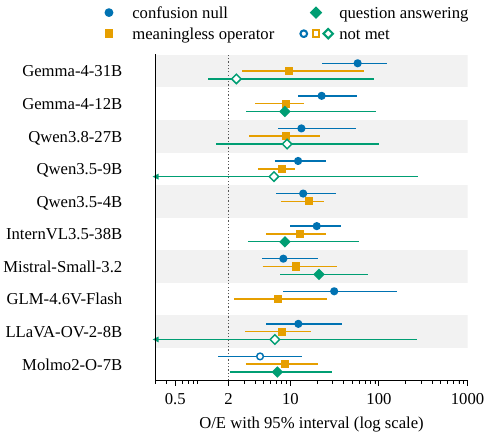}
  \caption{O/E with 95\% intervals per model and test (log scale).}
  \label{fig:effects}
\end{figure}

\paragraph{Confirmatory result on held-out pairs.}
The rule, fixed before the \transHeldPairs{} held-out pairs from \transHeldWriters{} test writers were generated and transcribed, required \transHeldMinimum{} of 10 VLMs to meet the criterion against the meaningless-operator null, and \transHeldPassed{} of 10 meet it, and the rule is thus met ($O/E$ \transHeldRatioPassedMin--\transHeldRatioPassedMax; ``rep.'' in Table~\ref{tab:heldout} is the share of repairs among readings of false equations); GLM-4.6V-Flash alone does not meet it (\transHeldGlm\ci{\transHeldGlmLo}{\transHeldGlmHi}).
Over the 10 VLMs, a random-effects mean gives a typical $O/E$ of \hsTypical\ci{\hsTypicalLo}{\hsTypicalHi}, with a 95\% prediction interval of \hsPredLo--\hsPredHi{} for a further VLM of this kind (post hoc).
The same test was met on the design-writer pairs by \sixPassedA{} VLMs, and in the confusion-null test on \testTwoNPairs{} pairs of test writers by \testTwoPassed{} of 10, with $O/E$ from \testTwoRatioPassedMin{} to \testTwoRatioPassedMax{} (Table~\ref{tab:phenomenon}, Fig.~\ref{fig:effects}); only Molmo2-O-7B ($O/E$ \testTwoRatioMin) does not.
In Table~\ref{tab:phenomenon}, ``temp.\ 0'' and ``legible'' give $O/E$ from temperature-0 readings and legible equations (lower bound in parentheses).

\paragraph{Commercial models.}
We applied the held-out test, whose criterion was fixed before any of the commercial models read a stimulus, to GPT-6 Luna and GPT-6.1 Sol through their API at their default reasoning effort and temperature (Sec.~\ref{S-sec:frontier}) with a Holm correction over the two models; Luna meets the criterion (\frontierLunaRatio\ci{\frontierLunaRatioLo}{\frontierLunaRatioHi}) with \frontierLunaO{} repairs against \frontierLunaE{} expected.
Sol transcribes \frontierSolExactF\% of false and \frontierSolExactT\% of true equations exactly, and its \frontierSolO{} repairs against \frontierSolE{} expected (\frontierSolRatio\ci{\frontierSolRatioLo}{\frontierSolRatioHi}) do not meet the criterion; the test lacks power, and this is no evidence that repair vanishes.
Luna outputs an equation of the expected form for only \frontierLunaEqForm\% of true equations (post hoc), so its exact-match rates of \frontierLunaExactT\% for true and \frontierLunaExactF\% for false equations also reflect unstable transcription.
Claude Opus 5.5, run post hoc at API defaults, transcribes \fcExactF\% of false and \fcExactT\% of true equations exactly and meets the criterion with \fcO{} repairs against \fcE{} expected (\fcRatio\ci{\fcRatioLo}{\fcRatioHi}), a lower bound close to 2.

\paragraph{Control 1: the null is calibrated.}
As a post hoc check of calibration, we replace the true result by placebo targets, the true result with its changed digit replaced by any digit that is neither written nor true; a calibrated null predicts these as often as they occur, and the confusion null gives $O/E$ of \placeboTransMin--\placeboTransMax{} (upper bounds at most \placeboTransHiMax), so it does not inflate $O/E$.
Restricted to placebo targets 1 to 3 steps from the written digit, like the true digit, $O/E$ is \placeboStepMin--\placeboStepMax{} and \placeboStepAboveTwo{} lower bounds exceed 2 (post hoc).

\paragraph{Control 2: legible ink.}
To address the alternative that the VLMs guess illegible digits from the arithmetic, we restrict the confusion-null test post hoc to the false equations that each reader finds legible (\legFalseMin--\legFalseMax{} of \legFalseOf); \legPassed{} of 10 VLMs still meet the criterion (lower bounds at least \legLoPassedMin; column ``legible'' of Table~\ref{tab:phenomenon}), and \legFailNames{} does not.
A null built from each changed glyph, the share of its digits-only readings that give the true digit, predicts \gnESum{} of the \gnOSum{} held-out repairs over the 10 VLMs ($E=0$ for \gnZeroE{} VLMs; post hoc), and \gnClearRepairs{} repairs, in all \gnClearVLMs{} VLMs, fall on glyphs that are read as written in all three digits-only readings.
Two authors, not told which equations were false, transcribed \hcFalse{} false and \hcTrue{} true held-out equations (post hoc); they read \hcExactMin{} and \hcExactMax{} of the false equations as written, and \hcRepairs{} of their misreadings give the true equation.

\paragraph{Recognizers.}
The four recognizers repair only \revRecRepairs{} of \revRecReadings{} readings of false equations, where their confusions predict \testTwoRecESum, but their power to detect an $O/E$ of 10 is \revRecPowerMin--\revRecPowerMax{} (post hoc).

\paragraph{Control 3: the same glyph in different contexts.}
\label{sec:sameglyph}

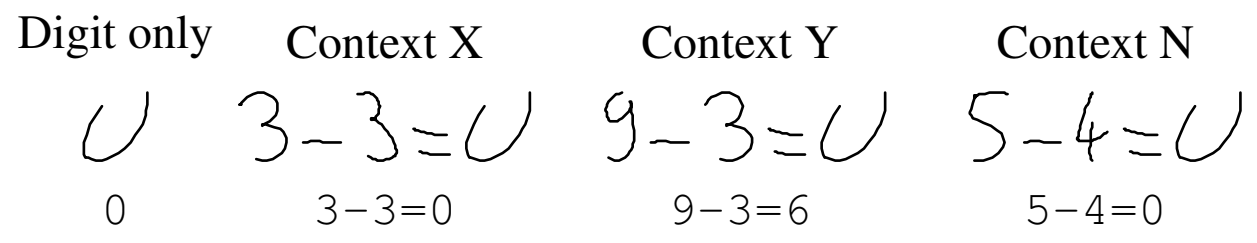
\begin{figure}[tb]
  \centering
  \begin{tikzpicture}
\draw[line width=0.5pt, line cap=round, line join=round] (0.42,-0.94) -- (0.38,-1.00) -- (0.34,-1.06) -- (0.32,-1.13) -- (0.30,-1.18) -- (0.30,-1.22) -- (0.30,-1.26) -- (0.31,-1.28) -- (0.33,-1.28) -- (0.35,-1.29) -- (0.38,-1.30) -- (0.42,-1.30) -- (0.45,-1.29) -- (0.49,-1.27) -- (0.53,-1.25) -- (0.57,-1.22) -- (0.61,-1.18) -- (0.64,-1.13) -- (0.67,-1.09) -- (0.69,-1.06) -- (0.70,-1.03) -- (0.71,-1.00) -- (0.71,-0.97) -- (0.72,-0.93) -- (0.72,-0.89) -- (0.72,-0.85);
\node[anchor=south, font=\small] at (0.51,-0.75) {Digit only};
\node[anchor=north, font=\small\ttfamily] at (0.51,-1.40) {0};
\draw[line width=0.5pt, line cap=round, line join=round] (1.34,-0.93) -- (1.33,-0.93) -- (1.33,-0.93) -- (1.34,-0.93) -- (1.34,-0.92) -- (1.36,-0.90) -- (1.38,-0.88) -- (1.40,-0.87) -- (1.42,-0.86) -- (1.45,-0.85) -- (1.47,-0.85) -- (1.49,-0.85) -- (1.51,-0.86) -- (1.53,-0.87) -- (1.55,-0.88) -- (1.57,-0.90) -- (1.58,-0.92) -- (1.58,-0.94) -- (1.58,-0.97) -- (1.57,-1.00) -- (1.55,-1.02) -- (1.53,-1.04) -- (1.51,-1.05) -- (1.49,-1.06) -- (1.48,-1.07) -- (1.48,-1.07) -- (1.48,-1.06) -- (1.48,-1.06) -- (1.49,-1.06) -- (1.50,-1.06) -- (1.53,-1.07) -- (1.56,-1.08) -- (1.60,-1.09) -- (1.62,-1.11) -- (1.63,-1.13) -- (1.64,-1.15) -- (1.64,-1.18) -- (1.63,-1.22) -- (1.61,-1.25) -- (1.58,-1.27) -- (1.55,-1.29) -- (1.51,-1.30) -- (1.48,-1.30) -- (1.44,-1.29);
\draw[line width=0.5pt, line cap=round, line join=round] (1.75,-1.18) -- (1.76,-1.18) -- (1.78,-1.18) -- (1.80,-1.17) -- (1.82,-1.17) -- (1.84,-1.17) -- (1.86,-1.17) -- (1.88,-1.18) -- (1.91,-1.18) -- (1.94,-1.18) -- (1.97,-1.18) -- (2.00,-1.18);
\draw[line width=0.5pt, line cap=round, line join=round] (2.11,-0.86) -- (2.11,-0.86) -- (2.11,-0.86) -- (2.12,-0.86) -- (2.12,-0.86) -- (2.12,-0.86) -- (2.13,-0.86) -- (2.14,-0.86) -- (2.15,-0.86) -- (2.16,-0.85) -- (2.17,-0.85) -- (2.18,-0.85) -- (2.19,-0.85) -- (2.20,-0.86) -- (2.21,-0.86) -- (2.22,-0.87) -- (2.23,-0.88) -- (2.24,-0.89) -- (2.24,-0.90) -- (2.25,-0.91) -- (2.25,-0.93) -- (2.26,-0.94) -- (2.25,-0.96) -- (2.25,-0.97) -- (2.25,-0.99) -- (2.24,-1.00) -- (2.23,-1.01) -- (2.23,-1.02) -- (2.22,-1.02) -- (2.21,-1.03) -- (2.21,-1.03) -- (2.20,-1.03) -- (2.20,-1.03) -- (2.20,-1.03) -- (2.20,-1.02) -- (2.21,-1.02) -- (2.21,-1.02) -- (2.21,-1.02) -- (2.22,-1.03) -- (2.23,-1.03) -- (2.24,-1.04) -- (2.25,-1.05) -- (2.27,-1.06) -- (2.29,-1.09) -- (2.31,-1.11) -- (2.32,-1.13) -- (2.34,-1.15) -- (2.35,-1.18) -- (2.36,-1.20) -- (2.36,-1.22) -- (2.37,-1.23) -- (2.37,-1.25) -- (2.36,-1.26) -- (2.36,-1.27) -- (2.35,-1.27) -- (2.34,-1.28) -- (2.32,-1.29) -- (2.30,-1.29) -- (2.28,-1.30) -- (2.26,-1.30) -- (2.24,-1.30) -- (2.23,-1.29) -- (2.21,-1.29) -- (2.20,-1.28) -- (2.18,-1.27);
\draw[line width=0.5pt, line cap=round, line join=round] (2.48,-1.09) -- (2.48,-1.09) -- (2.48,-1.09) -- (2.48,-1.09) -- (2.49,-1.09) -- (2.49,-1.10) -- (2.50,-1.10) -- (2.51,-1.10) -- (2.53,-1.11) -- (2.55,-1.11) -- (2.57,-1.12) -- (2.60,-1.12) -- (2.62,-1.13) -- (2.64,-1.13) -- (2.66,-1.13) -- (2.68,-1.13) -- (2.69,-1.13) -- (2.70,-1.13);
\draw[line width=0.5pt, line cap=round, line join=round] (2.52,-1.24) -- (2.52,-1.24) -- (2.53,-1.24) -- (2.54,-1.25) -- (2.56,-1.25) -- (2.57,-1.25) -- (2.59,-1.26) -- (2.62,-1.26) -- (2.64,-1.26) -- (2.67,-1.26) -- (2.70,-1.26) -- (2.72,-1.26);
\draw[line width=0.5pt, line cap=round, line join=round] (2.95,-0.94) -- (2.91,-1.00) -- (2.88,-1.06) -- (2.85,-1.13) -- (2.83,-1.18) -- (2.83,-1.22) -- (2.84,-1.26) -- (2.85,-1.28) -- (2.86,-1.28) -- (2.89,-1.29) -- (2.92,-1.30) -- (2.95,-1.30) -- (2.99,-1.29) -- (3.03,-1.27) -- (3.07,-1.25) -- (3.11,-1.22) -- (3.14,-1.18) -- (3.18,-1.13) -- (3.20,-1.09) -- (3.22,-1.06) -- (3.23,-1.03) -- (3.24,-1.00) -- (3.25,-0.97) -- (3.25,-0.93) -- (3.25,-0.89) -- (3.25,-0.85);
\node[anchor=south, font=\small] at (2.29,-0.75) {Context X};
\node[anchor=north, font=\small\ttfamily] at (2.29,-1.40) {3-3=0};
\draw[line width=0.5pt, line cap=round, line join=round] (3.86,-0.85) -- (3.84,-0.86) -- (3.83,-0.86) -- (3.82,-0.86) -- (3.81,-0.87) -- (3.80,-0.88) -- (3.79,-0.89) -- (3.78,-0.90) -- (3.77,-0.91) -- (3.76,-0.93) -- (3.76,-0.94) -- (3.75,-0.96) -- (3.75,-0.97) -- (3.76,-0.98) -- (3.76,-1.00) -- (3.76,-1.01) -- (3.77,-1.02) -- (3.77,-1.03) -- (3.78,-1.04) -- (3.79,-1.04) -- (3.80,-1.04) -- (3.81,-1.04) -- (3.82,-1.04) -- (3.83,-1.04) -- (3.84,-1.03) -- (3.86,-1.02) -- (3.87,-1.01) -- (3.89,-0.99) -- (3.90,-0.98) -- (3.91,-0.97) -- (3.91,-0.96) -- (3.92,-0.95) -- (3.92,-0.94) -- (3.92,-0.93) -- (3.92,-0.93) -- (3.92,-0.92) -- (3.92,-0.92) -- (3.92,-0.91) -- (3.92,-0.91) -- (3.91,-0.91) -- (3.91,-0.91) -- (3.91,-0.91) -- (3.91,-0.91) -- (3.91,-0.91) -- (3.92,-0.92) -- (3.92,-0.92) -- (3.92,-0.93) -- (3.92,-0.94) -- (3.92,-0.96) -- (3.92,-0.98) -- (3.93,-1.00) -- (3.93,-1.03) -- (3.93,-1.06) -- (3.94,-1.10) -- (3.94,-1.13) -- (3.94,-1.15) -- (3.94,-1.17) -- (3.93,-1.19) -- (3.93,-1.21) -- (3.92,-1.23) -- (3.91,-1.24) -- (3.90,-1.26) -- (3.88,-1.28) -- (3.86,-1.29) -- (3.84,-1.30) -- (3.81,-1.30) -- (3.79,-1.29) -- (3.77,-1.28);
\draw[line width=0.5pt, line cap=round, line join=round] (4.05,-1.18) -- (4.06,-1.18) -- (4.08,-1.18) -- (4.09,-1.17) -- (4.12,-1.17) -- (4.14,-1.17) -- (4.15,-1.17) -- (4.18,-1.18) -- (4.20,-1.18) -- (4.23,-1.18) -- (4.26,-1.18) -- (4.30,-1.18);
\draw[line width=0.5pt, line cap=round, line join=round] (4.41,-0.93) -- (4.41,-0.93) -- (4.41,-0.93) -- (4.41,-0.93) -- (4.42,-0.92) -- (4.43,-0.90) -- (4.45,-0.88) -- (4.48,-0.87) -- (4.50,-0.86) -- (4.52,-0.85) -- (4.54,-0.85) -- (4.57,-0.85) -- (4.59,-0.86) -- (4.61,-0.87) -- (4.63,-0.88) -- (4.64,-0.90) -- (4.65,-0.92) -- (4.66,-0.94) -- (4.65,-0.97) -- (4.64,-1.00) -- (4.63,-1.02) -- (4.61,-1.04) -- (4.59,-1.05) -- (4.57,-1.06) -- (4.55,-1.07) -- (4.55,-1.07) -- (4.55,-1.06) -- (4.55,-1.06) -- (4.56,-1.06) -- (4.58,-1.06) -- (4.60,-1.07) -- (4.64,-1.08) -- (4.67,-1.09) -- (4.69,-1.11) -- (4.71,-1.13) -- (4.72,-1.15) -- (4.72,-1.18) -- (4.71,-1.22) -- (4.68,-1.25) -- (4.66,-1.27) -- (4.62,-1.29) -- (4.59,-1.30) -- (4.55,-1.30) -- (4.51,-1.29);
\draw[line width=0.5pt, line cap=round, line join=round] (4.83,-1.09) -- (4.83,-1.09) -- (4.83,-1.09) -- (4.83,-1.09) -- (4.84,-1.09) -- (4.84,-1.10) -- (4.85,-1.10) -- (4.86,-1.10) -- (4.88,-1.11) -- (4.90,-1.11) -- (4.92,-1.12) -- (4.95,-1.12) -- (4.97,-1.13) -- (4.99,-1.13) -- (5.01,-1.13) -- (5.03,-1.13) -- (5.04,-1.13) -- (5.05,-1.13);
\draw[line width=0.5pt, line cap=round, line join=round] (4.87,-1.24) -- (4.87,-1.24) -- (4.88,-1.24) -- (4.89,-1.25) -- (4.91,-1.25) -- (4.92,-1.25) -- (4.94,-1.26) -- (4.97,-1.26) -- (4.99,-1.26) -- (5.02,-1.26) -- (5.05,-1.26) -- (5.07,-1.26);
\draw[line width=0.5pt, line cap=round, line join=round] (5.30,-0.94) -- (5.26,-1.00) -- (5.23,-1.06) -- (5.20,-1.13) -- (5.18,-1.18) -- (5.18,-1.22) -- (5.19,-1.26) -- (5.20,-1.28) -- (5.21,-1.28) -- (5.24,-1.29) -- (5.27,-1.30) -- (5.30,-1.30) -- (5.34,-1.29) -- (5.38,-1.27) -- (5.42,-1.25) -- (5.46,-1.22) -- (5.49,-1.18) -- (5.53,-1.13) -- (5.55,-1.09) -- (5.57,-1.06) -- (5.58,-1.03) -- (5.59,-1.00) -- (5.60,-0.97) -- (5.60,-0.93) -- (5.60,-0.89) -- (5.60,-0.85);
\node[anchor=south, font=\small] at (4.64,-0.75) {Context Y};
\node[anchor=north, font=\small\ttfamily] at (4.64,-1.40) {9-3=6};
\draw[line width=0.5pt, line cap=round, line join=round] (6.41,-0.86) -- (6.40,-0.86) -- (6.40,-0.86) -- (6.40,-0.86) -- (6.39,-0.85) -- (6.37,-0.85) -- (6.36,-0.85) -- (6.34,-0.85) -- (6.32,-0.85) -- (6.29,-0.85) -- (6.26,-0.86) -- (6.24,-0.86) -- (6.22,-0.86) -- (6.21,-0.87) -- (6.20,-0.87) -- (6.19,-0.87) -- (6.19,-0.87) -- (6.19,-0.87) -- (6.20,-0.87) -- (6.20,-0.87) -- (6.20,-0.88) -- (6.19,-0.88) -- (6.19,-0.89) -- (6.18,-0.92) -- (6.17,-0.96) -- (6.16,-0.98) -- (6.16,-0.99) -- (6.17,-1.00) -- (6.18,-1.02) -- (6.21,-1.03) -- (6.25,-1.05) -- (6.30,-1.07) -- (6.34,-1.10) -- (6.37,-1.12) -- (6.38,-1.15) -- (6.39,-1.18) -- (6.38,-1.21) -- (6.35,-1.25) -- (6.31,-1.28) -- (6.25,-1.30) -- (6.19,-1.30);
\draw[line width=0.5pt, line cap=round, line join=round] (6.52,-1.18) -- (6.53,-1.18) -- (6.54,-1.18) -- (6.56,-1.17) -- (6.58,-1.17) -- (6.60,-1.17) -- (6.62,-1.17) -- (6.65,-1.18) -- (6.67,-1.18) -- (6.70,-1.18) -- (6.73,-1.18) -- (6.76,-1.18);
\draw[line width=0.5pt, line cap=round, line join=round] (6.97,-0.85) -- (6.97,-0.85) -- (6.97,-0.85) -- (6.97,-0.85) -- (6.97,-0.85) -- (6.97,-0.86) -- (6.97,-0.86) -- (6.96,-0.86) -- (6.96,-0.86) -- (6.96,-0.86) -- (6.96,-0.87) -- (6.96,-0.87) -- (6.96,-0.88) -- (6.95,-0.89) -- (6.95,-0.90) -- (6.95,-0.91) -- (6.94,-0.93) -- (6.93,-0.94) -- (6.92,-0.96) -- (6.92,-0.98) -- (6.91,-1.01) -- (6.90,-1.04) -- (6.89,-1.07) -- (6.89,-1.09) -- (6.88,-1.11) -- (6.88,-1.13) -- (6.88,-1.14) -- (6.87,-1.15) -- (6.87,-1.16) -- (6.88,-1.16) -- (6.88,-1.17) -- (6.88,-1.17) -- (6.89,-1.17) -- (6.90,-1.17) -- (6.91,-1.17) -- (6.92,-1.17) -- (6.93,-1.17) -- (6.94,-1.17) -- (6.96,-1.17) -- (6.97,-1.17) -- (6.98,-1.16) -- (6.99,-1.16) -- (7.01,-1.15) -- (7.02,-1.15) -- (7.03,-1.14) -- (7.03,-1.14) -- (7.04,-1.13) -- (7.05,-1.12) -- (7.06,-1.12) -- (7.07,-1.11);
\draw[line width=0.5pt, line cap=round, line join=round] (6.97,-1.05) -- (6.97,-1.07) -- (6.97,-1.08) -- (6.96,-1.10) -- (6.96,-1.13) -- (6.96,-1.15) -- (6.96,-1.17) -- (6.96,-1.20) -- (6.95,-1.22) -- (6.95,-1.23) -- (6.95,-1.25) -- (6.96,-1.27) -- (6.96,-1.28) -- (6.96,-1.29) -- (6.96,-1.29) -- (6.97,-1.30);
\draw[line width=0.5pt, line cap=round, line join=round] (7.18,-1.09) -- (7.18,-1.09) -- (7.18,-1.09) -- (7.18,-1.09) -- (7.19,-1.09) -- (7.19,-1.10) -- (7.20,-1.10) -- (7.21,-1.10) -- (7.23,-1.11) -- (7.25,-1.11) -- (7.27,-1.12) -- (7.30,-1.12) -- (7.32,-1.13) -- (7.34,-1.13) -- (7.36,-1.13) -- (7.38,-1.13) -- (7.39,-1.13) -- (7.40,-1.13);
\draw[line width=0.5pt, line cap=round, line join=round] (7.22,-1.24) -- (7.22,-1.24) -- (7.23,-1.24) -- (7.24,-1.25) -- (7.26,-1.25) -- (7.27,-1.25) -- (7.29,-1.26) -- (7.32,-1.26) -- (7.34,-1.26) -- (7.37,-1.26) -- (7.40,-1.26) -- (7.42,-1.26);
\draw[line width=0.5pt, line cap=round, line join=round] (7.65,-0.94) -- (7.61,-1.00) -- (7.58,-1.06) -- (7.55,-1.13) -- (7.53,-1.18) -- (7.53,-1.22) -- (7.54,-1.26) -- (7.55,-1.28) -- (7.56,-1.28) -- (7.59,-1.29) -- (7.62,-1.30) -- (7.65,-1.30) -- (7.69,-1.29) -- (7.73,-1.27) -- (7.77,-1.25) -- (7.81,-1.22) -- (7.84,-1.18) -- (7.88,-1.13) -- (7.90,-1.09) -- (7.92,-1.06) -- (7.93,-1.03) -- (7.94,-1.00) -- (7.95,-0.97) -- (7.95,-0.93) -- (7.95,-0.89) -- (7.95,-0.85);
\node[anchor=south, font=\small] at (6.99,-0.75) {Context N};
\node[anchor=north, font=\small\ttfamily] at (6.99,-1.40) {5-4=0};
\end{tikzpicture}

  \caption{One glyph in four images, read by Gemma-4-12B.}
  \label{fig:example}
\end{figure}
In the same-glyph test, the target glyph sits at the last position of the result (Fig.~\ref{fig:example}).
Its alternative $y$ is the digit that the VLMs most often confuse with $d$ in the ink of the design writers, so we rebuilt the same items with $y$ drawn at random from the other digits (post hoc); the effect remains above zero in \srPassed{} of \srVLMs{} VLMs (\srDeltaMin{}--\srDeltaMax{} points, Holm-corrected bootstrap) and is \srRatio{} times its size with the most-confused $y$.
The target region is pixel-identical across contexts, and the \emph{context effect} is the percentage of readings that write $y$ in context Y minus that in context X.
For the \testThreeNItems{} glyphs from \testThreeNWriters{} test writers, \testThreePassed{} of 10 VLMs write $y$ more often in context Y than in context X, by \testThreeDeltaPassedMin--\testThreeDeltaPassedMax{} points, whereas the recognizers show almost no difference (last row of Table~\ref{tab:phenomenon}).
On clear glyphs the effect is \testThreeCropDeltaMin--\testThreeCropDeltaMax{} points, so it is absent for some readers.
When asked only for the digit at the target position (the \emph{position question}), the effect is much smaller (\testThreePosMin{} to \testThreePosMax{} points, lower bound above 0 for \testThreePosAbove{} of 10), although \testThreeInternPosUnparsedMin--\testThreeInternPosUnparsedMax\% of the answers of InternVL3.5-38B to this question cannot be parsed; restricted to parsed answers, the effect is \posParsedMin{} to \posParsedMax{} points (post hoc).
As the two questions also differ in the form of the answer, we asked for both in one reply, first the digit at the target position and then the whole equation (three readings per image, post hoc).
Within the same replies, the context effect is \fmDigitMin{} to \fmDigitMax{} points on the digit line and \fmEqMin{} to \fmEqMax{} points on the equation line, with a lower bound above 0 for \fmEqAbove{} of 10 VLMs on the equation line and for \fmDigitAbove{} on the digit line; in context Y, up to \fmSplitMax\% of the replies give $d$ as the digit and $y$ in the equation, against at most \fmReverseMax\% the other way round.
When the equation is asked for first, the digit line follows it, with \fmAfterDigitMin{} to \fmAfterDigitMax{} points.
The repair thus arises mostly while the equation is being written, with a small effect on the reported digit in some VLMs; GLM-4.6V-Flash leaves unparsed \fmGlmUnparsed\% of the equations of these replies.
An alternative account is that the reader, as an ideal observer would, combines uncertain visual evidence with an arithmetic prior; it predicts repairs mainly on ambiguous glyphs, whereas repairs remain on legible equations and on glyphs read as written alone (see Control 2) and in replies that report the written digit, but an integration acting only while the equation is generated cannot be excluded.

\begin{figure}[tb]
  \centering
  \includegraphics[width=\columnwidth]{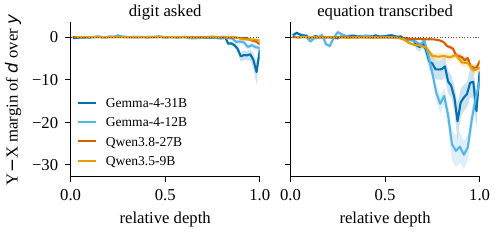}
  \caption{Logit-lens context effect on the target digit by relative depth (95\% writer intervals).}
  \label{fig:lens}
\end{figure}

\paragraph{Control 4: typed text and layers.}
\label{sec:layers}
Were repair a habit of the language model when it copies false arithmetic, it would also occur without an image.
Given the \transHeldPairs{} held-out pairs as typed \LaTeX{} with the instruction to copy them exactly (three readings, post hoc), no VLM writes the true equation for more than \toRepairMax\% of the false equations (upper bounds at most \toRepairHiMax\%; \toRepairZero{} of 10 VLMs never), no true equation is changed into its false mate, and \toCopyMin--\toCopyMax\% of the copies are exact.
Repairs land on the result of the written operation rather than on any other familiar arithmetic fact: on false equations with a changed result digit in the confusion-null test, the readings that keep the written operands land on the true result \atTrueO{} times where the confusion null predicts \atTrueE{} (summed over the 10 VLMs), on the result of a different operator \atSwapO{} times (\atSwapE{} predicted), and \atNearO{} times on the result with one operand off by 1 (\atNearE{} predicted; post hoc).
We applied the logit lens~\cite{nostalgebraist2020logitlens}, which projects the hidden state of each layer to the output vocabulary, to the same-glyph images of four VLMs to see in which layers the context acts (post hoc).
For transcription, we force the equation up to the digit at the target and take the margin $\log p(d)-\log p(y)$ of the next token in each layer.
The lens reads the digit only in late layers: the top digit is the written one for 80\% of the X images only from \lensReadableMin--\lensReadableMax\% of the depth on, so earlier layers say nothing about the context (Fig.~\ref{fig:lens}).
In the last layer, context Y lowers the margin by \lensEqMin--\lensEqMax{} when the equation is transcribed, \lensRatioMin--\lensRatioMax{} times more than when only the digit is asked (\lensDigitMin--\lensDigitMax), and $y$ outranks $d$ for \lensFlipYMin--\lensFlipYMax\% of the Y images against at most \lensFlipXMax\% of the X images.
An attention knockout~\cite{geva2023dissecting} shows that the override depends on the written operands (post hoc): when positions from ``='' onward cannot attend to the operand digits, $y$ outranks $d$ in \koForcedQ\% instead of \koForcedBaseQ\% of the Y images for Qwen3.5-9B (late layers) and \koForcedG\% instead of \koForcedBaseG\% for Gemma-4-12B (all layers); blocking the same number of instruction tokens does not (\koForcedCtrlQ\%, \koForcedCtrlG\%), and neither does subtracting the mean Y$-$X hidden state at a single late layer (\stBestMin--\stBestMax\% against \stBaseMin--\stBaseMax\%).
With all full-attention layers, chosen before this post hoc run, it lowers this share in all \kfVLMs{} VLMs we could load (\kfNoneMin--\kfNoneMax\% to \kfFlowMin--\kfFlowMax\%, bootstrap intervals below 0), whereas blocking as many random tokens shifts it by at most \kfRandomMaxShift{} points.
In free transcription the knockout lowers exact repairs of the Y images from \koGenBaseQ\% to \koGenQ\% and from \koGenBaseG\% to \koGenG\%.
On the held-out pairs and with the same layers (\khfVLMs{} VLMs), repairs fall from \khfRepairBaseMin--\khfRepairBaseMax\% to \khfRepairMin--\khfRepairMax\% (interval below 0 for \khfRepairLower{}), and exact transcription of true equations drops by more than 5 points for \khfExactDropFive{}; $Q$ is \khfQMin--\khfQMax{}, but its interval lies below 1 for only \khfQResolved{}, since the null falls too, and \khfPassed{} still meet the criterion.

\paragraph{What the rates mean in use.}
A repair yields a true and well-formed equation that no consistency check and no grader relying on the transcription can detect, unlike other reading errors.
Each repair turns a wrong answer into one that is graded correct; the repairs in the confusion-null test are \legRateMin--\legRateMax\% of the readings of false equations, small relative to all readings but large relative to each reader's confusions, and they are silent false acceptances on clean renderings (on the held-out pairs, \hsFalseMin{} to \hsFalseMax{} of 1{,}000 answers if a fifth of them are wrong), while the recognizers produce \revRecRepairs{} of them in \revRecReadings{} readings.
A count of corrections made without a reference, as used in earlier work~\cite{seong2026fix}, mixes how often a reader misreads with where these misreadings land: in the confusion-null test, GLM-4.6V-Flash repairs only \rawGlm\% of its readings of false equations but \rawGlmRatio{} times as often as its confusions predict, whereas Qwen3.5-9B repairs \rawQwenVL\% at an $O/E$ of \rawQwenVLRatio, and the rank correlation between the plain rate and $O/E$ over the 10 VLMs is \rawConfRho{} in this test and \rawHeldRho{} on held-out pairs (post hoc).
Overall accuracy therefore need not reveal this failure, and a user who applies a VLM to grading or transcription should report the repair rate on false inputs and $O/E$ against the reader's own confusions next to accuracy.

\section{Boundaries and further tests}
\label{sec:further}
\label{sec:headline}
\label{sec:mechanism}

\paragraph{Degradation, writing order and thinking.}
On held-out pairs, \degScanPassed{} of 10 VLMs meet the criterion under scan-like degradation (meaningless-operator null), fewer than the \degMinimum{} required, and under blur and under noise with JPEG, \degBlurPassed{} and \degNoisePassed{} meet it, so this test is not met.
Over the 10 VLMs $O/E$ is \snScanRatioMin--\snScanRatioMax{}, and the lower bounds are below 2 for \degScanFailNames{}.
Post hoc, the failure comes from fewer observed repairs, not from a larger $E$ (Sec.~\ref{S-sec:degradation}).
When the model must give the result before the operands, \rfPassed{} of 10 VLMs still meet the criterion (geometric mean ratio of $O/E$ \rfQGeoMean{}), so writing order does not explain the repairs.
With the thinking mode of Qwen3.8-27B switched on (post hoc, one model, \thinkPairs{} design pairs), repair persists ($O/E$ \thinkOnRatio\ci{\thinkOnRatioLo}{\thinkOnRatioHi} against \thinkOffRatio\ci{\thinkOffRatioLo}{\thinkOffRatioHi}), and the context effect on pixel-identical glyphs is \thinkOnCtx\ci{\thinkOnCtxLo}{\thinkOnCtxHi} against \thinkOffCtx\ci{\thinkOffCtxLo}{\thinkOffCtxHi} points.

\paragraph{Other tests in the supplement.}
In question answering on design writers, the criterion (\qaRegRuleMinimum{} of \qaRegRuleOf{} VLMs) was not met (\qaRegRulePassed{} passed); leakage of the computed value into reports of the written answer was repeated on held-out pairs for two VLMs named in advance (Table~\ref{S-tab:heldout}).
On the \fermatItems{} solutions of FERMAT~\cite{fermat2025} whose introduced error is one wrong integer after an equals sign (\fermatPages{} photographed pages), all \fermatVLMs{} VLMs write the arithmetically correct value in place of the written one in \fermatRepairMin{} to \fermatRepairMax{} of \fermatReadings{} readings, in line with the over-corrections reported by Seong \etal~\cite{seong2026fix}; a placebo value at the same distance from the written value is written in \fpPlaceboSum{} readings in total (none for \fpPlaceboZero{} VLMs), so the registered null is uncalibrated and its decision uninterpretable, while a confusion null cannot be estimated here (Sec.~\ref{S-sec:fermat}).
Column layouts (\colPassed{} of 10 VLMs) and a scale set (\sclPassed{} of 2) show the effect, inequalities made false by one digit do not ($O/E$ \ineqTruthMin--\ineqTruthMax; Table~\ref{S-tab:col}), and Table~\ref{S-tab:tests} lists further tests.

\section{Remedies and their cost}
\label{sec:remedies}

A remedy is useful only if it reduces repair without harming the reading of correct equations.
We say that a remedy brings a VLM \emph{below the criterion} when the VLM fails the criterion under the remedy and the upper bound of $Q$ is below 0.5 (with against without the remedy); this does not show that repair is absent.
We say that a remedy \emph{keeps accuracy} for a VLM when its exact transcription of true equations falls by at most 5 points; this does not cover other expressions.
Table~\ref{tab:remedies} summarizes four remedies and the post hoc operand knockout of Sec.~\ref{sec:layers}.

\begin{table}[tb]
  \centering
  \small
  \setlength{\tabcolsep}{3pt}
  \begin{tabular}{lcccc}
    \toprule
    Remedy & VLMs & below crit. & keeps acc. & both \\
    \midrule
    Instruction to copy & 10 & \vbRemoved & \the\numexpr10-\vbExactTDropFive\relax & \vbBoth \\
    Position questions & 10 & \swRemoved & \swPractical & \swBoth \\
    Symbol crops & \cropVLMs & \cropRemoved & \cropPractical & \cropBoth \\
    LoRA, plain & 3 & \loraPlainRemoved & \loraPlainPractical & \loraPlainBoth \\
    LoRA, mixed & 3 & \loraMixedRemoved & \loraMixedPractical & \loraMixedBoth \\
    Operand knockout & \khfVLMs & \khfRemoved & \khfPractical & \khfBoth \\
    \bottomrule
  \end{tabular}
  \caption{Outcome of each remedy (number of VLMs).}
  \label{tab:remedies}
\end{table}

\paragraph{Instruction and questions.}
An instruction to transcribe exactly what is written even if the equation is wrong lowers repairs in all \vbRead{} VLMs (from \vbRepairStdMin--\vbRepairStdMax\% to \vbRepairMin--\vbRepairMax\% of readings of false equations), but \vbPassed{} of 10 still meet the criterion (geometric mean $Q$ \vbQGeoMean), and \vbExactTDropFive{} lose more than 5 points of exact transcription of true equations; of the \vbRemoved{} VLMs brought below it, GLM-4.6V-Flash failed it also without the instruction.
Asking for the equation symbol by symbol, one position per question, brings \swRemoved{} VLM below the criterion, but the VLMs lose track of positions and exact transcription of true equations reaches at most \swExactTMax\%.

\paragraph{Symbol crops.}
When each symbol is cropped and read alone, no arithmetic context reaches the reader, and \cropRemoved{} of the \cropVLMs{} VLMs fall below the criterion (geometric mean $Q$ \cropQGeo).
The same step removes the context that helps to read ambiguous glyphs, and exact transcription of true equations falls by \cropExactDropMin--\cropExactDropMax{} points where $Q$ is defined, so the remedy keeps accuracy for \cropPractical{} VLMs.

\paragraph{Fine-tuning.}
We trained LoRA adapters (rank 16, a single epoch) for Qwen3.5-9B, Gemma-4-12B and Qwen3.8-27B on 3,000 false and true equations from the ink of CROHME 2019 training writers, disjoint from the test writers, with the written symbols as targets.
This plain adapter brings \loraPlainRemoved{} of the 3 VLMs below the criterion (Gemma-4-12B, $Q$ \loraPlainGemmaTwelveQ\ci{\loraPlainGemmaTwelveQLo}{\loraPlainGemmaTwelveQHi}), and keeps accuracy on true equations in all 3, but exact transcription of general CROHME 2019 expressions changes by \loraPlainGemmaTwelveGenDiff{} to \loraPlainQwenNineGenDiff{} points (post hoc; Table~\ref{S-tab:lora}).
With 3,000 general CROHME 2019 training expressions mixed in, \loraMixedPassed{} of 3 VLMs meet the criterion, yet \loraMixedRemoved{} is brought below it.
The repair rate changes by \loraMixedQwenTwentySevenRepairDiff{} to \loraMixedGemmaTwelveRepairDiff{} points and exact transcription of true equations by \loraMixedGemmaTwelveExactDiff{} to \loraMixedQwenTwentySevenExactDiff{} points, general transcription rises, and the reduction transfers to the unseen operand position in \loraMixedTransfer{} of 3 VLMs.

\paragraph{Trade-off.}
Removing the context reduces repair together with accuracy, while an instruction and the knockout cost little but leave most VLMs above the criterion, so a remedy must be judged on false and true equations together.

\section{Limitations and conclusion}
\label{sec:limits}
\label{sec:audits}

An automatic check finds no mismatched or reused glyph; none of \sensDecisions{} per-model decisions changes without pairs of identical ink, with a permissive parser, without extraction rules or with unrounded $p$ (post hoc).
Each stimulus is a rendering of a single-operation equation on a regular layout and the test fails its minimum under scan-like degradation; the rates are therefore not those of real grading.
The readers are 10 open VLMs with 4B to 38B parameters and three commercial models; the human baseline consists of two authors on \hcFalse{} false equations, and student work remains untested.
The confusion null was selected after the first test failed.
The logit lens is correlational, the knockout (\kfVLMs{} VLMs) does not show how the operands are read from the image, and repairs are a small share of all readings.
In sum, on clean renderings most of 10 VLMs repair written digits beyond their own confusions, also on held-out pairs; the same pixels are read as written when only the digit is asked; blocking attention to the written operands lowers repair (post hoc); no remedy brings more than one VLM below the criterion without losing accuracy.

\section*{LLM Usage Statement}
This work used LLM-based assistants.
Claude was used to write and revise the code generating the stimuli, running the evaluated models and computing the statistics.
It was also used to survey the related work, to draft and edit the text and to produce simulated reviews of drafts.
ChatGPT was used for ideation, for surveying related work and to produce simulated reviews of drafts.
The evaluated models included GPT-6 Luna, GPT-6.1 Sol and Claude Opus 5.5, each read through its API by the released code.
All reported results come from the released code applied to the outputs of the evaluated models, the open VLMs served locally and the three commercial models; no LLM outside the evaluated models produced, edited or selected any reading, score or reported number.
The authors checked all code, analyses and text.

{
    \small
\providecommand{\url}[1]{\texttt{#1}}

}

\clearpage
\setcounter{section}{0}\setcounter{table}{0}\setcounter{figure}{0}\setcounter{footnote}{0}
\renewcommand{\theHsection}{S\arabic{section}}\renewcommand{\theHtable}{S\arabic{table}}\renewcommand{\theHfigure}{S\arabic{figure}}
\twocolumn[{\centering\Large\bfseries DiploMat Supplementary Material\par\vspace{1.5em}}]
\renewcommand{\thetable}{S\arabic{table}}
\renewcommand{\thefigure}{S\arabic{figure}}
\providecommand{\ci}{}
\renewcommand{\ci}[2]{\,[#1,\,#2]}
\providecommand{\prompt}{}
\renewcommand{\prompt}[1]{\begin{quote}\small ``#1''\end{quote}}

\appendix
\renewcommand{\thesection}{S\arabic{section}}

This supplement adds detail to the main paper and is not required to follow it.
Section numbers from the main paper are written as ``main Sec.''; tables and sections of this supplement carry the prefix S.
All numbers are the stored results of the run reported in the main paper.

\section{Pre-registration record}
\label{S-sec:prereg}

\paragraph{What was fixed in advance.}
For each test, a document under version control stated the question, the stimuli, the readers, the measures, the decision rule and the predictions of the authors before the confirmatory readings of that test were collected.
The fixed elements are the confusion null and its smoothing count of 0.1, the meaningless-operator null of the meaningless-operator test, the criterion that the 95\% interval of $O/E$ has its lower bound above 2 and a Holm-adjusted one-sided $p$ is below 0.05, the weaker criterion of a lower bound above 0 for both excesses and differences, the writer bootstrap with 10{,}000 replicates, and the estimation targets, which are reported with intervals and no decision.
For the white-box tests, the documents also fixed the candidate layer bands, the share of the X--Y difference that a patch must remove (one half), and the share of repaired readings in which the logit lens must rank $d$ above $y$ (0.6).
The documents and the analysis code are part of the release.

\paragraph{What changed afterwards.}
Decision rules were written during an earlier run of five VLMs on the earlier stimuli; these five included Gemma-4-31B, InternVL3.5-38B and Mistral-Small-3.2, so the decisions for these three models are not independent of the rules, although every reading reported in the paper comes from the present run.
The 10 VLMs were fixed before any of their readings in the present run were collected.
Each document named a Holm family of five models (four in the question-answering test); the adjustment now covers the models with a testable estimate in each test, namely 10 in the transcription tests and \qaDissTested{} in the question-answering test.
Each document also required a minimum number of the five models to meet the criterion; because the model set was enlarged, this requirement is not applied, and the paper reports the number of models that meet each criterion.
All the stimuli were regenerated after the rules were fixed, to remove the two generation defects described in Sec.~\ref{S-sec:stimuli}.
The regenerated stimuli were read first in the meaningless-operator test and the question-answering test, and in the other tests afterwards with the same models and settings.
The regenerated stimuli were read once with every model, all their results are reported, and no reading of the earlier stimuli enters the paper.
The confusion-null metric itself was adopted after the failure of an earlier pre-registered test that compared repairs of false equations with misreadings of true equations.
Table~\ref{S-tab:deviations} summarizes these changes.

\begin{table}[tb]
  \centering
  \footnotesize
  \setlength{\tabcolsep}{2pt}
  \begin{tabular}{@{}>{\raggedright\arraybackslash}p{0.2\columnwidth}>{\raggedright\arraybackslash}p{0.36\columnwidth}>{\raggedright\arraybackslash}p{0.4\columnwidth}@{}}
    \toprule
    item & registered & present paper \\
    \midrule
    decision rules & fixed per test & unchanged \\
    models & five VLMs of an earlier run & 10 VLMs fixed before their readings; three overlap with the earlier run \\
    Holm family & the five models & models with a testable estimate in each test \\
    passing models & a minimum number of the five & not applied; counts reported \\
    stimuli & earlier generation & all regenerated after the rules; meaningless-operator and question-answering tests read first \\
    copy-only Q1 & not registered & three paraphrases added post hoc, after the other results were known \\
    held-out pairs & not in the earlier registrations & question answering repeated on test writers under a rule registered before its readings, including the copy-only question \\
    held-out transcription & not in the earlier registrations & meaningless-operator test repeated on the same held-out pairs under a rule registered before their transcriptions \\
    instruction to copy & not in the earlier registrations & held-out test repeated with an instruction to transcribe as written, registered before its readings \\
    commercial models & not in the earlier registrations & held-out test applied to two commercial models, registered before their readings; models and settings fixed in an addendum before scoring \\
    metric & confusion null & adopted after an earlier registered test failed \\
    main effect size & restricted ratio (dissociation) & unrestricted ratio, chosen in revision after the results were known \\
    \bottomrule
  \end{tabular}
  \caption{Deviations from the pre-registrations.}
  \label{S-tab:deviations}
\end{table}

\paragraph{Post hoc analyses.}
The following analyses were chosen after the data were seen and are labeled post hoc in the main paper: the null estimated from Q1 answers, the copy-only paraphrases of Q1 on the design writers, the restriction of leakage to Q2 alone, the masked-image interpretation of Q1, the sweep of the smoothing count, the power of the recognizer comparison, the slopes over the size of the error, the equivalence margin for the patching shares, and all variants of the sensitivity analyses (main Sec.~\ref{sec:audits}).
No correction for multiple comparisons is applied across tests or to post hoc analyses.

\paragraph{Registered decision rule.}
A Holm adjustment over 8 tests is the registered rule of the question-answering test, namely over the Q1 leakage and the restricted leakage of each of Gemma-4-31B, InternVL3.5-38B, Qwen3-VL-32B and Mistral-Small-3.2, with a test passing if its adjusted $p$ is at most 0.05 and the lower bound of $O/E$ is above 2; the dissociation is supported if the restricted leakage passes for at least \qaRegRuleMinimum{} of the \qaRegRuleOf{} VLMs of the earlier run.
Qwen3-VL-32B and Qwen3.6-27B are not among the present readers; the family includes both of the tests of Qwen3-VL-32B with $p=1$, and Qwen3.6-27B, which is an estimation target outside the family, counts as failing.
Applied post hoc to the present readings, the rule is \qaRegRuleMet: \qaRegRulePassed{} models pass (\qaRegRuleNames).

\paragraph{Leave-one-writer-out.}
If a single writer is left out, the $O/E$ (post hoc, point estimates) of the restricted leakage ranges over \qaLowoMinGemmaTwelve--\qaLowoMaxGemmaTwelve{} for Gemma-4-12B, over \qaLowoMinIntern--\qaLowoMaxIntern{} for InternVL3.5-38B, over \qaLowoMinMistral--\qaLowoMaxMistral{} for Mistral-Small-3.2 and over \qaLowoMinMolmo--\qaLowoMaxMolmo{} for Molmo2-O-7B.
These models draw $O$ from \qaLowoWritersGemmaTwelve, \qaLowoWritersIntern, \qaLowoWritersMistral{} and \qaLowoWritersMolmo{} writers, and the largest share of $O$ from one writer is \qaLowoShareGemmaTwelve\%, \qaLowoShareIntern\%, \qaLowoShareMistral\% and \qaLowoShareMolmo\%.

\paragraph{A selection-matched null.}
Re-estimating the confusions from Q1 answers to true equations in readings with Q2 equal to the result and with Q3 ``yes'', \qaSelPassed{} VLMs meet the criterion; this variant is not used in the main paper because the selected readings contain few errors, which biases $E$ downward.

\section{Stimulus construction}
\label{S-sec:stimuli}

\paragraph{Ink and glyphs.}
The ink, taken from the CROHME 2023 release~\cite{S-crohme2023}, is the online handwriting of the CROHME test expressions with its symbol segmentation.
Each segmented symbol of a writer is a \emph{glyph instance}, labeled with its symbol class.
An equation $a\circ b=c$ is composed by placing glyph instances of one writer from left to right on a common baseline and rendering the strokes at $448{\times}448$ pixels.
A writer is eligible for an equation only when it has enough instances of every label that the false and the true equation need.

\paragraph{Position-wise assignment.}
Glyph instances are assigned by position.
Every position whose label is the same in the false and in the true equation receives the same instance in both, so that the two images differ only at the changed digit.
No instance is used at two different positions of a pair.
The meaningless-operator view replaces the operator by a handwritten triangle, taken from the same writer when that writer wrote one and otherwise from another eligible writer, which was the case for \sixDeltaOther{} of the \sixNStarts{} equations.

\paragraph{Defects of the earlier stimuli.}
The earlier generator drew instances for each equation separately.
Then an unchanged position could use a different instance in the false and in the true equation (glyph mismatch), and one instance could appear at two positions of an equation (glyph reuse).
The earlier generator also drew ink from the CROHME 2014 test set, which the regenerated stimuli do not use.
The regenerated sets contain neither defect, which was checked automatically on every set, and all results of the paper come from them.

\paragraph{Writers and sets.}
Design writers provided the ink of the exploratory pilot runs made before the pre-registrations, and test writers were held out from these runs.
Table~\ref{S-tab:sets} lists the sets, their writers and the tests that use each set.
The writers of the design-writer pairs are disjoint from those of the confusion-null and same-glyph tests and from the design items of the white-box tests.
All sets come from the CROHME 2016, 2019 and 2023 test sets, and all \sixNWriters{} writers of the design-writer pairs are design writers.

\begin{table}[tb]
  \centering
  \footnotesize
  \setlength{\tabcolsep}{3pt}
  \begin{tabular}{@{}lll@{}}
    \toprule
    set & size & tests \\
    \midrule
    confusion-null pairs & \testTwoNPairs{} pairs & confusion null \\
    same-glyph items & \testThreeNItems{} glyphs & same glyph, white box \\
    column pairs & \colPairs{} pairs & column arithmetic \\
    inequality pairs & \ineqPairs{} pairs & inequalities \\
    design-writer pairs & \sixNPairs{} pairs & meaningless operator, questions \\
    held-out pairs & \qaHeldPairs{} pairs & confirmatory test, degradation, result first, remedies, commercial VLMs, questions \\
    \bottomrule
  \end{tabular}
  \caption{Stimulus sets and the tests that use them.}
  \label{S-tab:sets}
\end{table}

The same-glyph items come from \testThreeNWriters{} test writers.
The scale test uses its own set of same-glyph items, built in the same way.
The white-box tests use the same-glyph items of test writers as test items and separate items of design writers for the choice of the layer band and for training the probe.

\paragraph{False equations.}
Each false equation differs from its true equation in one digit of the result or of an operand by a step drawn uniformly from $\pm1$, $\pm2$ and $\pm3$, and every number keeps its number of digits.
The question-answering test decides on the \qaNFalse{} false equations in which the changed digit lies in the result.
In the same-glyph test, the target glyph is the last digit of the result, and context X and context Y differ only in the operands, so the target region is pixel-identical across contexts.
The answer-masked image of the question-answering test is the same image with the result digits removed.

\section{Prompts and scoring}
\label{S-sec:prompts}

\paragraph{Transcription.}
Every transcription test except the column and result-first tests uses one prompt.
\prompt{The image shows a handwritten mathematical expression. Transcribe it into LaTeX. Output only the LaTeX code, without \$ delimiters or explanations.}
The result-first transcription test (main Sec.~\ref{sec:further}) uses the following prompt.
\prompt{The image shows a handwritten mathematical expression. Transcribe it into LaTeX, but write the right-hand side of the equals sign first, then the equals sign, then the left-hand side. For example, an image showing a + b = c is written as c = a + b. Output only the LaTeX code, without \$ delimiters or explanations.}
The column test uses the following prompt.
\prompt{The image shows handwritten arithmetic written in column form: numbers stacked with an operator, a horizontal line, and the result below the line. Transcribe it line by line from top to bottom, one line per row, writing the operator (+, -, \textbackslash times) where it is written and skipping the horizontal line. Output only those lines, without \$ delimiters or explanations.}
The position question of the same-glyph test and the reading of an isolated crop use the following prompts, where $k$ is the position of the target.
\prompt{The image shows a handwritten equation. What is the digit at position $k$ in this equation, counting every symbol from the left starting at 1 (it is the last symbol)? Answer with that single digit only.}
\prompt{The image shows one handwritten digit. Which digit from 0 to 9 is it? Answer with that single digit only.}
The free-generation test (Sec.~\ref{S-sec:whitebox}) uses the following prompt with three steps.
\prompt{The image shows a handwritten equation. Transcribe it in three steps. First, on a line starting with ``Right:'', write the right-hand side (everything after the equals sign) in LaTeX. Second, on a line starting with ``Left:'', write the left-hand side (everything before the equals sign) in LaTeX. Third, on a line starting with ``Equation:'', write the whole equation in LaTeX. Output only these three lines, without \$ delimiters.}

\paragraph{Question answering.}
Every prompt with an image begins with ``The image shows a student's handwritten equation.'' and continues with one of the paraphrases below, where the place in Q4 is the units, tens or hundreds place.
\begin{itemize}\setlength{\itemsep}{1pt}\small
  \item Q1: ``What number did the student write as the answer, to the right of the equals sign? Answer with the number only.''; ``What number is written to the right of the equals sign? Answer with the number only.''; ``Read the student's answer, the number after the equals sign. Answer with the number only.''
  \item Q2: ``What is the correct result of the calculation to the left of the equals sign? Answer with the number only.''; ``Compute the left-hand side of this equation. Answer with the number only.''; ``What should the result of this calculation be? Answer with the number only.''
  \item Q3: ``Is the student's answer correct? Answer yes or no.''; ``Is this equation true? Answer yes or no.''; ``Did the student get the right result? Answer yes or no.''
  \item Q4: ``What is the [place] digit of the number to the right of the equals sign? Answer with that single digit only.''; ``Look at the number after the equals sign. Which digit is in its [place] place? Answer with that single digit only.''; ``In the student's answer (after the equals sign), what is the [place] digit? Answer with that single digit only.''
  \item Q1, copy only (post hoc): ``Copy the number written to the right of the equals sign exactly as it is written, even if it is not the correct result. Answer with the number only.''; ``Ignore whether the equation is correct. Which number is written after the equals sign? Answer with the number only.''; ``Transcribe the student's answer after the equals sign digit by digit as it appears, without correcting it. Answer with the number only.''
  \item Q5: ``How much larger is the student's written answer than the correct result? Answer with a single number (negative if smaller, 0 if equal).''; ``By how much does the student's answer exceed the correct result? Answer with a single number (negative if it is smaller, 0 if equal).''; ``What is the student's written answer minus the correct result? Answer with a single number.''
\end{itemize}
For the answer-masked image, Q1 and Q2 are asked.
The text-only control replaces the image and the first sentence by ``A student wrote the equation [equation].'', where the equation is given as text, and asks Q1, Q2 and Q3.
Every prompt is run once at temperature 0.

\paragraph{Parsing.}
A transcription is reduced to one \LaTeX{} string by the extraction rules of main Sec.~\ref{sec:method} and is converted into numbers by removing markup and spaces and then reading the operator, the equals sign and the digit strings between them.
For a numeric answer, the scorer removes dollar signs and the wrappers \texttt{\textbackslash text}, \texttt{\textbackslash mathrm}, \texttt{\textbackslash mathbf} and \texttt{\textbackslash boxed}, joins digit groups separated by spaces or commas, and accepts the answer when exactly one digit string remains.
A yes/no answer is the first of the words ``yes'' and ``no'' among the first three words of the reply.
An answer that cannot be parsed is counted as invalid and is never counted as leakage.

\section{Statistical procedures}
\label{S-sec:stats}

\paragraph{Confusion null.}
For each reader, the confusion matrix $C$ is estimated from the aligned misreadings that this reader makes of a reference view, with 0.1 added to every cell before each row is normalized (see main Sec.~\ref{sec:method}).
For each aligned misreading of a false equation, $q$ is the probability that independent digit errors turn the written equation into the true one, given that at least one digit is misread.
$E$ is the sum of $q$ over the aligned misreadings, $O$ is the number of repairs, and $O/E$ is the ratio of observed to expected repairs.
The excess $(O-E)/n$, where $n$ is the number of aligned misreadings, is the share of misreadings that are repairs beyond the null.

\paragraph{Meaningless-operator null.}
For the meaningless-operator test, the matrix $C$ is estimated from transcriptions of the meaningless-operator view, in which the operator symbol is a triangle and the digits carry no arithmetic relation.
A misreading of this view cannot be pulled toward the true equation by the arithmetic, so the null is not lowered by a prior that suppresses misreadings of true equations.
In the question-answering test, $C$ is estimated from the result digits of the same transcriptions, and $O$ counts wrong Q1 answers to false equations that keep the digit count and equal the computed value.

\paragraph{Null from Q1 answers.}
The post hoc null of main Sec.~\ref{sec:further} instead estimates the digit confusions from the Q1 answers themselves.
The counts come from wrong Q1 answers to the true equations of result-error pairs that keep the digit count, compared digit by digit with the written result, with 0.1 added to every cell and the rows normalized.
$O$ counts wrong Q1 answers to the false equations with a result error that keep the digit count and are equal to the computed value, and $E$ sums over these wrong answers the probability that independent digit errors turn the written result into the computed one.
The intervals come from the same writer bootstrap.

\paragraph{Writer bootstrap and $p$ values.}
Each one of the 10{,}000 replicates resamples the writers with replacement, retains all readings of the drawn writers, and re-estimates $C$, $E$ and $O$; for differences between contexts (same-glyph and scale tests), the readings of each drawn writer are resampled as well.
Intervals are the 2.5 and 97.5 percentiles of the replicates.
The one-sided $p$ is $(1+k)/(1+B)$ with $k$ the number of replicates whose excess is at most zero, so the smallest attainable value is $1/(1+B)$.
The $p$ values are rounded to five decimals before the Holm adjustment; the sensitivity analyses repeat the adjustment with unrounded values.

\paragraph{Holm adjustment.}
In each single test, the $m$ models with a testable estimate are ordered by $p$, the $j$-th smallest $p$ is multiplied by $m-j+1$, and the adjusted values are made non-decreasing in this order.
The family is the set of models of one test and one criterion; the clear-glyph criterion of the question-answering test is adjusted separately from the dissociation.

\paragraph{Equivalence test.}
For the patching shares, two one-sided tests at level 0.05 declare a share equivalent to zero when its 90\% bootstrap interval lies within $\pm\tostMargin$, a margin chosen post hoc.

\paragraph{Signal detection for Q3.}
With $H$ and $F$ the rates of ``yes'' answers to true and to false equations, we compute $d'=z(H)-z(F)$ and $c=-(z(H)+z(F))/2$, where $z$ is the inverse of the standard normal distribution function~\cite{S-macmillan2005detection}.

\section{Detailed results}

\begin{table*}[tb]
  \centering
  \scriptsize
  \setlength{\tabcolsep}{4pt}
\begin{tabular}{@{}lccc@{}}
\toprule
pair & 266 & 237 & 94 \\
\midrule
written (faithful reading) & $8 5 \times 5 = 4 1 5$ & $7 8 \times 8 = 6 2 6$ & $2 6 \times 6 = 4 5 6$ \\
true equation & $8 5 \times 5 = 4 2 5$ & $7 8 \times 8 = 6 2 4$ & $2 6 \times 6 = 1 5 6$ \\
VLMs writing the true equation & 8 of 10 & 7 of 10 & 5 of 10 \\
\midrule
Gemma-4-31B & \texttt{85 \textbackslash{}times 5 = 425} & \texttt{78 \textbackslash{}times 8 = 626} & \texttt{26 \textbackslash{}times 6 = 456} \\
Gemma-4-12B & \texttt{85 \textbackslash{}times 5 = 425} & \texttt{78 \textbackslash{}times 8 = 624} & \texttt{26 \textbackslash{}times 6 = 156} \\
Qwen3.8-27B & \texttt{8 5 \textbackslash{}times 5 = 4 2 5} & \texttt{7 8 \textbackslash{}times 8 = 6 2 4} & \texttt{2 6 \textbackslash{}times 6 = 4 5 6} \\
Qwen3.5-9B & \texttt{8 5 \textbackslash{}times 5 = 4 2 5} & \texttt{7 8 \textbackslash{}times 8 = 6 2 4} & \texttt{2 6 \textbackslash{}times 6 = 1 5 6} \\
Qwen3.5-4B & \texttt{8 5 \textbackslash{}times 5 = 4 2 5} & \texttt{7 8 \textbackslash{}times 8 = 6 2 4} & \texttt{2 6 \textbackslash{}times 6 = 1 5 6} \\
InternVL3.5-38B & \texttt{85\textbackslash{}alpha5=425} & \texttt{78\textbackslash{}times8=624} & \texttt{26\textbackslash{}times6=156} \\
Mistral-Small-3.2 & \texttt{85 \textbackslash{}times 5 = 445} & \texttt{78 \textbackslash{}times 9 = 621} & \texttt{26 \textbackslash{}times 6 - 456} \\
GLM-4.6V-Flash & \texttt{85 \textbackslash{}times 5 = 425} & \texttt{78 \textbackslash{}times 8 = 626} & \texttt{26 \textbackslash{}times 6 = 156} \\
LLaVA-OneVision-2-8B & \texttt{85\textbackslash{}times5=425} & \texttt{78\textbackslash{}times8=624} & \texttt{26\textbackslash{}times6=456} \\
Molmo2-O-7B & \texttt{85 \textbackslash{}times 5 = 425} & \texttt{78 \textbackslash{}times 8 = 624} & \texttt{2 \textbackslash{}times 6 \textbackslash{}div 456} \\
\bottomrule
\end{tabular}

  \caption{Readings of the three held-out false equations that the most VLMs transcribed as the true equation.}
  \label{S-tab:examples}
\end{table*}

Table~\ref{S-tab:examples} shows readings of the held-out repetition: the three false equations that the largest number of VLMs transcribed as the true equation at temperature 0 with ties broken by pair id, chosen by a script, with each VLM's reading as extracted by the scoring code.
A reading counts as the true equation only when it matches it exactly, so a reading that restores the result but misreads the operator, such as InternVL3.5-38B's reading of the first equation, is not counted.

\label{S-sec:results}

Table~\ref{S-tab:tests} lists the registered and post hoc predictions of the transcription and question tests, including those in the tables of the main paper, with the number of models that meet them (the degradation, result-first, remedy, FERMAT and commercial-model tests appear in the main text); the main paper reports the held-out transcription test, the test against the confusion null, the meaningless-operator test and their controls, and refers here for the rest.
For question answering, the registered minimum on the count of passing models is \qaRegRuleMet{} when applied post hoc (\qaRegRulePassed{} of \qaRegRuleOf{} models pass, \qaRegRuleMinimum{} required); Table~\ref{S-tab:tests} counts the per-model criterion with the Holm family of all \qaDissTested{} testable models, so its \qaDissPassed{} of \qaDissTested{} is not a test of the registered rule.

\begin{table}[tb]
  \centering
  \footnotesize
  \setlength{\tabcolsep}{2pt}
  \begin{tabular}{@{}llr@{}}
    \toprule
    test & prediction & models \\
    \midrule
    \multicolumn{3}{@{}l}{\emph{fully confirmatory: held-out pairs}} \\
    meaningless operator & repair: $O/E>2$ & \transHeldPassed{} of \transHeldRead{} \\
    question answering & leakage, Q1 null and restricted & \qaHeldReplicated{} of 2 \\
    \multicolumn{3}{@{}l}{\emph{registered, rule unchanged}} \\
    first transcription test & earlier measure & not met \\
    meaningless operator & repair: $O/E>2$ & \sixPassedA{} of 10 \\
    same glyph & effect of Y over X above 0 & \testThreePassed{} of 10 \\
    logit lens & $d$ leads in repaired readings & \mchPOnePassed{} of 3 \\
    prefix split & digits carry more than other tokens & \pfxPassed{} of 2 \\
    free generation & other glyphs carry the effect & \bidHOnePassed{} of 2 \\
    column arithmetic & repair: $O/E>2$ & \colPassed{} of 10 \\
    column, clear glyphs & repair on clear glyphs & \colClearPassed{} of 10 \\
    scale & same-glyph effect above 0 & \sclPassed{} of 2 \\
    scale, clear glyphs & same-glyph effect on clear glyphs & \sclClearPassed{} of 2 \\
    \multicolumn{3}{@{}l}{\emph{registered after the first test had failed}} \\
    confusion null, second test & repair: $O/E>2$ & \testTwoPassed{} of 10 \\
    \multicolumn{3}{@{}l}{\emph{registered, rule changed after results}} \\
    question answering & registered rule: $\ge$\qaRegRuleMinimum{} pass & \qaRegRulePassed{} of \qaRegRuleOf{} \\
    \quad per model & dissociation: $O/E>2$ & \qaDissPassed{} of \qaDissTested{} \\
    \quad clear glyphs & leakage excess above 0 & \qaClearPassed{} of \qaClearTested{} \\
    \multicolumn{3}{@{}l}{\emph{post hoc}} \\
    question answering & leakage, Q1 null: $O/E>2$ & \qOneNullPassed{} of 10 \\
    confusion null & position-split $C$: $O/E>2$ & \posSplitPassed{} of 10 \\
    \bottomrule
  \end{tabular}
  \caption{All predictions and the models meeting them.}
  \label{S-tab:tests}
\end{table}

\begin{table*}[tb]
  \centering
  \footnotesize
  \setlength{\tabcolsep}{1.5pt}
  \begin{tabular}{@{}lcccccccc@{}}
    \toprule
    & Q1 & Q2 & Q3 $d'$ & \multicolumn{2}{c}{leakage, all readings} & \multicolumn{2}{c}{leakage, Q2 correct and Q3 ``no''} & clear glyphs \\
    \cmidrule(lr){5-6}\cmidrule(lr){7-8}
    reader & computed, \% & correct, \% & image / text & $O/E$ & $O/E$, Q1 null & computed / wrong & $O/E$ & excess, points \\
    \midrule
    \mbox{Gemma-4-31B} & \qaQOneCompGemma & \qaQTwoCompGemma & \qaQThreeDpGemma{} / \qaTextDpGemma & \qaLeakGemma\ci{\qaLeakGemmaLo}{\qaLeakGemmaHi} & \qOneNullGemma\ci{\qOneNullGemmaLo}{\qOneNullGemmaHi}$^{\dagger}$ & \qaKnowOGemma/\qaKnowErrGemma & \qaKnowGemma\ci{\qaKnowGemmaLo}{\qaKnowGemmaHi} & \qaClearExGemma\ci{\qaClearExGemmaLo}{\qaClearExGemmaHi} \\
    \mbox{Gemma-4-12B} & \qaQOneCompGemmaTwelve & \qaQTwoCompGemmaTwelve & \qaQThreeDpGemmaTwelve{} / \qaTextDpGemmaTwelve & \qaLeakGemmaTwelve\ci{\qaLeakGemmaTwelveLo}{\qaLeakGemmaTwelveHi}$^{\dagger}$ & \qOneNullGemmaTwelve\ci{\qOneNullGemmaTwelveLo}{\qOneNullGemmaTwelveHi}$^{\dagger}$ & \qaKnowOGemmaTwelve/\qaKnowErrGemmaTwelve & \qaKnowGemmaTwelve\ci{\qaKnowGemmaTwelveLo}{\qaKnowGemmaTwelveHi}$^{\checkmark}$ & \qaClearExGemmaTwelve\ci{\qaClearExGemmaTwelveLo}{\qaClearExGemmaTwelveHi}$^{\checkmark}$ \\
    Qwen3.8-27B & \qaQOneCompQwen & \qaQTwoCompQwen & \qaQThreeDpQwen{} / \qaTextDpQwen & \qaLeakQwen\ci{\qaLeakQwenLo}{\qaLeakQwenHi} & \qOneNullQwen\ci{\qOneNullQwenLo}{\qOneNullQwenHi}$^{\dagger}$ & \qaKnowOQwen/\qaKnowErrQwen & \qaKnowQwen\ci{\qaKnowQwenLo}{\qaKnowQwenHi} & \qaClearExQwen\ci{\qaClearExQwenLo}{\qaClearExQwenHi}$^{\checkmark}$ \\
    Qwen3.5-9B & \qaQOneCompQwenVL & \qaQTwoCompQwenVL & \qaQThreeDpQwenVL{} / \qaTextDpQwenVL & \qaLeakQwenVL\ci{\qaLeakQwenVLLo}{\qaLeakQwenVLHi} & \qOneNullQwenVL\ci{\qOneNullQwenVLLo}{\qOneNullQwenVLHi} & \qaKnowOQwenVL/\qaKnowErrQwenVL & \qaKnowQwenVL\ci{\qaKnowQwenVLLo}{\qaKnowQwenVLHi} & \qaClearExQwenVL\ci{\qaClearExQwenVLLo}{\qaClearExQwenVLHi}$^{\checkmark}$ \\
    Qwen3.5-4B & \qaQOneCompQwenFour & \qaQTwoCompQwenFour & \qaQThreeDpQwenFour{} / \qaTextDpQwenFour & \qaLeakQwenFour\ci{\qaLeakQwenFourLo}{\qaLeakQwenFourHi} & -- & \qaKnowOQwenFour/\qaKnowErrQwenFour & -- & -- \\
    InternVL3.5-38B & \qaQOneCompIntern & \qaQTwoCompIntern & \qaQThreeDpIntern{} / \qaTextDpIntern & \qaLeakIntern\ci{\qaLeakInternLo}{\qaLeakInternHi}$^{\dagger}$ & \qOneNullIntern\ci{\qOneNullInternLo}{\qOneNullInternHi}$^{\dagger}$ & \qaKnowOIntern/\qaKnowErrIntern & \qaKnowIntern\ci{\qaKnowInternLo}{\qaKnowInternHi}$^{\checkmark}$ & \qaClearExIntern\ci{\qaClearExInternLo}{\qaClearExInternHi}$^{\checkmark}$ \\
    Mistral-Small-3.2 & \qaQOneCompMistral & \qaQTwoCompMistral & \qaQThreeDpMistral{} / \qaTextDpMistral & \qaLeakMistral\ci{\qaLeakMistralLo}{\qaLeakMistralHi}$^{\dagger}$ & \qOneNullMistral\ci{\qOneNullMistralLo}{\qOneNullMistralHi}$^{\dagger}$ & \qaKnowOMistral/\qaKnowErrMistral & \qaKnowMistral\ci{\qaKnowMistralLo}{\qaKnowMistralHi}$^{\checkmark}$ & \qaClearExMistral\ci{\qaClearExMistralLo}{\qaClearExMistralHi}$^{\checkmark}$ \\
    GLM-4.6V-Flash & \qaQOneCompGlm & \qaQTwoCompGlm & n/a / \qaTextDpGlm & \qaLeakGlm\ci{\qaLeakGlmLo}{\qaLeakGlmHi} & -- & \qaKnowOGlm/\qaKnowErrGlm & -- & -- \\
    LLaVA-OneVision-2-8B & \qaQOneCompLlava & \qaQTwoCompLlava & \qaQThreeDpLlava{} / \qaTextDpLlava & \qaLeakLlava\ci{\qaLeakLlavaLo}{\qaLeakLlavaHi} & \qOneNullLlava\ci{\qOneNullLlavaLo}{\qOneNullLlavaHi} & \qaKnowOLlava/\qaKnowErrLlava & \qaKnowLlava\ci{\qaKnowLlavaLo}{\qaKnowLlavaHi} & \qaClearExLlava\ci{\qaClearExLlavaLo}{\qaClearExLlavaHi} \\
    Molmo2-O-7B & \qaQOneCompMolmo & \qaQTwoCompMolmo & \qaQThreeDpMolmo{} / \qaTextDpMolmo & \qaLeakMolmo\ci{\qaLeakMolmoLo}{\qaLeakMolmoHi} & \qOneNullMolmo\ci{\qOneNullMolmoLo}{\qOneNullMolmoHi} & \qaKnowOMolmo/\qaKnowErrMolmo & \qaKnowMolmo\ci{\qaKnowMolmoLo}{\qaKnowMolmoHi}$^{\checkmark}$ & \qaClearExMolmo\ci{\qaClearExMolmoLo}{\qaClearExMolmoHi}$^{\checkmark}$ \\
    \bottomrule
  \end{tabular}
  \caption{Question answering on \qaNFalse{} false result equations; $^{\checkmark}$: registered criterion met; $^{\dagger}$: met post hoc; --: not testable; n/a: no answer.}
  \label{S-tab:qa}
\end{table*}

\paragraph{Question answering.}
Table~\ref{S-tab:qaleak} gives for each VLM the unrestricted leakage, the leakage restricted to the readings with Q2 equal to the computed value (post hoc), and the leakage under the null from Q1 answers (post hoc).
The unrestricted leakage counts the wrong Q1 reports that keep the digit count, of which the first column gives the number equal to the computed value and the total.
The restriction to the readings with Q2 correct and Q3 ``no'' is reported in Table~\ref{S-tab:qa}.
The null from Q1 answers could not be estimated for Qwen3.5-4B or for GLM-4.6V-Flash.
The last column counts, under the three copy-only paraphrases of Q1 (post hoc), the wrong reports that keep the digit count and those equal to the computed value; no VLM meets the leakage criterion under them.

\begin{table*}[tb]
  \centering
  \footnotesize
  \setlength{\tabcolsep}{6pt}
  \begin{tabular}{@{}lccccc@{}}
    \toprule
    & \multicolumn{2}{c}{unrestricted} & Q2 correct & Q1 null & copy only \\
    \cmidrule(lr){2-3}
    reader & computed / wrong & $O/E$ & $O/E$ & $O/E$ & computed / wrong \\
    \midrule
    Gemma-4-31B & \qaLeakOGemma/\qaLeakErrGemma & \qaLeakGemma\ci{\qaLeakGemmaLo}{\qaLeakGemmaHi} & \qaKnowQTwoGemma\ci{\qaKnowQTwoGemmaLo}{\qaKnowQTwoGemmaHi} & \qOneNullGemma\ci{\qOneNullGemmaLo}{\qOneNullGemmaHi} & \qaCopyOGemma/\qaCopyErrGemma \\
    Gemma-4-12B & \qaLeakOGemmaTwelve/\qaLeakErrGemmaTwelve & \qaLeakGemmaTwelve\ci{\qaLeakGemmaTwelveLo}{\qaLeakGemmaTwelveHi} & \qaKnowQTwoGemmaTwelve\ci{\qaKnowQTwoGemmaTwelveLo}{\qaKnowQTwoGemmaTwelveHi} & \qOneNullGemmaTwelve\ci{\qOneNullGemmaTwelveLo}{\qOneNullGemmaTwelveHi} & \qaCopyOGemmaTwelve/\qaCopyErrGemmaTwelve \\
    Qwen3.8-27B & \qaLeakOQwen/\qaLeakErrQwen & \qaLeakQwen\ci{\qaLeakQwenLo}{\qaLeakQwenHi} & \qaKnowQTwoQwen\ci{\qaKnowQTwoQwenLo}{\qaKnowQTwoQwenHi} & \qOneNullQwen\ci{\qOneNullQwenLo}{\qOneNullQwenHi} & \qaCopyOQwen/\qaCopyErrQwen \\
    Qwen3.5-9B & \qaLeakOQwenVL/\qaLeakErrQwenVL & \qaLeakQwenVL\ci{\qaLeakQwenVLLo}{\qaLeakQwenVLHi} & \qaKnowQTwoQwenVL\ci{\qaKnowQTwoQwenVLLo}{\qaKnowQTwoQwenVLHi} & \qOneNullQwenVL\ci{\qOneNullQwenVLLo}{\qOneNullQwenVLHi} & \qaCopyOQwenVL/\qaCopyErrQwenVL \\
    Qwen3.5-4B & \qaLeakOQwenFour/\qaLeakErrQwenFour & \qaLeakQwenFour\ci{\qaLeakQwenFourLo}{\qaLeakQwenFourHi} & \qaKnowQTwoQwenFour\ci{\qaKnowQTwoQwenFourLo}{\qaKnowQTwoQwenFourHi} & -- & \qaCopyOQwenFour/\qaCopyErrQwenFour \\
    InternVL3.5-38B & \qaLeakOIntern/\qaLeakErrIntern & \qaLeakIntern\ci{\qaLeakInternLo}{\qaLeakInternHi} & \qaKnowQTwoIntern\ci{\qaKnowQTwoInternLo}{\qaKnowQTwoInternHi} & \qOneNullIntern\ci{\qOneNullInternLo}{\qOneNullInternHi} & \qaCopyOIntern/\qaCopyErrIntern \\
    Mistral-Small-3.2 & \qaLeakOMistral/\qaLeakErrMistral & \qaLeakMistral\ci{\qaLeakMistralLo}{\qaLeakMistralHi} & \qaKnowQTwoMistral\ci{\qaKnowQTwoMistralLo}{\qaKnowQTwoMistralHi} & \qOneNullMistral\ci{\qOneNullMistralLo}{\qOneNullMistralHi} & \qaCopyOMistral/\qaCopyErrMistral \\
    GLM-4.6V-Flash & \qaLeakOGlm/\qaLeakErrGlm & \qaLeakGlm\ci{\qaLeakGlmLo}{\qaLeakGlmHi} & \qaKnowQTwoGlm\ci{\qaKnowQTwoGlmLo}{\qaKnowQTwoGlmHi} & -- & \qaCopyOGlm/\qaCopyErrGlm \\
    LLaVA-OneVision-2-8B & \qaLeakOLlava/\qaLeakErrLlava & \qaLeakLlava\ci{\qaLeakLlavaLo}{\qaLeakLlavaHi} & \qaKnowQTwoLlava\ci{\qaKnowQTwoLlavaLo}{\qaKnowQTwoLlavaHi} & \qOneNullLlava\ci{\qOneNullLlavaLo}{\qOneNullLlavaHi} & \qaCopyOLlava/\qaCopyErrLlava \\
    Molmo2-O-7B & \qaLeakOMolmo/\qaLeakErrMolmo & \qaLeakMolmo\ci{\qaLeakMolmoLo}{\qaLeakMolmoHi} & \qaKnowQTwoMolmo\ci{\qaKnowQTwoMolmoLo}{\qaKnowQTwoMolmoHi} & \qOneNullMolmo\ci{\qOneNullMolmoLo}{\qOneNullMolmoHi} & \qaCopyOMolmo/\qaCopyErrMolmo \\
    \bottomrule
  \end{tabular}
  \caption{Leakage of the computed value in Q1 under restrictions, nulls and the copy-only paraphrases.}
  \label{S-tab:qaleak}
\end{table*}

\paragraph{Held-out pairs.}
The held-out question-answering set has \qaHeldPairs{} pairs (\qaHeldPairs{} false and \qaHeldPairs{} true equations) in the strata of the design-writer pairs, from \qaHeldWriters{} test writers, with glyphs assigned by position and no glyph instance used twice within a pair; its equations differ from those of every other set.
Its rule was registered before the readings were made; the two named VLMs are decided by a Bonferroni bound over \qaHeldFamily{} tests, and Table~\ref{S-tab:heldout} gives all 10 VLMs with 95\% intervals and no correction.

\begin{table*}[tb]
  \centering
  \footnotesize
  \setlength{\tabcolsep}{6pt}
  \begin{tabular}{@{}llll@{}}
    \toprule
    reader & Q1 null: computed / wrong, $O/E$ & restricted: computed / wrong, $O/E$ & copy only: computed / wrong, $O/E$ \\
    \midrule
    Gemma-4-31B & \qaHeldNullOGemma/\qaHeldNullErrorsGemma, \qaHeldNullGemma\ci{\qaHeldNullGemmaLo}{\qaHeldNullGemmaHi} & \qaHeldKnowOGemma/\qaHeldKnowErrorsGemma, \qaHeldKnowGemma\ci{\qaHeldKnowGemmaLo}{\qaHeldKnowGemmaHi} & \qaHeldCopyOGemma/\qaHeldCopyErrorsGemma, \qaHeldCopyGemma\ci{\qaHeldCopyGemmaLo}{\qaHeldCopyGemmaHi} \\
    Gemma-4-12B & \qaHeldNullOGemmaTwelve/\qaHeldNullErrorsGemmaTwelve, \qaHeldNullGemmaTwelve\ci{\qaHeldNullGemmaTwelveLo}{\qaHeldNullGemmaTwelveHi} & \qaHeldKnowOGemmaTwelve/\qaHeldKnowErrorsGemmaTwelve, \qaHeldKnowGemmaTwelve\ci{\qaHeldKnowGemmaTwelveLo}{\qaHeldKnowGemmaTwelveHi} & \qaHeldCopyOGemmaTwelve/\qaHeldCopyErrorsGemmaTwelve, \qaHeldCopyGemmaTwelve\ci{\qaHeldCopyGemmaTwelveLo}{\qaHeldCopyGemmaTwelveHi} \\
    Qwen3.8-27B & \qaHeldNullOQwen/\qaHeldNullErrorsQwen, \qaHeldNullQwen\ci{\qaHeldNullQwenLo}{\qaHeldNullQwenHi} & \qaHeldKnowOQwen/\qaHeldKnowErrorsQwen, \qaHeldKnowQwen\ci{\qaHeldKnowQwenLo}{\qaHeldKnowQwenHi} & \qaHeldCopyOQwen/\qaHeldCopyErrorsQwen, \qaHeldCopyQwen\ci{\qaHeldCopyQwenLo}{\qaHeldCopyQwenHi} \\
    Qwen3.5-9B & \qaHeldNullOQwenVL/\qaHeldNullErrorsQwenVL, \qaHeldNullQwenVL\ci{\qaHeldNullQwenVLLo}{\qaHeldNullQwenVLHi} & \qaHeldKnowOQwenVL/\qaHeldKnowErrorsQwenVL, \qaHeldKnowQwenVL\ci{\qaHeldKnowQwenVLLo}{\qaHeldKnowQwenVLHi} & \qaHeldCopyOQwenVL/\qaHeldCopyErrorsQwenVL, \qaHeldCopyQwenVL\ci{\qaHeldCopyQwenVLLo}{\qaHeldCopyQwenVLHi} \\
    Qwen3.5-4B & \qaHeldNullOQwenFour/\qaHeldNullErrorsQwenFour, \qaHeldNullQwenFour\ci{\qaHeldNullQwenFourLo}{\qaHeldNullQwenFourHi} & \qaHeldKnowOQwenFour/\qaHeldKnowErrorsQwenFour, \qaHeldKnowQwenFour\ci{\qaHeldKnowQwenFourLo}{\qaHeldKnowQwenFourHi} & \qaHeldCopyOQwenFour/\qaHeldCopyErrorsQwenFour, \qaHeldCopyQwenFour\ci{\qaHeldCopyQwenFourLo}{\qaHeldCopyQwenFourHi} \\
    InternVL3.5-38B & \qaHeldNullOIntern/\qaHeldNullErrorsIntern, \qaHeldNullIntern\ci{\qaHeldNullInternLo}{\qaHeldNullInternHi} & \qaHeldKnowOIntern/\qaHeldKnowErrorsIntern, \qaHeldKnowIntern\ci{\qaHeldKnowInternLo}{\qaHeldKnowInternHi} & \qaHeldCopyOIntern/\qaHeldCopyErrorsIntern, \qaHeldCopyIntern\ci{\qaHeldCopyInternLo}{\qaHeldCopyInternHi} \\
    Mistral-Small-3.2 & \qaHeldNullOMistral/\qaHeldNullErrorsMistral, \qaHeldNullMistral\ci{\qaHeldNullMistralLo}{\qaHeldNullMistralHi} & \qaHeldKnowOMistral/\qaHeldKnowErrorsMistral, \qaHeldKnowMistral\ci{\qaHeldKnowMistralLo}{\qaHeldKnowMistralHi} & \qaHeldCopyOMistral/\qaHeldCopyErrorsMistral, \qaHeldCopyMistral\ci{\qaHeldCopyMistralLo}{\qaHeldCopyMistralHi} \\
    GLM-4.6V-Flash & \qaHeldNullOGlm/\qaHeldNullErrorsGlm, \qaHeldNullGlm\ci{\qaHeldNullGlmLo}{\qaHeldNullGlmHi} & \qaHeldKnowOGlm/\qaHeldKnowErrorsGlm, \qaHeldKnowGlm\ci{\qaHeldKnowGlmLo}{\qaHeldKnowGlmHi} & \qaHeldCopyOGlm/\qaHeldCopyErrorsGlm, \qaHeldCopyGlm\ci{\qaHeldCopyGlmLo}{\qaHeldCopyGlmHi} \\
    LLaVA-OneVision-2-8B & \qaHeldNullOLlava/\qaHeldNullErrorsLlava, \qaHeldNullLlava\ci{\qaHeldNullLlavaLo}{\qaHeldNullLlavaHi} & \qaHeldKnowOLlava/\qaHeldKnowErrorsLlava, \qaHeldKnowLlava\ci{\qaHeldKnowLlavaLo}{\qaHeldKnowLlavaHi} & \qaHeldCopyOLlava/\qaHeldCopyErrorsLlava, \qaHeldCopyLlava\ci{\qaHeldCopyLlavaLo}{\qaHeldCopyLlavaHi} \\
    Molmo2-O-7B & \qaHeldNullOMolmo/\qaHeldNullErrorsMolmo, \qaHeldNullMolmo\ci{\qaHeldNullMolmoLo}{\qaHeldNullMolmoHi} & \qaHeldKnowOMolmo/\qaHeldKnowErrorsMolmo, \qaHeldKnowMolmo\ci{\qaHeldKnowMolmoLo}{\qaHeldKnowMolmoHi} & \qaHeldCopyOMolmo/\qaHeldCopyErrorsMolmo, \qaHeldCopyMolmo\ci{\qaHeldCopyMolmoLo}{\qaHeldCopyMolmoHi} \\
    \bottomrule
  \end{tabular}
  \caption{Question answering on the held-out pairs (pre-registered).}
  \label{S-tab:heldout}
\end{table*}

Table~\ref{S-tab:qactrl} gives the rates of Q3, the criterion $c$ and the controls.
The mask column gives the share of Q1 answers on the answer-masked image that report a result, the text column the share of Q1 answers to the equation as text that report the computed value, and the Q4 column the share of Q4 answers that give the digit of the computed value.

\begin{table}[tb]
  \centering
  \footnotesize
  \setlength{\tabcolsep}{3pt}
  \begin{tabular}{@{}lccccc@{}}
    \toprule
    & \multicolumn{2}{c}{Q3 on the image} & mask & text & Q4 \\
    \cmidrule(lr){2-3}
    reader & $H$ / $F$ & $c$ & \% & \% & \% \\
    \midrule
    Gemma-4-31B & \qaQThreeHitGemma / \qaQThreeFaGemma & \qaQThreeCGemma & \qaMaskGemma & \qaTextGemma & \qaQFourCompGemma \\
    Gemma-4-12B & \qaQThreeHitGemmaTwelve / \qaQThreeFaGemmaTwelve & \qaQThreeCGemmaTwelve & \qaMaskGemmaTwelve & \qaTextGemmaTwelve & \qaQFourCompGemmaTwelve \\
    Qwen3.8-27B & \qaQThreeHitQwen / \qaQThreeFaQwen & \qaQThreeCQwen & \qaMaskQwen & \qaTextQwen & \qaQFourCompQwen \\
    Qwen3.5-9B & \qaQThreeHitQwenVL / \qaQThreeFaQwenVL & \qaQThreeCQwenVL & \qaMaskQwenVL & \qaTextQwenVL & \qaQFourCompQwenVL \\
    Qwen3.5-4B & \qaQThreeHitQwenFour / \qaQThreeFaQwenFour & \qaQThreeCQwenFour & \qaMaskQwenFour & \qaTextQwenFour & \qaQFourCompQwenFour \\
    InternVL3.5-38B & \qaQThreeHitIntern / \qaQThreeFaIntern & \qaQThreeCIntern & \qaMaskIntern & \qaTextIntern & \qaQFourCompIntern \\
    Mistral-Small-3.2 & \qaQThreeHitMistral / \qaQThreeFaMistral & \qaQThreeCMistral & \qaMaskMistral & \qaTextMistral & \qaQFourCompMistral \\
    GLM-4.6V-Flash & n/a & n/a & \qaMaskGlm & \qaTextGlm & \qaQFourCompGlm \\
    LLaVA-OneVision-2-8B & \qaQThreeHitLlava / \qaQThreeFaLlava & \qaQThreeCLlava & \qaMaskLlava & \qaTextLlava & \qaQFourCompLlava \\
    Molmo2-O-7B & \qaQThreeHitMolmo / \qaQThreeFaMolmo & \qaQThreeCMolmo & \qaMaskMolmo & \qaTextMolmo & \qaQFourCompMolmo \\
    \bottomrule
  \end{tabular}
  \caption{Q3 rates and controls of the question-answering test.}
  \label{S-tab:qactrl}
\end{table}

\paragraph{Confusion-null test.}
Table~\ref{S-tab:cn} gives the counts for main Table~\ref{tab:phenomenon}: the repairs $O$, the expected count $E$, the number $n$ of aligned misreadings, the excess, $O/E$, and the share of all false-equation readings that are repairs.
The four recognizers repair \revRecRepairs{} of \revRecReadings{} false-equation readings, with an expected count of \testTwoRecEMin--\testTwoRecEMax{} per recognizer.

\begin{table*}[tb]
  \centering
  \footnotesize
  \setlength{\tabcolsep}{6pt}
  \begin{tabular}{@{}lrrrccr@{}}
    \toprule
    reader & $O$ & $E$ & $n$ & excess, points & $O/E$ & repairs, \% of readings \\
    \midrule
    Gemma-4-31B & \testTwoGemmaO & \testTwoGemmaE & \testTwoGemmaAligned & \testTwoGemmaExcess\ci{\testTwoGemmaExcessLo}{\testTwoGemmaExcessHi} & \testTwoGemmaRatio\ci{\testTwoGemmaRatioLo}{\testTwoGemmaRatioHi} & \testTwoGemmaRepair \\
    Gemma-4-12B & \testTwoGemmaTwelveO & \testTwoGemmaTwelveE & \testTwoGemmaTwelveAligned & \testTwoGemmaTwelveExcess\ci{\testTwoGemmaTwelveExcessLo}{\testTwoGemmaTwelveExcessHi} & \testTwoGemmaTwelveRatio\ci{\testTwoGemmaTwelveRatioLo}{\testTwoGemmaTwelveRatioHi} & \testTwoGemmaTwelveRepair \\
    Qwen3.8-27B & \testTwoQwenO & \testTwoQwenE & \testTwoQwenAligned & \testTwoQwenExcess\ci{\testTwoQwenExcessLo}{\testTwoQwenExcessHi} & \testTwoQwenRatio\ci{\testTwoQwenRatioLo}{\testTwoQwenRatioHi} & \testTwoQwenRepair \\
    Qwen3.5-9B & \testTwoQwenVLO & \testTwoQwenVLE & \testTwoQwenVLAligned & \testTwoQwenVLExcess\ci{\testTwoQwenVLExcessLo}{\testTwoQwenVLExcessHi} & \testTwoQwenVLRatio\ci{\testTwoQwenVLRatioLo}{\testTwoQwenVLRatioHi} & \testTwoQwenVLRepair \\
    Qwen3.5-4B & \testTwoQwenFourO & \testTwoQwenFourE & \testTwoQwenFourAligned & \testTwoQwenFourExcess\ci{\testTwoQwenFourExcessLo}{\testTwoQwenFourExcessHi} & \testTwoQwenFourRatio\ci{\testTwoQwenFourRatioLo}{\testTwoQwenFourRatioHi} & \testTwoQwenFourRepair \\
    InternVL3.5-38B & \testTwoInternO & \testTwoInternE & \testTwoInternAligned & \testTwoInternExcess\ci{\testTwoInternExcessLo}{\testTwoInternExcessHi} & \testTwoInternRatio\ci{\testTwoInternRatioLo}{\testTwoInternRatioHi} & \testTwoInternRepair \\
    Mistral-Small-3.2 & \testTwoMistralO & \testTwoMistralE & \testTwoMistralAligned & \testTwoMistralExcess\ci{\testTwoMistralExcessLo}{\testTwoMistralExcessHi} & \testTwoMistralRatio\ci{\testTwoMistralRatioLo}{\testTwoMistralRatioHi} & \testTwoMistralRepair \\
    GLM-4.6V-Flash & \testTwoGlmO & \testTwoGlmE & \testTwoGlmAligned & \testTwoGlmExcess\ci{\testTwoGlmExcessLo}{\testTwoGlmExcessHi} & \testTwoGlmRatio\ci{\testTwoGlmRatioLo}{\testTwoGlmRatioHi} & \testTwoGlmRepair \\
    LLaVA-OneVision-2-8B & \testTwoLlavaO & \testTwoLlavaE & \testTwoLlavaAligned & \testTwoLlavaExcess\ci{\testTwoLlavaExcessLo}{\testTwoLlavaExcessHi} & \testTwoLlavaRatio\ci{\testTwoLlavaRatioLo}{\testTwoLlavaRatioHi} & \testTwoLlavaRepair \\
    Molmo2-O-7B & \testTwoMolmoO & \testTwoMolmoE & \testTwoMolmoAligned & \testTwoMolmoExcess\ci{\testTwoMolmoExcessLo}{\testTwoMolmoExcessHi} & \testTwoMolmoRatio\ci{\testTwoMolmoRatioLo}{\testTwoMolmoRatioHi} & \testTwoMolmoRepair \\
    \bottomrule
  \end{tabular}
  \caption{Confusion-null test on \testTwoNPairs{} pairs of test writers.}
  \label{S-tab:cn}
\end{table*}

\paragraph{Meaningless-operator test.}
Table~\ref{S-tab:six} gives the counts of the meaningless-operator test.
Over the VLMs, the excess ranges from \sixExcessMin{} to \sixExcessMax{} points and the lower bound of $O/E$ from \sixRatioLoMin{} to \sixRatioLoMax.

\begin{table}[tb]
  \centering
  \footnotesize
  \setlength{\tabcolsep}{4pt}
  \begin{tabular}{@{}lrrrc@{}}
    \toprule
    reader & $O$ & $E$ & $n$ & $O/E$ \\
    \midrule
    Gemma-4-31B & 31 & 3.24 & 62 & 9.6 [2.8, 67.0] \\
    Gemma-4-12B & 55 & 6.17 & 138 & 8.9 [4.0, 14.3] \\
    Qwen3.8-27B & 30 & 3.38 & 101 & 8.9 [3.4, 21.8] \\
    Qwen3.5-9B & 50 & 6.24 & 142 & 8.0 [4.3, 11.2] \\
    Qwen3.5-4B & 68 & 4.18 & 144 & 16.3 [7.8, 24.2] \\
    InternVL3.5-38B & 50 & 3.86 & 103 & 12.9 [5.3, 25.0] \\
    Mistral-Small-3.2 & 34 & 2.94 & 109 & 11.6 [4.9, 33.5] \\
    GLM-4.6V-Flash & 13 & 1.80 & 59 & 7.2 [2.3, 25.8] \\
    LLaVA-OneVision-2-8B & 29 & 3.62 & 82 & 8.0 [3.1, 17.2] \\
    Molmo2-O-7B & 12 & 1.40 & 64 & 8.6 [3.1, 20.5] \\
    \bottomrule

{}
  \end{tabular}
  \caption{Meaningless-operator test on \sixNPairs{} pairs.}
  \label{S-tab:six}
\end{table}

\paragraph{Same-glyph test.}
Table~\ref{S-tab:sg} lists the context effect Y$-$X together with the effects of Y and X against the neutral context (Y$-$N and X$-$N), the effect seen under the position question, the effect on glyphs that the reader transcribes as $d$ in all three crop readings, and the shares of parsed readings in contexts X and Y.
Mistral-Small-3.2 and Molmo2-O-7B return a parsable transcription for fewer than half of the images, so their effects rest on fewer readings.

\begin{table*}[tb]
  \centering
  \footnotesize
  \setlength{\tabcolsep}{6pt}
  \begin{tabular}{@{}lcrrrcc@{}}
    \toprule
    reader & Y$-$X & Y$-$N & X$-$N & clear glyphs & position question & parsed X / Y, \% \\
    \midrule
    Gemma-4-31B & \testThreeGemmaDelta\ci{\testThreeGemmaDeltaLo}{\testThreeGemmaDeltaHi} & \testThreeGemmaYN & \testThreeGemmaXN & \testThreeGemmaCropDelta & \testThreeGemmaPos\ci{\testThreeGemmaPosLo}{\testThreeGemmaPosHi} & \revThreeGemmaParseX / \revThreeGemmaParseY \\
    Gemma-4-12B & \testThreeGemmaTwelveDelta\ci{\testThreeGemmaTwelveDeltaLo}{\testThreeGemmaTwelveDeltaHi} & \testThreeGemmaTwelveYN & \testThreeGemmaTwelveXN & \testThreeGemmaTwelveCropDelta & \testThreeGemmaTwelvePos\ci{\testThreeGemmaTwelvePosLo}{\testThreeGemmaTwelvePosHi} & \revThreeGemmaTwelveParseX / \revThreeGemmaTwelveParseY \\
    Qwen3.8-27B & \testThreeQwenDelta\ci{\testThreeQwenDeltaLo}{\testThreeQwenDeltaHi} & \testThreeQwenYN & \testThreeQwenXN & \testThreeQwenCropDelta & \testThreeQwenPos\ci{\testThreeQwenPosLo}{\testThreeQwenPosHi} & \revThreeQwenParseX / \revThreeQwenParseY \\
    Qwen3.5-9B & \testThreeQwenVLDelta\ci{\testThreeQwenVLDeltaLo}{\testThreeQwenVLDeltaHi} & \testThreeQwenVLYN & \testThreeQwenVLXN & \testThreeQwenVLCropDelta & \testThreeQwenVLPos\ci{\testThreeQwenVLPosLo}{\testThreeQwenVLPosHi} & \revThreeQwenVLParseX / \revThreeQwenVLParseY \\
    Qwen3.5-4B & \testThreeQwenFourDelta\ci{\testThreeQwenFourDeltaLo}{\testThreeQwenFourDeltaHi} & \testThreeQwenFourYN & \testThreeQwenFourXN & \testThreeQwenFourCropDelta & \testThreeQwenFourPos\ci{\testThreeQwenFourPosLo}{\testThreeQwenFourPosHi} & \revThreeQwenFourParseX / \revThreeQwenFourParseY \\
    InternVL3.5-38B & \testThreeInternDelta\ci{\testThreeInternDeltaLo}{\testThreeInternDeltaHi} & \testThreeInternYN & \testThreeInternXN & \testThreeInternCropDelta & \testThreeInternPos\ci{\testThreeInternPosLo}{\testThreeInternPosHi} & \revThreeInternParseX / \revThreeInternParseY \\
    Mistral-Small-3.2 & \testThreeMistralDelta\ci{\testThreeMistralDeltaLo}{\testThreeMistralDeltaHi} & \testThreeMistralYN & \testThreeMistralXN & \testThreeMistralCropDelta & \testThreeMistralPos\ci{\testThreeMistralPosLo}{\testThreeMistralPosHi} & \revThreeMistralParseX / \revThreeMistralParseY \\
    GLM-4.6V-Flash & \testThreeGlmDelta\ci{\testThreeGlmDeltaLo}{\testThreeGlmDeltaHi} & \testThreeGlmYN & \testThreeGlmXN & \testThreeGlmCropDelta & \testThreeGlmPos\ci{\testThreeGlmPosLo}{\testThreeGlmPosHi} & \revThreeGlmParseX / \revThreeGlmParseY \\
    LLaVA-OneVision-2-8B & \testThreeLlavaDelta\ci{\testThreeLlavaDeltaLo}{\testThreeLlavaDeltaHi} & \testThreeLlavaYN & \testThreeLlavaXN & \testThreeLlavaCropDelta & \testThreeLlavaPos\ci{\testThreeLlavaPosLo}{\testThreeLlavaPosHi} & \revThreeLlavaParseX / \revThreeLlavaParseY \\
    Molmo2-O-7B & \testThreeMolmoDelta\ci{\testThreeMolmoDeltaLo}{\testThreeMolmoDeltaHi} & \testThreeMolmoYN & \testThreeMolmoXN & \testThreeMolmoCropDelta & \testThreeMolmoPos\ci{\testThreeMolmoPosLo}{\testThreeMolmoPosHi} & \revThreeMolmoParseX / \revThreeMolmoParseY \\
    \bottomrule
  \end{tabular}
  \caption{Same-glyph test on \testThreeNItems{} glyphs of \testThreeNWriters{} test writers, in points.}
  \label{S-tab:sg}
\end{table*}

\paragraph{Column arithmetic and inequalities.}
Table~\ref{S-tab:col} lists for each VLM the ratio $O/E$ in the column test against the meaningless-operator null, the excess on clear glyphs, the share of readings that are repairs, and for the inequalities the ratio of misreadings that make the inequality true and the share of readings that give the specific true inequality; no value could be computed for GLM-4.6V-Flash on clear glyphs and on inequalities.

\begin{table*}[tb]
  \centering
  \footnotesize
  \setlength{\tabcolsep}{6pt}
  \begin{tabular}{@{}lccrcr@{}}
    \toprule
    & \multicolumn{3}{c}{column arithmetic} & \multicolumn{2}{c}{inequalities} \\
    \cmidrule(lr){2-4}\cmidrule(lr){5-6}
    reader & $O/E$ & clear glyphs, excess & repairs, \% & $O/E$ of truth & true inequality, \% \\
    \midrule
    Gemma-4-31B & \colGemma\ci{\colGemmaLo}{\colGemmaHi} & \colClearExGemma\ci{\colClearExGemmaLo}{\colClearExGemmaHi} & \colRepGemma & \ineqTruthGemma\ci{\ineqTruthGemmaLo}{\ineqTruthGemmaHi} & \ineqRepGemma \\
    Gemma-4-12B & \colGemmaTwelve\ci{\colGemmaTwelveLo}{\colGemmaTwelveHi} & \colClearExGemmaTwelve\ci{\colClearExGemmaTwelveLo}{\colClearExGemmaTwelveHi} & \colRepGemmaTwelve & \ineqTruthGemmaTwelve\ci{\ineqTruthGemmaTwelveLo}{\ineqTruthGemmaTwelveHi} & \ineqRepGemmaTwelve \\
    Qwen3.8-27B & \colQwen\ci{\colQwenLo}{\colQwenHi} & \colClearExQwen\ci{\colClearExQwenLo}{\colClearExQwenHi} & \colRepQwen & \ineqTruthQwen\ci{\ineqTruthQwenLo}{\ineqTruthQwenHi} & \ineqRepQwen \\
    Qwen3.5-9B & \colQwenVL\ci{\colQwenVLLo}{\colQwenVLHi} & \colClearExQwenVL\ci{\colClearExQwenVLLo}{\colClearExQwenVLHi} & \colRepQwenVL & \ineqTruthQwenVL\ci{\ineqTruthQwenVLLo}{\ineqTruthQwenVLHi} & \ineqRepQwenVL \\
    Qwen3.5-4B & \colQwenFour\ci{\colQwenFourLo}{\colQwenFourHi} & \colClearExQwenFour\ci{\colClearExQwenFourLo}{\colClearExQwenFourHi} & \colRepQwenFour & \ineqTruthQwenFour\ci{\ineqTruthQwenFourLo}{\ineqTruthQwenFourHi} & \ineqRepQwenFour \\
    InternVL3.5-38B & \colIntern\ci{\colInternLo}{\colInternHi} & \colClearExIntern\ci{\colClearExInternLo}{\colClearExInternHi} & \colRepIntern & \ineqTruthIntern\ci{\ineqTruthInternLo}{\ineqTruthInternHi} & \ineqRepIntern \\
    Mistral-Small-3.2 & \colMistral\ci{\colMistralLo}{\colMistralHi} & \colClearExMistral\ci{\colClearExMistralLo}{\colClearExMistralHi} & \colRepMistral & \ineqTruthMistral\ci{\ineqTruthMistralLo}{\ineqTruthMistralHi} & \ineqRepMistral \\
    GLM-4.6V-Flash & \colGlm\ci{\colGlmLo}{\colGlmHi} & -- & \colRepGlm & -- & \ineqRepGlm \\
    LLaVA-OneVision-2-8B & \colLlava\ci{\colLlavaLo}{\colLlavaHi} & \colClearExLlava\ci{\colClearExLlavaLo}{\colClearExLlavaHi} & \colRepLlava & \ineqTruthLlava\ci{\ineqTruthLlavaLo}{\ineqTruthLlavaHi} & \ineqRepLlava \\
    Molmo2-O-7B & \colMolmo\ci{\colMolmoLo}{\colMolmoHi} & \colClearExMolmo\ci{\colClearExMolmoLo}{\colClearExMolmoHi} & \colRepMolmo & \ineqTruthMolmo\ci{\ineqTruthMolmoLo}{\ineqTruthMolmoHi} & \ineqRepMolmo \\
    \bottomrule
  \end{tabular}
  \caption{Column arithmetic on \colPairs{} pairs and inequalities on \ineqPairs{} pairs.}
  \label{S-tab:col}
\end{table*}

\paragraph{Question-answering details.}
\paragraph{Design.}
We run the question-answering test on all of the \sixNPairs{} design-writer pairs (\sixNWriters{} writers); its decisions concern the \qaNFalse{} false equations with an error in the result.
The five registered questions and the post hoc copy-only question are each asked in three paraphrases, and every prompt is run once at temperature 0, so that every VLM gave \qaNAnswers{} answers per question.
Every image prompt begins with the fixed sentence ``The image shows a student's handwritten equation.'', and the first paraphrases continue with ``What number did the student write as the answer, to the right of the equals sign?'' (Q1), ``What is the correct result of the calculation to the left of the equals sign?'' (Q2) and ``Is the student's answer correct?'' (Q3).
Q1 and Q2 end with ``Answer with the number only.'' and Q3 with ``Answer yes or no.''; all eighteen prompts are given in Sec.~\ref{S-sec:prompts}.
As controls we use the answer-masked image, with the result digits removed, and the equation given as text.
Leakage is the number of Q1 errors that keep the digit count and are equal to the computed value, compared with the confusion null of the reader's result digits.
The paraphrases of the questions Q1, Q2 and Q3 are paired by an index fixed in advance, so each reading combines one answer to every question for the same image.
In the dissociation, leakage is restricted to the readings in which the Q2 answer equals the computed value and the Q3 answer is ``no''.
The null uses digit confusions estimated from transcriptions rather than from Q1 answers, and it is not conditioned on how the readings are selected by Q2 and Q3.
We therefore take the unrestricted leakage as the main effect size, a choice made after the results were known, and the restricted ratio as a secondary estimate that might be biased upward; in the data, the restriction raised $O/E$ for three of the four models that meet the criterion and lowered it slightly for InternVL3.5-38B.
For Q3 we report the sensitivity $d'$ and the criterion $c$ of signal detection theory~\cite{S-macmillan2005detection}, from the ``yes'' rates on true and false equations ($c>0$ indicates a bias toward ``no'').
In Table~\ref{S-tab:qa}, the Q1 and Q2 columns give the shares of answers equal to the computed value, Q3 $d'$ is given on the image and on the text, ``$O/E$, Q1 null'' is the unrestricted ratio against the post hoc null from Q1 answers to true equations, described below, ``computed / wrong'' gives, among the restricted readings, the number of wrong Q1 reports with the written digit count that equal the computed value over the total number of such wrong reports, and the last column gives the excess of leakage on clear false equations, $(O-E)$ divided by the number of aligned wrong Q1 reports, in points.
The other $O/E$ values use the meaningless-operator null; the registered decision uses the restricted $O/E$ (marked $^{\checkmark}$), and the marks $^{\dagger}$ apply the same criterion post hoc.

\paragraph{The dissociation.}
Leakage in the readings with a correct Q2 answer and a Q3 answer of ``no'' met the criterion for \qaDissPassed{} of the \qaDissTested{} testable VLMs (\qaDissNames; see Table~\ref{S-tab:qa} and main Fig.~\ref{fig:effects}).
In these restricted readings, \mbox{Gemma-4-12B} gives the computed value in \qaKnowOGemmaTwelve{} of its \qaKnowErrGemmaTwelve{} wrong reports with the written digit count, InternVL3.5-38B in \qaKnowOIntern{} of \qaKnowErrIntern{} of the written answer and Mistral-Small-3.2 in \qaKnowOMistral{} of \qaKnowErrMistral, with $O/E$ ranging from \qaKnowRatioMin{} to \qaKnowRatioMax{} over the models that meet the criterion.
\mbox{Gemma-4-31B} (lower bound \qaKnowGemmaLo), Qwen3.8-27B and Qwen3.5-9B produced too few errors of this kind to meet the criterion.
The restriction differs across the four models: InternVL3.5-38B answers ``no'' to \qaQThreeNoTrueIntern\% of the true equations, and Mistral-Small-3.2 and Molmo2-O-7B judge correctness near chance ($d'$ of \qaQThreeDpMistral{} and \qaQThreeDpMolmo), so Q3 is informative mainly for \mbox{Gemma-4-12B}.

\paragraph{A null from Q1 answers.}
As the null above comes from transcriptions, whose format differs from Q1, we estimate post hoc the digit confusions from those wrong Q1 answers to true equations that keep the digit count and compare the unrestricted Q1 leakage with this null.
The criterion is met by \qOneNullPassed{} VLMs, namely \qOneNullNames, whose $O/E$ values lie in \qOneNullRatioMin--\qOneNullRatioMax.
With placebo targets in place of the computed value (post hoc), the Q1 null gives $O/E$ of \placeboQaMin--\placeboQaMax{} on the main pairs and \placeboQaHeldMin--\placeboQaHeldMax{} on the held-out pairs, with no lower bound above 2.

On clear false equations, whose excess is given in the last column, the leakage of \qaClearPassed{} of the \qaClearTested{} testable VLMs met the criterion (\qaClearNames).

\paragraph{Controls.}
Given the equation as text, the VLMs report the computed value as written in \qaTextMin--\qaTextMax\% of their answers to Q1.
With the answer masked, the VLMs still give a result in \qaMaskMin--\qaMaskMax\% of Q1 answers, so some of them report a written answer although no result digits are visible, as if they read Q1 as a request for the correct result.
\mbox{Gemma-4-12B} and InternVL3.5-38B do this in only \qaMaskGemmaTwelve{} and \qaMaskIntern\%, whereas Mistral-Small-3.2 and Molmo2-O-7B do this in \qaMaskMistral{} and \qaMaskMolmo\%, so for these two a reading of Q1 as a request to compute cannot be ruled out even on the image.

\paragraph{A copy-only question.}
Post hoc, we asked Q1 again on the same images in three paraphrases that request a copy of the number after the equals sign even if it is not the correct result (Sec.~\ref{S-sec:prompts}).
Leakage is then not detected in any VLM; \mbox{Gemma-4-12B} and InternVL3.5-38B give the computed value in \qaCopyRateGemmaTwelve\% of their answers (upper bounds \qaCopyRateHiGemmaTwelve\% and \qaCopyRateHiIntern\%, against \qaQOneCompGemmaTwelve\% and \qaQOneCompIntern\% for Q1).
On the held-out pairs, their copy-only $O/E$ is \qaHeldCopyGemmaTwelve\ci{\qaHeldCopyGemmaTwelveLo}{\qaHeldCopyGemmaTwelveHi} and \qaHeldCopyIntern\ci{\qaHeldCopyInternLo}{\qaHeldCopyInternHi}, which does not exclude an effect relative to the few remaining errors.
One possible reading is that the models take Q1 as a request for the result the student should have written; another is that they guess illegible digits.

\paragraph{Held-out pairs.}

On \qaHeldPairs{} new pairs from \qaHeldWriters{} test writers and with the same VLMs, prompts and decoding, the rule registered before the readings required a bound of $O/E$ above 2 for each of \mbox{Gemma-4-12B} and InternVL3.5-38B under both the Q1 null and the restricted criterion, with a Bonferroni correction over these \qaHeldFamily{} tests (2 models times 2 nulls).
No test writer is a design writer, and some of them also wrote ink of the confusion-null and same-glyph sets, in other equations.
\mbox{Gemma-4-12B} gives $O/E$ \qaHeldNullGemmaTwelve\ci{\qaHeldNullGemmaTwelveLo}{\qaHeldNullGemmaTwelveHi} against the Q1 null and \qaHeldKnowGemmaTwelve\ci{\qaHeldKnowGemmaTwelveLo}{\qaHeldKnowGemmaTwelveHi} in the restricted readings, and InternVL3.5-38B gives \qaHeldNullIntern\ci{\qaHeldNullInternLo}{\qaHeldNullInternHi} and \qaHeldKnowIntern\ci{\qaHeldKnowInternLo}{\qaHeldKnowInternHi}, so the leakage of both models is confirmed on ink never used before.
Without correction, \qaHeldKnowPassed{} of the 10 VLMs meet the restricted criterion, and \qaHeldCopyAnyMet{} meet it with the copy-only question.

\section{Scan-like degradation}
\label{S-sec:degradation}

Under scan-like degradation (main Sec.~\ref{sec:further}), post hoc, the failure comes from fewer observed repairs: summed over the 10 VLMs, $O$ falls from \degCleanOSum{} to \degScanOSum{}, while $E$ is \degCleanESum{} on clean and \degScanESum{} on degraded renderings; aligned misreadings fall from \degCleanAlignedSum{} to \degScanAlignedSum{}, and the share among them that are repairs from \degCleanRepairShare\% to \degScanRepairShare\%, so degraded ink yields fewer misreadings that keep the form of the equation and, among them, a smaller share of repairs; this fits an override that acts on a well-formed equation, but the counts cannot tell an absent from an undetected effect.

\section{White-box tests}
\label{S-sec:whitebox}

\paragraph{Models and token groups.}
The white-box tests here run Qwen3.8-27B, Qwen3.5-9B and Qwen3.5-4B in bf16 with forward hooks on the transcription prompt.
The group T of target tokens consists of the visual tokens whose cells on the 32-pixel token grid overlap the bounding box of the target glyph, O of the visual tokens of all other glyphs, and P of the prefix tokens before the output position.

\paragraph{Results moved from the main paper.}
\paragraph{Decoding the written digit.}
On the same-glyph items, a linear probe trained on design writers (a logistic-regression classifier on internal activations) reads the written digit $d$ from the target glyph's visual tokens in \mchProbeXQwen\% of X and \mchProbeYQwen\% of Y images in Qwen3.8-27B, and in \mchProbeXQwenVL\% and \mchProbeYQwenVL\% in Qwen3.5-9B, while the transcription writes $d$ in \phYdQwen\% and \phYdQwenVL\% of Y readings.

\paragraph{Activation patching.}
Activation patching takes two runs on the transcription prompt, then replaces the activations of some selected token positions in one run by those of the other run and measures the change in the output.
We call the transcribed text before the target digit the \emph{prefix}.
We patch from context X into context Y the visual tokens that overlap the target glyph (T), the visual tokens of all other glyphs (O), or the prefix (P), and we report the removed share of the X--Y difference in the probability $P(y)$ of the output $y$, $(P_Y(y)-P_{\text{patched}}(y))/(P_Y(y)-P_X(y))$, where a value of 1 means the whole difference and 0 none of it (Table~\ref{S-tab:mech}).
For each model, patching acts on a band of consecutive layers that spans a quarter of its depth; the band was chosen from five candidate bands on design-writer items before the test items were run.
In all three models, P removes almost all of the X--Y difference in the patching test, where the Y run is given its own transcription up to the target digit, whereas T and O remove at most \mchPatchQwenFourO.
As the X and Y prefixes contain different operands, a P share near 1 is expected by construction; the output follows the transcribed text, but where the context acts is open.
The prefix split patches the digit tokens and the other tokens (operator, equals sign, markup) of the prefix separately and predicts that the digit tokens carry more of the effect.
In Qwen3.8-27B, as predicted, the digit tokens remove \pfxDQwen{} of the difference and the other tokens remove \pfxFQwen.
The prediction is not met in Qwen3.5-9B, where the other tokens remove \pfxFQwenVL{} and the prefix digit tokens \pfxDQwenVL.
The logit lens reads the token that the head of the language model would predict from intermediate hidden states; its prediction, that in repaired readings $d$ is ranked above $y$ at intermediate layers before the output switches to $y$, is met for \mchPOnePassed{} of 3 models (the two Qwen3.5 models).
With the prefix fixed, T is equivalent to zero (two one-sided tests, post hoc margin $\pm\tostMargin$; 90\% intervals [\tostQwenVLTLo, \tostQwenVLTHi] and [\tostQwenTLo, \tostQwenTHi] in Qwen3.5-9B and Qwen3.8-27B).

\begin{table}[tb]
  \centering
  \footnotesize
  \setlength{\tabcolsep}{2pt}
  \begin{tabular}{@{}llccc@{}}
    \toprule
    condition & patched & Qwen3.8-27B & Qwen3.5-9B & Qwen3.5-4B \\
    \midrule
    prefix fixed & T, target glyph & \mchPatchQwenT & \mchPatchQwenVLT & \mchPatchQwenFourT \\
     & O, other glyphs & \mchPatchQwenO & \mchPatchQwenVLO & \mchPatchQwenFourO \\
     & P, prefix & \mchPatchQwenP & \mchPatchQwenVLP & \mchPatchQwenFourP \\
     & prefix digits & \pfxDQwen & \pfxDQwenVL & -- \\
    free & O, other glyphs & \bidFreeOQwen & \bidFreeOQwenVL & -- \\
     & T, target glyph & \bidFreeTQwen & \bidFreeTQwenVL & -- \\
    \midrule
    items & & \mchPairsQwen & \mchPairsQwenVL & \mchPairsQwenFour \\
    \bottomrule
  \end{tabular}
  \caption{Share of the X--Y difference removed by patching.}
  \label{S-tab:mech}
\end{table}

\paragraph{The image route before the operands are written.}
The registered prediction of the free-generation test is that the fraction removed by O is above 0 (one-sided test), and a fraction of 0.5 was registered only as an estimation target value.
In Qwen3.8-27B, O then removes \bidFreeOQwen\ci{\bidFreeOQwenLo}{\bidFreeOQwenHi} of the difference, against \mchPatchQwenO{} with the prefix fixed, and the model writes the operands of context X followed by the faithful digit.
The prediction is also met in Qwen3.5-9B (Holm-adjusted $p=\bidHolm$), but in that model O removes only \bidFreeOQwenVL\ci{\bidFreeOQwenVLLo}{\bidFreeOQwenVLHi} of the difference, so a large effect of the other glyphs before the operands are written appears only for Qwen3.8-27B.
Because patching O changes the operands that the model then writes, the free-generation test cannot distinguish an image route to the result from a route through the written operands.
Intervals resample writers (\bidWritersQwen{} writers in Qwen3.8-27B).

\paragraph{Layer bands.}
Patching overwrites the residual stream of the selected token positions across a band of consecutive layers that spans a quarter of the model's depth.
For each model, the band was chosen among five candidate bands on design-writer items before the test items were run, and the logit lens was read over a band of late layers (Table~\ref{S-tab:bands}; layers are numbered from 0).

\begin{table}[tb]
  \centering
  \footnotesize
  \setlength{\tabcolsep}{4pt}
  \begin{tabular}{@{}lccc@{}}
    \toprule
    model & layers & patching band & logit-lens band \\
    \midrule
    Qwen3.8-27B & 64 & 24--39 & 51--62 \\
    Qwen3.5-9B & 32 & 20--27 & 19--30 \\
    Qwen3.5-4B & 32 & 16--23 & 20--30 \\
    \bottomrule
  \end{tabular}
  \caption{Layer bands of the white-box tests.}
  \label{S-tab:bands}
\end{table}

\paragraph{Logit lens.}
The measure is, among the context-Y readings that output $y$, the share in which the logit of $d$ exceeds that of $y$ at some layer of the lens band, and the prediction is met when this share reaches 0.6 with a one-sided bootstrap test against 0.5.
The measured share is \mchPOneQwen\ci{\mchPOneQwenLo}{\mchPOneQwenHi} in Qwen3.8-27B, \mchPOneQwenVL\ci{\mchPOneQwenVLLo}{\mchPOneQwenVLHi} in Qwen3.5-9B, and \mchPOneQwenFour\ci{\mchPOneQwenFourLo}{\mchPOneQwenFourHi} in Qwen3.5-4B.
On the readings of the meaningless-operator view, which act as a null control, the same share is \mchNullQwen, \mchNullQwenVL{} and \mchNullQwenFour.

\paragraph{Prefix split.}
The prefix split patches only the item pairs whose X and Y prefixes have the same length and the same arrangement of digit tokens and other tokens (\pfxPairsQwen{} pairs in Qwen3.8-27B and \pfxPairsQwenVL{} pairs in Qwen3.5-9B).
In a secondary analysis, the digit tokens are further split into the operand digits before the equals sign and the result digits that are already written after it.
In Qwen3.8-27B the operand digits remove \pfxDopQwen\ci{\pfxDopQwenLo}{\pfxDopQwenHi} of the difference and the result digits \pfxDresQwen\ci{\pfxDresQwenLo}{\pfxDresQwenHi}; in Qwen3.5-9B the values are \pfxDopQwenVL\ci{\pfxDopQwenVLLo}{\pfxDopQwenVLHi} and \pfxDresQwenVL\ci{\pfxDresQwenVLLo}{\pfxDresQwenVLHi}.

\paragraph{Free generation.}
In the free-generation test, we patch O or T from context X into context Y over the band during the processing of the image and the prompt, and the model then generates greedily without a fixed prefix.
The prediction is that O removes a positive share of the difference, and the Holm adjustment covers both models.
Under the result-first prompt of Sec.~\ref{S-sec:prompts}, which writes the target digit ahead of any operand, O removes \bidRevOQwen\ci{\bidRevOQwenLo}{\bidRevOQwenHi} in Qwen3.8-27B and \bidRevOQwenVL\ci{\bidRevOQwenVLLo}{\bidRevOQwenVLHi} in Qwen3.5-9B.
With the standard prompt the values are \bidFreeOQwen\ci{\bidFreeOQwenLo}{\bidFreeOQwenHi} and \bidFreeOQwenVL\ci{\bidFreeOQwenVLLo}{\bidFreeOQwenVLHi} (Table~\ref{S-tab:mech}), on \bidPairsQwen{} pairs from \bidWritersQwen{} writers and \bidPairsQwenVL{} pairs from \bidWritersQwenVL{} writers.

\section{An instruction to transcribe as written}
\label{S-sec:verbatim}

A rule registered before the readings repeats the held-out transcription test, for both views and all 10 VLMs, with a single sentence added to the standard prompt, ``Transcribe exactly what is written, even if the equation is wrong; do not correct anything.'', and with the seeds derived from the new prompt name.
Each VLM's null comes from its own meaningless-operator readings under the same prompt, and $Q$ is the ratio of $O/E$ with and without the sentence, with a paired writer bootstrap.
A VLM is counted as removed when it fails the criterion with the sentence and the upper bound of $Q$ is below 0.5, and as persisting when it meets the criterion and the lower bound of $Q$ is above 0.5.
In all, \vbRemoved{} VLMs are removed, \vbPersists{} persists and the others are undecided (Table~\ref{S-tab:verbatim}); GLM-4.6V-Flash, counted as removed by the rule, also fails the criterion without the sentence.
Repairs fall in all 10 VLMs, with writer-bootstrap intervals of the difference below zero, while the exact transcription of true equations falls by up to \vbExactTDropMax{} points.
In a first launch, eight VLMs were not served because the server of the previous model was still running; all 28{,}800 of their requests failed with an unknown-model error, none was scored, and the eight VLMs were read again.

\begin{table*}[tb]
  \centering
  \small
  \begin{tabular}{lcccrrrrl}
    \toprule
    VLM & $O/E$ standard & $O/E$ with sentence & $Q$ & \multicolumn{2}{c}{repair (\%)} & \multicolumn{2}{c}{exact true (\%)} & class \\
    & & & & std & sent. & std & sent. & \\
    \midrule
    Gemma-4-31B & 12.7 [4.6, 38.3] & 6.8 [2.7, 21.8] & 0.54 [0.14, 1.59] & 3.3 & 1.6 & 97.1 & 96.2 & undecided \\
    Gemma-4-12B & 15.2 [6.8, 28.9] & 0.0 [0.0, 0.0] & 0.00 [0.00, 0.00] & 6.8 & 0.0 & 92.0 & 91.0 & removed \\
    Qwen3.8-27B & 6.6 [3.6, 12.3] & 7.0 [3.5, 19.2] & 1.06 [0.61, 2.64] & 4.6 & 3.0 & 89.6 & 88.9 & persists \\
    Qwen3.5-9B & 7.3 [4.0, 11.4] & 5.8 [3.1, 8.6] & 0.80 [0.47, 1.19] & 6.4 & 4.1 & 90.6 & 90.7 & undecided \\
    Qwen3.5-4B & 19.5 [9.1, 32.6] & 14.4 [7.6, 29.1] & 0.74 [0.42, 1.71] & 8.0 & 6.2 & 90.2 & 89.7 & undecided \\
    InternVL3.5-38B & 18.4 [6.8, 57.3] & 8.9 [4.1, 27.2] & 0.48 [0.19, 1.18] & 6.9 & 2.6 & 94.1 & 93.0 & undecided \\
    Mistral-Small-3.2 & 9.1 [4.7, 23.0] & 4.4 [1.6, 11.2] & 0.48 [0.17, 1.01] & 5.4 & 1.6 & 41.2 & 37.6 & undecided \\
    GLM-4.6V-Flash & 2.1 [0.3, 9.3] & 0.0 [0.0, 0.0] & 0.00 [0.00, 0.00] & 0.9 & 0.0 & 85.7 & 73.3 & removed \\
    LLaVA-OneVision-2-8B & 7.9 [3.8, 20.9] & 8.0 [2.2, 30.1] & 1.01 [0.29, 3.54] & 2.7 & 1.1 & 91.0 & 89.8 & undecided \\
    Molmo2-O-7B & 11.1 [5.3, 23.9] & 0.9 [0.0, 7.5] & 0.08 [0.00, 0.67] & 2.9 & 0.1 & 48.7 & 38.2 & undecided \\
    \bottomrule
  \end{tabular}
  \caption{The held-out test with an instruction to transcribe as written, with 95\% intervals.}
  \label{S-tab:verbatim}
\end{table*}

\section{Fine-tuning}
\label{S-sec:lora}

Two pre-registered tests fine-tune three of the VLMs with LoRA (rank 16, alpha 32, learning rate $10^{-4}$, batch size 32, one epoch, language model only) on 3,000 false and true equations composed from the ink of CROHME 2019 training writers, none of whom is a test writer, with errors only in the result and the written symbols as the targets; the second test adds 3,000 general expressions from the same training set.
The three VLMs are Gemma-4-12B and the two Qwen models Qwen3.5-9B and Qwen3.8-27B.
Both are read on the held-out test with the standard prompt and three samples.
General transcription is the exact canonical match on the 1,196 expressions of the CROHME 2019 test set, read once at temperature 0; for the plain version this check is post hoc, for the mixed version it is pre-registered; Table~\ref{S-tab:lora} gives the per-model results.

\begin{table*}[tb]
  \centering
  \small
  \setlength{\tabcolsep}{4pt}
  \begin{tabular}{lccccccc}
    \toprule
    Model, version & $O/E$ base & $O/E$ tuned & $Q$ & $\Delta$ repair & $\Delta$ exact true & general base & general tuned \\
    \midrule
    Qwen3.5-9B, plain & \loraPlainQwenNineRatioBase & \loraPlainQwenNineRatio & \loraPlainQwenNineQ\ci{\loraPlainQwenNineQLo}{\loraPlainQwenNineQHi} & ${\loraPlainQwenNineRepairDiff}$ & ${\loraPlainQwenNineExactDiff}$ & \loraPlainQwenNineGenBase & \loraPlainQwenNineGen \\
    Gemma-4-12B, plain & \loraPlainGemmaTwelveRatioBase & \loraPlainGemmaTwelveRatio & \loraPlainGemmaTwelveQ\ci{\loraPlainGemmaTwelveQLo}{\loraPlainGemmaTwelveQHi} & ${\loraPlainGemmaTwelveRepairDiff}$ & ${\loraPlainGemmaTwelveExactDiff}$ & \loraPlainGemmaTwelveGenBase & \loraPlainGemmaTwelveGen \\
    Qwen3.8-27B, plain & \loraPlainQwenTwentySevenRatioBase & \loraPlainQwenTwentySevenRatio & \loraPlainQwenTwentySevenQ\ci{\loraPlainQwenTwentySevenQLo}{\loraPlainQwenTwentySevenQHi} & ${\loraPlainQwenTwentySevenRepairDiff}$ & ${\loraPlainQwenTwentySevenExactDiff}$ & \loraPlainQwenTwentySevenGenBase & \loraPlainQwenTwentySevenGen \\
    Qwen3.5-9B, mixed & \loraMixedQwenNineRatioBase & \loraMixedQwenNineRatio & \loraMixedQwenNineQ\ci{\loraMixedQwenNineQLo}{\loraMixedQwenNineQHi} & ${\loraMixedQwenNineRepairDiff}$ & ${\loraMixedQwenNineExactDiff}$ & \loraMixedQwenNineGenBase & \loraMixedQwenNineGen \\
    Gemma-4-12B, mixed & \loraMixedGemmaTwelveRatioBase & \loraMixedGemmaTwelveRatio & \loraMixedGemmaTwelveQ\ci{\loraMixedGemmaTwelveQLo}{\loraMixedGemmaTwelveQHi} & ${\loraMixedGemmaTwelveRepairDiff}$ & ${\loraMixedGemmaTwelveExactDiff}$ & \loraMixedGemmaTwelveGenBase & \loraMixedGemmaTwelveGen \\
    Qwen3.8-27B, mixed & \loraMixedQwenTwentySevenRatioBase & \loraMixedQwenTwentySevenRatio & \loraMixedQwenTwentySevenQ\ci{\loraMixedQwenTwentySevenQLo}{\loraMixedQwenTwentySevenQHi} & ${\loraMixedQwenTwentySevenRepairDiff}$ & ${\loraMixedQwenTwentySevenExactDiff}$ & \loraMixedQwenTwentySevenGenBase & \loraMixedQwenTwentySevenGen \\
    \bottomrule
  \end{tabular}
  \caption{LoRA fine-tuning: held-out test (differences in points) and CROHME 2019 test set (\%).}
  \label{S-tab:lora}
\end{table*}

\section{Photographed handwritten solutions (FERMAT)}
\label{S-sec:fermat}

The FERMAT set~\cite{S-fermat2025} (CC BY 4.0) contains photographs of handwritten solutions where a known error was introduced and written by hand, with the LaTeX of the written and of the original solution.
We kept the solutions whose first difference between the two texts replaces one integer right after an equals sign by another integer of the same length that differs in one or two digits, which yields \fermatItems{} solutions on \fermatPages{} pages.
Each of the 10 open VLMs then transcribed each image into LaTeX three times (once at temperature 0 and twice at temperature 1), which gives \fermatReadings{} readings per VLM.
The reading is aligned with the written text and the value at the error site counts as a repair when it equals the original, correct value.
Repairs are \fermatRepairMin{} to \fermatRepairMax{} readings per VLM (\fermatRateMin--\fermatRateMax\%).
There is no reading of these glyphs in a true equation, so these counts are not compared with each reader's own confusions.
A planned comparison with a uniform-landing null, in which a misread digit falls equally often on any value at the same distance, could not be used: placebo values drawn from the same set were read far less often than that null predicts (pooled ratio \fermatPlacebo), so that null is not calibrated on these images, and we report only the counts.
The placebo values themselves give a simpler comparison (post hoc): over the 10 VLMs, the placebo value is written in \fpPlaceboSum{} readings and the correct value in \fpRepairSum{} readings, although both are drawn from the same set at the same distance from the written value.

\section{Commercial models}
\label{S-sec:frontier}

We applied the rule of the held-out transcription test without change to two commercial models, under a pre-registration written before any commercial model read a stimulus and one addendum fixed before any score was computed.
The served versions were \texttt{\frontierLunaVersion} and \texttt{\frontierSolVersion}, which were read on 2 and 3 November 2026 (UTC) through an OpenAI-compatible gateway.
Each model read the \frontierReadings{} images of the held-out views (the equations and the meaningless-operator view, in three samples each) with the prompt, image encoding and seeds of the open run.
Both models reject temperature 0, so all the samples were drawn at the API default; the reasoning effort was the API default and the output limit, reasoning included, was 16{,}000 tokens, which no reply reached.
Mean output lengths were \frontierLunaTokens{} tokens (Luna) and \frontierSolTokens{} (Sol).
A pilot of 10 readings per view and model was kept in the full run, and the scope (all readings of both models) was fixed from the pilot's measured cost alone; the total cost was about \frontierLunaCost{} and \frontierSolCost{} JPY.
Three earlier starts of Luna with a broken reasoning setting (14 readings) were set aside before any score was computed and are released separately.
Under Holm's correction over the two models, Luna meets the criterion ($p_\mathrm{Holm}<0.001$, $O/E$ \frontierLunaRatio\ci{\frontierLunaRatioLo}{\frontierLunaRatioHi}), and Sol does not ($p_\mathrm{Holm}=\frontierSolPholm$, $O/E$ \frontierSolRatio\ci{\frontierSolRatioLo}{\frontierSolRatioHi}).
Repairs were \frontierLunaRepair\% (Luna) and \frontierSolRepair\% (Sol) of the readings of false equations.
As a post hoc description, \frontierLunaEqForm\% of Luna's and \frontierSolEqForm\% of Sol's readings of true equations had the form of a single-operation equation; most other Luna readings were transcriptions of unrelated notation, which the scoring code counts as non-equations.

After these results, the same rule was applied post hoc to Claude Opus 5.5 (served as \texttt{\fcVersion}), read on 15 November 2026 (UTC) through the Anthropic Messages endpoint of the same gateway with the same images, prompt and samples and an output limit of 16{,}000 tokens.
The thinking setting, effort and temperature were left at the API defaults, since the model rejects an explicit temperature; the mean output length was \fcTokens{} tokens and the cost about \fcCost{} JPY for \fcReadings{} readings.
Scored alone, without Holm's correction over the other models, it meets the criterion ($p=\fcP$, $O/E$ \fcRatio\ci{\fcRatioLo}{\fcRatioHi}); repairs were \fcRepair\% of the readings of false equations, and \fcEqForm\% of its readings of true equations had the form of a single-operation equation.

\section{Data release}
\label{S-sec:release}

\paragraph{Contents.}
The release contains the composed stimuli and their ground truth, all prompts (with the eighteen question prompts of Sec.~\ref{S-sec:prompts}), all model readings, the documents of the pre-registration, the code that composes the stimuli and reproduces every number of the paper from the stored readings, a datasheet and Croissant metadata.
The released benchmark is the regenerated set.

\paragraph{Datasheet summary.}
We, the authors, composed the data from the CROHME 2023 release of the CROHME test-set expressions, whose writers contributed handwriting to the competitions; no new handwriting was collected, and no personal information is added beyond that of the CROHME release.
The intended use is the evaluation of the faithful reading of handwritten arithmetic by recognition models and VLMs.
The stimuli are synthetic compositions of real ink and are not representative of student work, photographs or scanned pages.
The ground truth is fixed by construction and requires no annotation.
The human check of the main paper involved only two of the authors, who consented to it; no external participants were recruited and no personal data were collected, so no institutional review was sought.

\paragraph{Licences and metadata.}
The data are released under CC BY-NC-SA 3.0, which is the license of the CROHME 2023 release~\cite{S-crohme2023}, which restricts use to non-commercial purposes and requires derivatives to carry the same license; the code is distributed under the MIT license.
The dataset host will carry the same release with a persistent identifier.

{
    \small
    \providecommand{\url}[1]{\texttt{#1}}

}

\end{document}